\documentclass{article} 
\usepackage{iclr2023_conference,times}

\usepackage{amsmath,amsfonts,bm}

\def\eqref#1{equation~\ref{#1}}

\def\1{\bm{1}}

\DeclareMathAlphabet{\mathsfit}{\encodingdefault}{\sfdefault}{m}{sl}
\SetMathAlphabet{\mathsfit}{bold}{\encodingdefault}{\sfdefault}{bx}{n}

\newcommand{\data}{\textsc{Sangraha}}
\newcommand{\datasc}{\textsc{Sangraha Verified}}
\newcommand{\datasu}{\textsc{Sangraha Unverified}}
\newcommand{\datass}{\textsc{Sangraha Synthetic}}
\newcommand{\pipeline}{\textsc{Setu}}
\newcommand{\translation}{\textsc{IndicTrans2}}
\newcommand{\transliteration}{\textsc{IndicXlit}}
\newcommand{\llama}{\textsc{Llama2-70B Chat}}
\newcommand{\mixtral}{\textsc{Mixtral-8x7B-v0.1}}
\newcommand{\hhrlhf}{\textsc{HH-RLHF}}

\newcommand{\mistral}{\textsc{Mistral-7B Chat}}
\newcommand{\mc}{\textsc{mC4}}
\newcommand{\oscar}{\textsc{OSCAR}}
\newcommand{\roots}{\textsc{Roots}}
\newcommand{\culturax}{\textsc{CulturaX}}
\newcommand{\madlad}{\textsc{MADLAD-400}}

\newcommand{\setu}{\textsc{Setu}}
\newcommand{\setutranslate}{\textsc{Setu-Translate}}
\newcommand{\setutransliterate}{\textsc{Setu-Transliterate}}
\newcommand{\indiclid}{\textsc{IndicLID}}
\newcommand{\cld}{\textsc{cld3}}
\newcommand{\nllb}{\textsc{NLLB}}
\newcommand{\icvi}{\textsc{IndicCorp v1}}
\newcommand{\icvii}{\textsc{IndicCorp v2}}
\newcommand{\wikiconv}{\textsc{Wiki-Conv}}
\newcommand{\xlit}{\textsc{IndicXlit}}
\newcommand{\wikichat}{\textsc{Wiki-Chat}}
\newcommand{\indicalign}{\textsc{IndicAlign}}
\newcommand{\indicaligni}{\textsc{IndicAlign - Instruct}}
\newcommand{\indicalignt}{\textsc{IndicAlign - Toxic}}
\newcommand{\indicllmsuite}{\textsc{IndicLLMSuite}}
\newcommand{\aksharantar}{\textsc{Aksharantar}}
\newcommand{\guard}{\textsc{Toxic Matrix}}
\newcommand{\anudesh}{\textsc{Anudesh}}

\usepackage{hyperref}
\usepackage{url}
\usepackage{times}
\usepackage{latexsym}
\usepackage{xcolor}
\usepackage{multirow}
\usepackage{booktabs}
\usepackage{subfig}
\usepackage{enumitem}
\usepackage{inconsolata}
\usepackage{graphicx}
\usepackage{svg}
\usepackage{tabularx}
\usepackage{array}

\title{IndicLLMSuite: A Blueprint for Creating Pre-training and Fine-Tuning Datasets for Indian Languages}

\author{
  \textbf{Mohammed Safi Ur Rahman Khan$^{*1}$} \quad
  \textbf{Priyam Mehta\thanks{Equal Contribution. All author contributions listed in \ref{sec:authors}.}$~~^{1}$} 
  \quad
  \textbf{Ananth Sankar$^{1}$} \\
  \textbf{Umashankar Kumaravelan$^{1}$} \quad
  \textbf{Sumanth Doddapaneni$^{1,2}$} \quad 
  \textbf{Suriyaprasaad B$^{1,4}$}\thanks{Work done during an internship at Nilekani Center at AI4Bharat}\footnotemark[2]\\
  \textbf{Varun Balan G$^{1,5}$}\footnotemark[2]\quad
  \textbf{Sparsh Jain$^{1,6}$}\footnotemark[2]\quad
  \textbf{Anoop Kunchukuttan$^{1,2,3}$} \\
  \textbf{Pratyush Kumar$^{1,2,7}$} \quad
  \textbf{Raj Dabre$^{2,8}$} \quad
  \textbf{Mitesh M. Khapra$^{1,2}$}\thanks{\;Corresponding Author: Mitesh Khapra (\href{mailto:miteshk@cse.iitm.ac.in}{miteshk@cse.iitm.ac.in})}
  \\ \\
  $^{1}$Nilekani Centre at AI4Bharat \quad
  $^{2}$Indian Institute of Technology, Madras \quad
  $^{3}$Microsoft \quad \\
  $^{4}$Sant Longowal Institute of Engineering and Technology \quad
  $^{5}$IIIT D\&M Kancheepuram \quad \\
  $^{6}$Maharaja Agrasen Institute of Technology \quad 
  $^{7}$Sarvam AI \quad \\
  $^{8}$National Institute of Information and Communications Technology, Kyoto, Japan \quad
  \\
}

\iclrfinalcopy 
\begin{document}

\maketitle

\begin{abstract}
Despite the considerable advancements in English LLMs, the progress in building comparable models for other languages has been hindered due to the scarcity of tailored resources. 
Our work aims to bridge this divide by introducing an expansive suite of resources specifically designed for the development of Indic LLMs, covering 22 languages, containing a total of 251B tokens and 74.8M instruction-response pairs. Recognizing the importance of both data quality and quantity, our approach combines highly curated manually verified data, unverified yet valuable data, and synthetic data. We build a clean, open-source pipeline for curating pre-training data from diverse sources, including websites, PDFs, and videos, incorporating best practices for crawling, cleaning, flagging, and deduplication. For instruction-fine tuning, we amalgamate existing Indic datasets, translate/transliterate English datasets into Indian languages, and utilize LLaMa2 and Mixtral models to create conversations grounded in articles from Indian Wikipedia and Wikihow. Additionally, we address toxicity alignment by generating toxic prompts for multiple scenarios and then generate non-toxic responses by feeding these toxic prompts to an aligned LLaMa2 model. We hope that the datasets, tools, and resources released as a part of this work will not only propel the research and development of Indic LLMs but also establish an open-source blueprint for extending such efforts to other languages. The data and other artifacts created as part of this work are released with permissive licenses at \url{https://github.com/AI4Bharat/IndicLLMSuite}
\end{abstract}
\section{Introduction}

\begin{figure*}[htb]
    \centering
    \includegraphics[width=\linewidth]{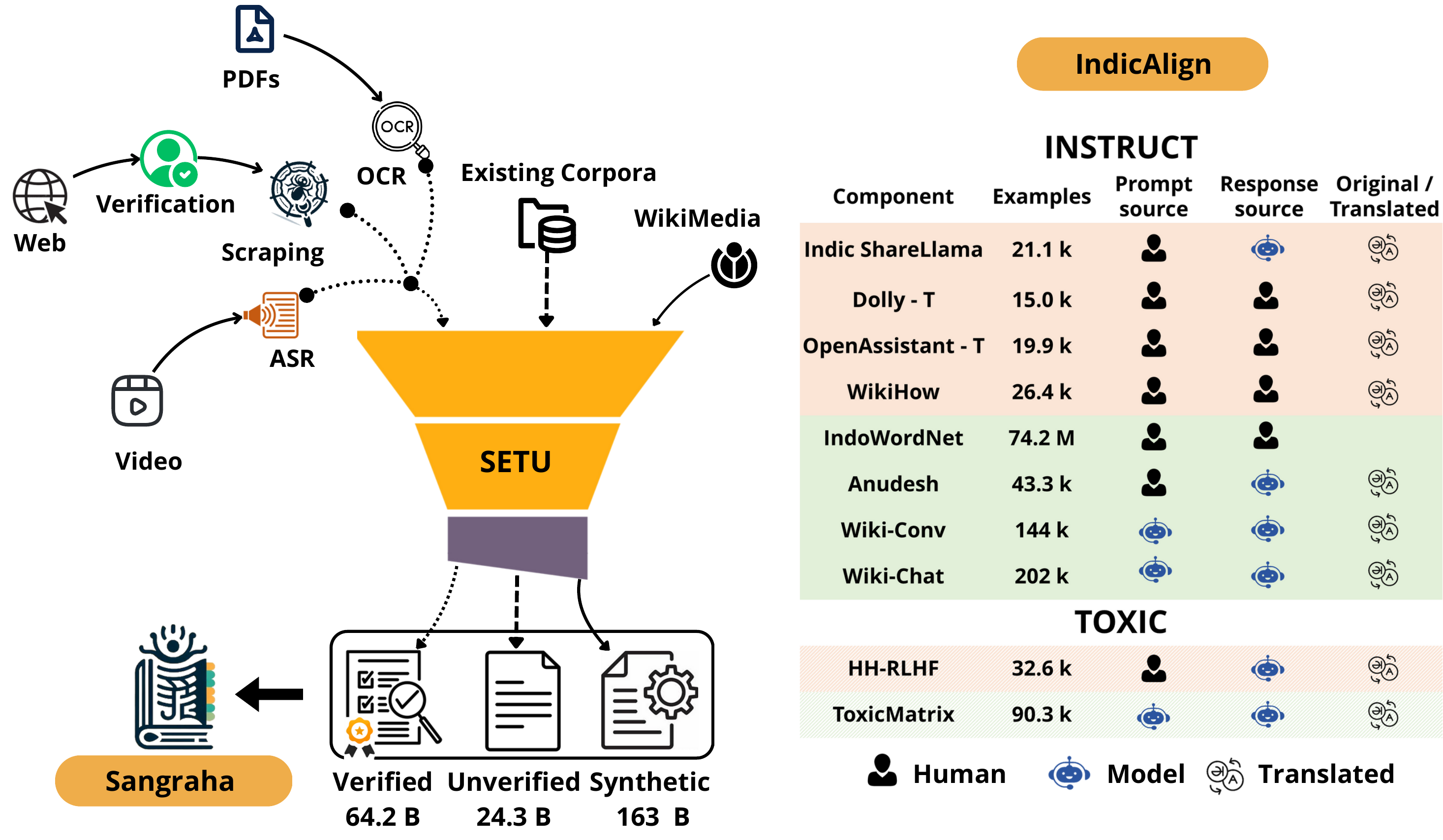}
    \caption{Overview of the different components present in \indicllmsuite{}. }
    \label{fig:sangraha_infographic}
\end{figure*}

Building Large Language Models (LLMs) is an inherently data-intensive process requiring a comprehensive set of resources for pre-training \citep{t5, xue2021mt5, pile, refinedweb, nguyen2023culturax, abadji-etal-2022-towards} and fine-tuning \citep{longpre2023flan, dolly, openassistant, ultrachat}. The last year has seen remarkable progress in building English LLMs, thanks to open-source models \citep{touvron2023llama1, touvron2023llama, jiang2023mistral, mixtral, almazrouei2023falcon} developed using comprehensive datasets containing such resources. Nonetheless, this progress has largely bypassed low and mid-resource languages due to the lack of data resulting from the lack of open-source pipelines for curating data for such languages from diverse sources such as websites (which require crawling and extraction), books (which require OCR) and videos (which require transcription). 
Further, for instruction fine-tuning, English LLMs now rely on model-generated data such as ShareGPT\footnote{\url{https://sharegpt.com/}}, Self-Instruct \citep{wang2023selfinstruct}, Evol-Instruct \citep{wizardlm}, UltraChat \citep{ultrachat}, etc. However, for low and mid resource languages, this option is not available due to lack of high-quality LLMs, leading to a chicken and egg problem, further widening the gap between the \textit{haves} and the \textit{have-nots}. 

A case in point is that of languages from the Indian sub-continent, which collectively are spoken by over 1.4 billion people. We focus on the 22 languages recognised in the 8th schedule of the Indian constitution.
These languages, despite their significant number of speakers, receive minimal representation in the training datasets and tokenizers of current open-source LLMs \citep{touvron2023llama, mixtral, almazrouei2023falcon} leading to a notable exclusion of their rich cultural contexts and nuances. In this work, we address this disparity 
by making the following contributions, as summarised in Figure \ref{fig:sangraha_infographic}:

\noindent \textbf{1. \data}: Pre-training data containing 251B tokens\footnote{We built a custom tokenizer which supports English and Indian languages and has an average fertility of 1.3 to 2.79 across the 22 languages. We use this tokenizer for all the reported statistics unless mentioned otherwise.} summed up over 22 languages, summarized in Table \ref{tab: sangraha_stats}, extracted from curated URLs, existing multilingual corpora, and large-scale translations.

\noindent \textbf{2. \pipeline}: Spark-based \citep{spark} distributed pipeline customized for Indian languages for extracting content from websites, PDFs and videos, with in-built stages for cleaning, filtering, toxicity removal and deduplication.

\noindent \textbf{3. \indicaligni}: A diverse collection of 74.7 million prompt-response pairs across 20 languages, summarized in Table \ref{tab:indicalign_stats}, collected through four methods: aggregating existing Instruction Fine-Tuning (IFT) datasets, translating English datasets into 14 Indian languages using an open-source translation model, creating context-grounded conversations from India-centric Wikipedia articles using open source LLMs, and establishing a crowdsourcing platform called \textit{Anudesh} for prompt collection. We also create a novel IFT dataset to teach the model language and grammar, by leveraging \textsc{IndoWordNet}~\citep{bhattacharyya2010indowordnet}, a lexically rich but rather neglected resource in the era of LLMs.

\noindent \textbf{4. \indicalignt}: 123K pairs of toxic prompt and non-toxic responses generated using open source English LLMs and translated to 14 Indian languages for safety alignment of Indic LLMs.

We collectively refer to the above as \indicllmsuite. We try to balance quality and quantity while acknowledging recent trends of using synthetic data for building powerful LLMs for English \citep{gunasekar2023textbooks,li2023textbooks} as well as low resource languages \citep{nguyen2023seallms,li2023bactrianx}. 
To ensure quality, we take help from humans to verify websites to flag noisy or machine translated content and to create toxicity lists for Indian languages. On the other hand, to ensure explicit representation of prompt-response pairs grounded in the Indian context we take the help of powerful open-source LLMs to generate grounded conversations from India-centric Wikipedia articles. 
We recognize the need to represent diverse knowledge and alignment information in Indic languages for better performance of LLMs in Indic languages. 
Hence, we undertake large-scale machine translation of rich English resources like Wikimedia as well as English fine-tuning datasets into Indian languages using SOTA open-source MT models.  
 
We thus balance source original data with translated and LLM-generated data to create the above collection.  

We believe that these choices can be replicated across other languages to create LLMSuites. All the code, tools and datasets developed as a part of this work will be publicly released and hopefully advance the development of LLMs for Indian languages. Given that LLM training is an expensive exercise, we plan to undertake community-effort to train LLMs, where multiple groups can pool together computing resources to build a high-quality Indic language LLM.

\section{Related Works}
We organise the Related Work into 3 sections in line with our main contributions.
\subsection{Multilingual Datasets}
Wikipedia has consistently served as the \textit{go to} repository of multilingual data and continues to be an important contributor for training data. Prior works such as \oscar~\citep{abadji-etal-2022-towards}, CC100~\citep{xlmr} (encompassing 100 languages), and \mc~\citep{xue2021mt5} (encompassing 101 languages) have been instrumental in generating data for a large set of languages through the meticulous processing of common crawl dumps\footnote{\url{https://commoncrawl.org/}}. The overarching methodology across all the works includes the systematic extraction of data, language identification, and subsequent stages of filtering and deduplication. Notably, CCNet~\citep{wenzek-etal-2020-ccnet} employs filters based on n-gram language models, while OSCAR and mC4~\citep{xue2021mt5} leverage heuristic-based filtering mechanisms. Efforts like Samanantar~\citep{ramesh-etal-2022-samanantar} and ROOTS~\citep{laurençon2023bigscience} have underscored the importance of aggregating existing datasets as an initial step towards multilingual corpus construction. ROOTS~\cite{laurençon2023bigscience}, with a specific focus on 59 languages, has additionally drawn attention to potential data duplications across disparate pipelines. Building on these works, \madlad~\citep{kudugunta2023madlad400} extends to 419 languages, while introducing human audit of data and iterative refinement processes, coupled with language family-specific filters. \culturax~\citep{nguyen2023culturax}, merges \mc{} and all versions of \oscar{}, followed by a rigorous cleaning pipeline to produce corpora in 167 languages.

\subsection{Data Curation}

\citet{DBLP:journals/tacl/KreutzerCWWEUTS22} has unilaterally showed the importance of auditing datasets. They discovered problems like wrong language, bad quality, offensive content, etc. Other works~\citet{gopher, refinedweb} reinforced the idea that using clean data is key to making better models.

\paragraph{Sources \& Scraping.} 
Majority of data curation pipelines start with Common Crawl as the internet source, followed by a series of cleaning steps to produce the final dataset~\citep{t5, xue2021mt5, xlmr, refinedweb}. Wikipedia is another common source of \emph{high quality} data that is used across many works~\citep{pile, together2023redpajama, soldaini2024dolma}. Research \citep{gpt3, touvron2023llama1}  further demonstrates the benefits of augmenting datasets with high-quality content, such as Wikipedia and Books corpus, for improved language model training. The first major hurdle for scraping data is extracting the actual text content from the HTML files and filtering out any boilerplate and unwanted HTML content. Tools like Trafilatura~\citep{DBLP:conf/acl/Barbaresi21} and jusText~\citep{DBLP:journals/polibits/EndredyN13} are widely used in recent works like RefinedWeb~\citep{refinedweb} and Pile~\citep{pile}.

\paragraph{Language Identification.} 
The subsequent challenge in the data cleaning process involves language detection, a task for which several LID tools have been employed in previous studies, namely, \cld~\citep{cld3}\footnote{\url{https://github.com/google/cld3/}}, langdetect\footnote{\url{https://github.com/shuyo/language-detection}}, fasttext~\citep{wenzek-etal-2020-ccnet, nllb}, SSLID~\citep{kudugunta2023madlad400}, among others. These tools demonstrate proficiency in identifying 55 to 500 languages; however, they fall short in encompassing all Indian languages considered in this study. Complications arise, particularly in the case of low-resource languages sharing the same script, leading to potential instances of mislabeling (e.g., Hindi and Marathi languages), necessitating the need for focused language family-specific detectors~\citep{madhani-etal-2023-bhasa}. 

\paragraph{Filtering.} 
One of the pivotal components in the data processing pipeline involves implementing various heuristics to ensure the quality of the processed data. These heuristics encompass rule-based approaches, targeting elements such as punctuation, repetitions, special characters, templated content, code snippets, etc. ~\citep{wenzek-etal-2020-ccnet, t5, xue2021mt5, laurençon2023bigscience, gopher}. Additionally, model-based tools, including n-gram language models perplexities~\citep{wenzek-etal-2020-ccnet, laurençon2023bigscience, nguyen2023culturax} and ML classifier-based approaches~\citep{gpt3}, play a crucial role in filtering out data that falls below certain quality thresholds. A prominent emphasis within the data cleaning process is the removal of toxic and harmful content, a priority shared across various datasets. Techniques employed for this purpose often include the utilization of word lists~\citep{t5}, blocklists of URLs~\citep{refinedweb}, and leveraging Google Safe Search~\citep{gopher}. Notably, \madlad~\citep{kudugunta2023madlad400} adopts language-specific heuristics, incorporating a manual audit of a small portion of data to enhance the cleaning process. This multifaceted approach underscores the importance of employing diverse strategies to ensure the reliability and safety of processed data in various contexts.

\paragraph{Deduplication.} 
Recent research findings shed light on the significant impact of deduplication on language models (LMs). \citet{DBLP:conf/iclr/CarliniIJLTZ23} demonstrated that deduplication effectively reduces memorization within LMs, while \citet{DBLP:conf/acl/LeeINZECC22} highlighted its role in enhancing LM performance. Furthermore, recent work \citet{DBLP:journals/corr/abs-2205-10487} underscored the detrimental effects of data repetition on model performance, particularly as model size increases. These insights underscore the critical importance of rigorous deduplication procedures in dataset preprocessing. Common approaches to deduplication encompass various techniques such as deduplication based on URLs, fuzzy techniques like MinHash~\citep{minhash} and SimHash~\citep{simhash}, alongside embedding techniques like those proposed by SemDedup~\citep{semdedup}.

With the rapid expansion of LLM development, a significant challenge arises from the proliferation of machine-generated text on the internet. It is important to devise effective strategies for filtering out such content to ensure the creation of high-quality data intended for human consumption. This necessitates a careful curation process that distinguishes between machine-generated and human-written text. Additionally, another issue pertains to the crawling and inclusion of benchmark/test data in the pre-training mixture i.e., Data Contamination~\citep{sainz-etal-2023-nlp, golchin2024time}. BigBench~\citep{DBLP:journals/corr/abs-2206-04615} advocates for filtering based on matching key strings, while recent research suggests encrypting benchmark data with passwords before distribution~\citep{DBLP:conf/emnlp/JacoviCGG23}.

\subsection{Supervised Fine-Tuning Datasets}
Pre-training demands huge amounts of data to effectively train on diverse linguistic patterns, while fine-tuning, specifically instruction tuning, necessitates comparatively smaller yet high-quality datasets~\citep{lima}. Broadly there exist two approaches for creating these datasets: (i) \emph{human-generated}, involving the manual input of humans for prompt creation and/or answer generation; and (ii) \emph{model-generated}, where both prompts and answers are generated by models.

\paragraph{Human Generated}
Dolly~\citep{dolly} and Open-Assistant~\citep{openassistant} are created through a comprehensive human annotation process from start to finish. Dolly, for instance, was curated by around 5000 Databricks employees to create nearly 15k instructions, whereas Open-Assistant emerged from a collaborative crowd-sourced initiative, accumulating 10k conversations spanning across 35 languages. These datasets encompass a wide array of tasks including question-answering, creative writing, classification, etc. On the flip side, datasets like ShareGPT\footnote{\url{https://sharegpt.com/}} and WildChat~\cite{zhao2024inthewildchat} are created by gathering real human interactions with ChatGPT\footnote{\url{https://chat.openai.com/}}. Similarly, HC3~\cite{hc3} gathered both human and ChatGPT responses for questions generated by humans on public datasets. These variations span diverse topics due to the extensive participation in creating these datasets.
\paragraph{Model Generated}
Distilling data from powerful models for the creation of datasets tailored for SFT tasks has become a widespread practice. Self-Instruct~\citep{selfinstruct} introduced an almost annotation-free data creation pipeline. Starting with just 175 seed prompts, they expand the dataset to 52k instructions iteratively. Alpaca~\citep{alpaca} streamlined the Self-Instruct process, leading to increased efficiency and cost reduction. Taking a further stride, Evol-Instruct~\citep{wizardlm} introduces a novel method of evolving the seed prompts iteratively along various axes. This resulted in the generation of an extensive dataset comprising 250k instructions. Further works generate data by simulating conversations between two or more model agents, with at least one acting as a User and the other acting as the Assistant.  CAMEL~\citep{camel}, for instance, crafted 115k instructions via multi-agent role play, while Ultrachat~\citep{ultrachat} produced 1.5 million multi-turn dialogues. Similarly, Baize~\citep{baize} generated 115k dialogues through self-chat with ChatGPT.

\section{\data{}}
\begingroup
\setlength{\tabcolsep}{15pt} 
\renewcommand{\arraystretch}{1} 
\begin{table}
    \centering
    \small
    \begin{tabular}{l|rrr|r}
        \toprule
        \textbf{Code} & \textbf{SV} & \textbf{SS} & \textbf{SU} & \textbf{Total Tokens} \\
        \midrule
        asm & 292.1   & 11,696.4  & 17.5   & 12,006.0  \\
        ben & 10,604.4 & 13,814.1  & 5,608.8 & 30,027.5  \\
        brx & 1.5      & -         & -       & 1.5  \\
        doi & 0.06    & -         & -       & 0.06\\
        eng & 12,759.9 & -         & -       & 12,759.9  \\
        gom & 10.1    & -          & -       & 10.1  \\
        guj & 3,647.9  & 12,934.5  & 597.0  & 17,179.4  \\
        hin & 12,617.3 & 9,578.7   & 12,348.3 & 34,544.3  \\
        kan & 1,778.3  & 12,087.4  & 388.8  & 14,254.5  \\
        kas & 0.5     & -         & -       & 0.5  \\
        mai & 14.6    & -         & -       & 14.6  \\
        mal & 2,730.8  & 13,130.0  & 547.8  & 16,408.6  \\
        mar & 2,827.0  & 10,816.7  & 652.1  & 14,295.8  \\
        mni & 7.4     & -         & -       & 7.4  \\
        npi & 1,822.5  & 10,588.7  & 485.5  & 12,896.7  \\
        ori & 1,177.1  & 11,338.0  & 23.7   & 12,538.8  \\
        pan & 1,075.3  & 9,969.6   & 136.9  & 11,181.8  \\
        san & 1,329.0  & 13,553.5   & 9.8    & 14,892.3  \\
        sat & 0.3     & -         & -       & 0.3  \\
        snd & 258.2   & -         & -       & 258.2  \\
        tam & 3,985.1  & 11,859.3  & 1,515.9  & 17,360.3  \\
        urd & 3,658.1  & 9,415.8   & 1,328.2  & 14,402.1 \\
        tel & 3,706.8  & 11,924.5  & 647.4  & 16,278.7  \\
        \midrule
        \textbf{Total} &  64,306.1 & 162,707.9 &  24,307.7 & \textbf{251,321.0}\\
        \bottomrule
    \end{tabular}
    \caption{Number of tokens (in Millions) in each split of Sangraha. (SV: \datasc, SS: \datass, SU: \datasu). We represent the languages in this document using the \textsc{ISO 639-3} standard codes}
    \label{tab: sangraha_stats}
\end{table}
\endgroup

In this section, we describe the composition and curation process of \data{} spanning verified (64B), unverified (24B), and synthetic (162B) content for a total of 251B tokens. Table \ref{tab: sangraha_stats} shows the language level tokens distribution in each of the splits. 

\subsection{Sangraha Verified}
We introduce \datasc{}, a high-quality dataset, which adds human verification at various stages of its curation. A major chunk of this includes data crawled from high-quality, manually verified Indic language websites. Additionally, recognizing the fact that a significant amount of Indic language text is locked in PDFs and audio, we also collect data from various books/documents and videos resulting in a total of 64B tokens. Table \ref{tab: sv stats} shows the language level tokens distribution across the Web, PDF, and Speech data in \datasc{}.


\begingroup
\setlength{\tabcolsep}{15pt} 
\renewcommand{\arraystretch}{1} 
\begin{table}[t]
    \centering
    \small
    \begin{tabular}{l|rrr}
    \toprule
 & \multicolumn{3}{c}{\textbf{Sangraha Verified}}\\

\multirow{-2}{*}{\textbf{Code}} & \multicolumn{1}{c}{\textbf{Web}} & \multicolumn{1}{c}{\textbf{PDFs}} & \multicolumn{1}{c}{\textbf{Speech}} \\
\midrule
asm  & 128.6        & 162.9   & 0.6 \\
ben  & 6,398.8       & 4,132.5 & 73.0 \\
brx  & 1.3          & 0.1   & -        \\
doi  & 0.03         & 0.03   & -        \\
eng  & 12,190.9      & 542.9 & 26.0       \\
gom  & 9.3          & 0.7   & -        \\
guj  & 2,370.2        & 1,256.6& 21.0       \\
hin  & 9,312.5       & 2,435.7 & 869.0      \\
kan  & 1,415.1       & 350.2 & 12.9     \\
kas  & 0.3          & 0.06   & 0.04     \\
mai  & 14.5         & 0.06   & -        \\
mal  & 2,539.5       & 78.8   & 112.4    \\
mar  & 2,415.5       & 397.2 & 14.2     \\
mni  & 0.9          & 5.8    & 0.6     \\
npi  & 1,809.0       & 13.3  & -        \\
ori  & 653.3        & 523.3  & 0.4     \\
pan  & 863.1        & 211.6 & 0.5     \\
san  & 54.3         & 1,274.6 & -        \\
sat  & 0.06          & -      & 0.2     \\
snd  & 257.9        & 0.23   & -        \\
tam  & 3,345.3       & 611.1 & 28.6     \\
tel  & 2,934.8       & 753.3 & 18.5    \\
urd  & 1,836.5       & 1,798.4 & 23.1    \\
\midrule
\textbf{Total} & 48,552.8     & 14,550.0 & 1,203.2   \\   
\bottomrule
\end{tabular}
    \caption{Number of tokens (in Millions) in each component of \datasc{}.}
    \label{tab: sv stats}
\end{table}
\endgroup

\subsubsection{Web Data}
Our web data, constituting most of Sangraha, diverges from traditional Common Crawl-based approaches by prioritizing data quality. This involves manual verification of each website before scraping. We adopt a three-fold strategy to collect a comprehensive collection of verified websites for scraping. Firstly, we extend the efforts of \citet{kakwani-etal-2020-indicnlpsuite} and \citet{doddapaneni-etal-2023-towards} of discovering web sources using existing news repositories and automated web searches using popular keywords to discover a large list of Indic language websites. But, unlike the previous efforts, we do not restrict ourselves to just news websites. Secondly, we identify various domains such as Indian Culture, Food, Health, Travel, among others and enlist volunteers to gather websites within these domains, prioritizing those in Indic languages or in English but pertaining to Indian context. Thirdly, we collect base URLs from \mc~\citep{xue2021mt5}, focusing on websites with high amount of content, and get them verified by volunteers. Additionally, we also include all the Indian Government websites\footnote{\url{https://igod.gov.in/}}, which serves as a valuable resource, given their multilingual content.

Volunteers review all the websites collected via automated methods and decide on acceptance or rejection based on the defined criteria. A website can be rejected if either of the below conditions were met:
\begin{itemize}
    \item Website is non-Indic or non-English.
    \item Website is an adult, gambling, or a general toxic website.
    \item Website has content that clearly appears to be machine-translated.
\end{itemize}

\begin{figure*}
    \centering
    \subfloat[\centering Accepted Website statistics - Domain information ]{\includegraphics[width=0.4\linewidth]{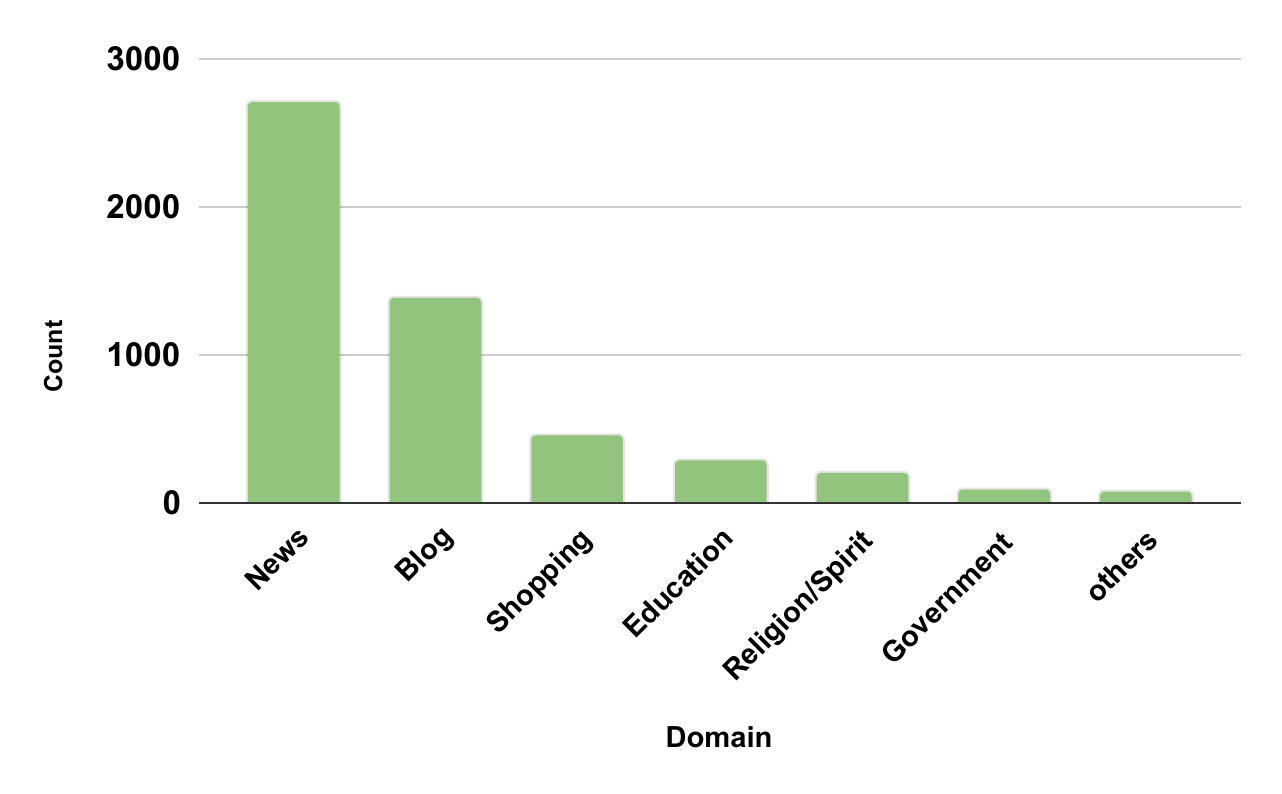}}
    \qquad
    \subfloat[\centering Rejected Website statistics - Rejection Reason]{\includegraphics[width=0.4\linewidth]{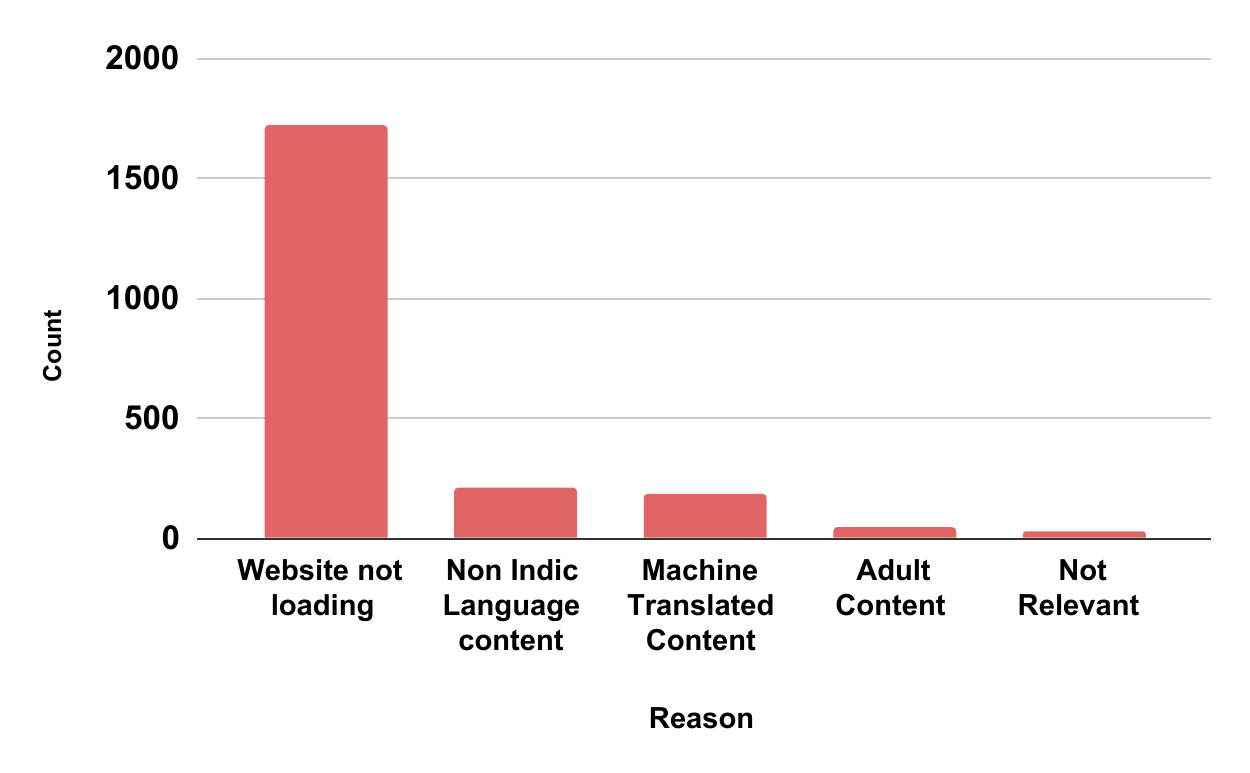}}
    \caption{}
    \label{image:url_verification_stats}
\end{figure*}

Figure \ref{image:url_verification_stats} presents the verification outcomes, highlighting a significant rejection rate due to website inactivity, particularly those sourced from \mc{}. This means that the information in existing collections is becoming outdated because of defunct websites. We make available both the verification portal and the list of validated URLs for further research utilization. We then used the open-source \textit{webcorpus}\footnote{\url{https://github.com/AI4Bharat/webcorpus}} toolkit to crawl the verified websites.

\subsubsection{PDF Data}
Given that a lot of Indic language content is locked in various digitized PDF documents, we focus on text extraction from them using high-quality OCR systems. We employ GCP's Vision Tool for performing OCR as it is known to give good performance across different categories \citep{ocr-perf}. We source the PDFs from 7 broad sources as shown in Table \ref{tab:pdf_sources}.
\begin{table}[]
    \centering
    \def\arraystretch{1.5}
    \begin{tabular}{lrr}
        \toprule
        \textbf{PDF Sources} & \textbf{\#PDFs} & \textbf{\#Pages} \\
        \midrule
        Internet Archive     & 437,225 &  74M\\
        eGyanKosh            & 5,133                 & 88K               \\
        Indian Parliament    & 30,964                & 2.7M              \\
        AIR News             & 74,353                & 148K    \\
        Govt. Magazines      & 895                  & 46K                \\
        School Books         & 4,315                 & 359K               \\
        Miscellaneous        & 27,988                  & 4.6M               \\
        \midrule
        \textbf{Total}       & \textbf{507,419}            & \textbf{82M}\\
        \bottomrule
        \end{tabular}
    \caption{Sangraha PDF sources - The final statistic of the PDFs on which OCR has been performed.}
        \label{tab:pdf_sources}
\end{table} 

\textbf{Internet Archive}

Utilizing the official API of the Internet Archive\footnote{\url{https://archive.org/developers/internetarchive/}}, we collected approximately 921K PDF documents across all Indic languages. This collection spans diverse categories such as religious texts, news articles, fiction, educational materials, and scientific literature. We subsequently filtered out PDFs incompatible with GCP Vision\footnote{\url{https://cloud.google.com/vision/docs/languages}}, specifically excluding PDFs in languages like Bodo, Dogri, Kashmiri, Konkani, Maithili, Manipuri, and Sindhi, with plans for future inclusion.

To optimize for quality and manage costs, OCR was performed only on high-quality PDFs. We first remove all the corrupted and encrypted PDFs. Additionally, resource limitations from GCP Vision necessitated the filtering of PDFs exceeding 2000 pages. We also filter out all  PDFs having less than 25 pages as these are often incoherent documents such as glossaries, comics, bills, and receipts. This was followed by filtering out scanned PDFs with a Pixel Per Inch (PPI) value below 300 to filter out blurry PDFs. Additionally, we analyzed images from 10 consecutive pages of each PDF, measuring the average area covered by images and their brightness. Pages with images were considered for further analysis if they covered less than 50\% of the page area and had a brightness level above 200. Table \ref{tab:internet_archive_filtering} shows the statistics of PDFs filtered after each filtering stage.

\textbf{eGyanKosh}

eGyanKosh, India's National Digital Repository, is a repository for digital learning resources from Open and Distance Learning Institutions. We collect PDF documents covering various subjects, such as History, Economics, Political Science, Public Administration, and Sociology, across various Indian languages.

\textbf{Indian Parliament}

This source comprises manually compiled summaries of debates and discussions from the Indian Parliament and various State Legislative Assemblies. These form a rich source of local and culturally relevant data. We collect all the publicly available Parliamentary and State Assembly materials. Table \ref{tab:parilament_stats} shows the state-wise statistics of the collected documents.

\textbf{AIR News}

All India Radio (AIR) is the national radio broadcaster of India, a Prasar Bharati division, that streams radio programs in all major Indian languages. Following the approach of \citep{bhogale2022effectiveness}, we download all news bulletins for 12 Indian languages. Table \ref{tab:air_news} shows the language level statistics of the collected data. 

\textbf{Government Magazines}

We aggregated content from magazines published by various governmental agencies, which include annual reports, details on governmental schemes, initiatives, cabinet decisions, and current affairs, that are published in multiple Indian languages.

\textbf{School Textbooks}

This set includes publicly available textbooks from various Indian states and those published by the National Council of Educational Research and Training (NCERT), providing a rich source of educational content in multiple Indian languages. Table \ref{tab:school_books} shows the statistics of the books collected from different states.

In addition to the above categorized sources, we also incorporated a variety of documents from government and public domains, focusing on content either in Indic languages or in English with relevance to India. Our future work will continue to explore digitization and OCR of new
public sources.

\subsubsection{Speech Data}
Similar to PDFs, a bulk of language data is present in audio forms either in videos, podcasts, radio broadcasts, etc. This data captures the most natural way of human interactions in the form of conversations. We collect a variety of both manually as well as automatically transcribed sources, covering a broad variety of content, as shown in Table \ref{tab:speech_stats}.
\begin{table}[]
    \centering
    \def\arraystretch{1.5}
    \begin{tabular}{lr}
        \toprule
        \textbf{Source}      & \textbf{Number of Instances} \\
        \midrule
        YouTube - Hindi      &  276K videos                         \\
        Open Subtitles       &  14K movies                           \\
        NPTEL - Transcripts  &  1.4K courses                         \\
        Mann Ki Baat         &  1.4K podcasts                       \\
        Others               &  15K                     \\
        \midrule
        \textbf{Total}       & \textbf{309K}    \\
        \bottomrule
        \end{tabular}
    \caption{Statistics of the various sources of Speech Data collected}
    \label{tab:speech_stats}
\end{table}

\textbf{Youtube - Hindi}

Following the approach of \citet{anonymous}, we collect around 80K hours of audio data from YouTube videos in Hindi language. We then chunk it into smaller segments by detecting silences using WebRTC VAD\footnote{\url{https://github.com/wiseman/py-webrtcvad}} and get each chunk transcribed using the Hindi Conformer model. We then piece together the transcripts of each individual chunk to get the transcript of the whole video.

\textbf{OpenSubtitles}

Following \citet{pile}, we collect all the Indic Language subtitles from OpenSubtitles\footnote{\url{https://www.opensubtitles.org/}}. We first process the SRT files using simple regex-based patterns to remove the timestamps and extract the text. We then define regex patterns to filter out other noisy content like character cues, continuation ellipses, etc. We then combine the different parts to form a single document per SRT file. Table \ref{tab:opensubtitles} shows the language-wise statistics of Subtitles.

\textbf{NPTEL - Transcripts}

The National Programme on Technology Enhanced Learning (NPTEL)\footnote{\url{https://nptel.ac.in/}} is an Indian e-learning platform for university-level science, technology, engineering, and mathematics subjects that is jointly developed by various Indian Institutes. Although the course content developed by NPTEL is primarily in English, much of it has been manually transcribed and translated into 11 different Indian Languages and reviewed before being made publicly available. The translated content has been compiled and released as course textbooks. Table \ref{tab:nptel_books_stats} shows the statistics of the course transcripts available in different languages.

\textbf{Mann Ki Baat}

Mann Ki Baat is an Indian Radio programme hosted by the Indian Prime Minister, usually with a frequency of 1 programme per month. This is transcribed and then manually translated into 13 Indian languages. Table \ref{tab:mkb_stats} shows the language-wise statistics.

\subsection{Sangraha Synthetic}
There is a huge disparity between the information-rich digital content and knowledge available in English compared to Indian languages.
To address this disparity, we introduce \datass, an initiative to democratize access to knowledge by translating a knowledge-rich English corpus into Indian languages. Utilizing \translation~\citep{gala2023indictrans2}, we translated the entirety of English Wikimedia into 14 Indian languages resulting in nearly 90B tokens.
Since \translation{} operates at the sentence level and does not retain the document level formatting such as newlines, markdowns and other structures, we developed the \setutranslate{} pipeline, described in Section \ref{setu-translate}. This pipeline facilitates the translation of documents and conversations while preserving the original document structure.

Recognizing the prevalent trend of ``Romanized'' Indic language usage, particularly in informal settings and in digital communication, we extend \citet{husain2024romansetu} and transliterate the above-translated content in 14 languages to Roman script using \transliteration~\citep{madhani-etal-2023-aksharantar} resulting in about 72B tokens. Going forward, we will extend \datass{} to cover all the 22 scheduled languages of India as well as translate other knowledge-rich sources.

\subsection{Sangraha Unverified}
We introduce the \datasu{} split to expand the Sangraha corpus while ensuring high quality. We employ a perplexity filtering pipeline, inspired by CCNet \citep{wenzek-etal-2020-ccnet}, to collect all the high-quality tagged documents from \culturax{}~\citep{nguyen2023culturax} and \madlad{}~\citep{kudugunta2023madlad400}. We consider \culturax{} and \madlad{} as these represent the latest and most comprehensive multilingual collections of Web data.

We first randomly sample 200,000 documents from the \datasc{} split for each language. We then normalize each document by converting text to lowercase, removing accents from characters, normalizing numbers to a uniform representation (specifically converting all digits to "0"), replacing a predefined set of Unicode punctuations with their ASCII counterparts, and removing non-printing characters. We then train a sentencepiece tokenizer and tokenize all of the sampled data. Then, we train a 5-gram Kneser-Ney model using the KenLM \citep{heafield-2011-kenlm} library. We binarize these models for quicker inference. 

For deciding the language-specific thresholds, we create a validation set by sampling another 100,000 documents from \datasc{} and calculate the perplexity of each document using the trained n-gram models. We then sort the perplexities and choose the 80th percentile value as the threshold for each language. Table \ref{tab:perplexity_stats} shows the thresholds chosen for each language. Higher percentile thresholds can be chosen to prefer more quality over volume, but that may result in reduced diversity and representativeness of the resultant data.
\begin{table*}[]
    \centering
    \small
    \begin{tabular}{lrrrrrrr}
\toprule
\textbf{Lang} &
  \textbf{\begin{tabular}[r]{@{}r@{}}Min \\ Perplexity\end{tabular}} &
  \textbf{\begin{tabular}[r]{@{}r@{}}Max \\ Perplexity\end{tabular}} &
  \textbf{\begin{tabular}[r]{@{}r@{}}Mean \\ Perplexity\end{tabular}} &
  \textbf{\begin{tabular}[r]{@{}r@{}}Perplexity \\ Threshold\end{tabular}} &
  \textbf{\begin{tabular}[r]{@{}r@{}}Total \\ Docs\end{tabular}} &
  \textbf{\begin{tabular}[r]{@{}r@{}}Chosen \\ Docs\end{tabular}} &
  \textbf{\begin{tabular}[r]{@{}r@{}}Filtering \\ Rate\end{tabular}} \\ \midrule
asm &
  {27.4} &
  {65155.6} &
  {1013.9} &
  {1216} &
  {25,617} &
  {18,713}&
  {26.9\%}\\
ben &
  {6.7} &
  {22941.5} &
  {286.6} &
  {606.7} &
  {6,838,196} &
  {6,274,727} &
  {8.24\%}\\
guj &
  {7.8} &
  {23184.4} &
  {421.7} &
  {792.5 } &
  {640,843} &
  {586,977} &
  {8.4\%}\\
hin &
  {5.7} &
  {160264.7} &
  {230.44} &
  {378.8} &
  {19,362,407} &
  {17,271,194} &
  {10.8\%}\\
kan &
  {8.6} &
  {25413.1} &
  {74.5} &
  {103.4} &
  {748,914} &
  {623,662} &
  {9.1\%}\\
mal &
  {5.6} &
  {43419.9} &
  {65.8} &
  {61.4} &
  {1,723,524} &
  {1,012,425} &
  {41.25\%}\\
mar &
  {8.3}  &
  {16032.2} &
  {214.2} &
  {277.8} &
  {1,322,324} &
  {1,051,722} &
  {20.4\%}\\
nep &
  {7.1} &
  {20334.8} &
  {140.0} &
  {120.32} &
  {1,625,754} &
  {961,637} &
  {40.84\%}\\
ori &
  {5.8} &
  {166,311} &
  {160.0} &
  {170.8} &
  {61,692} &
  {44,298} &
  {28.1\%}\\
pan &
  {8.0} &
  {23375.0} &
  {232.6} &
  {229.7} &
  {302,421} &
  {195,115} &
  {35.48\%}\\
san &
  {32.8} &
  {5919.0} &
  {823.8} &
  {1397.7} &
  {3,332} &
  {2,993} &
  {10.17\%}\\
tam &
  {6.2} &
  {22583.3} &
  {157.6} &
  {262.3} &
  {2,416,008} &
  {2,089,674} &
  {13.5\%}\\
tel &
  {12.6} &
  {65297.8} &
  {139.3} &
  {377} &
  {930,407} &
  {898,991} &
  {3.37\%}\\
urd &
  {2.4} &
  {25206.5} &
  {158.4} &
  {316.8} &
  {1,502,769} &
  {1,372,703} &
  {8.65\%}\\
  \bottomrule
\end{tabular}
    \caption{Perplexity Statistics of \culturax{} and \madlad{} datasets. Perplexity is calculated using n-gram language models trained on data sampled from \datasc{}.}
    \label{tab:perplexity_stats}
\end{table*}

We clean the entire \culturax{} and \madlad{} corpora using \setu{} and deduplicate it with the entire \datasc{} split. Finally, we calculate the perplexities of each document and filter out those that are above the chosen threshold. Table \ref{tab:perplexity_stats} shows the final number of documents chosen after perplexity based filtering.

\section{Setu: A Comprehensive Pipeline for Data Synthesis, Cleaning, Filtering, and Deduplication}
\label{setu}

\begin{figure*}
    \centering
    \includegraphics[width=\linewidth]{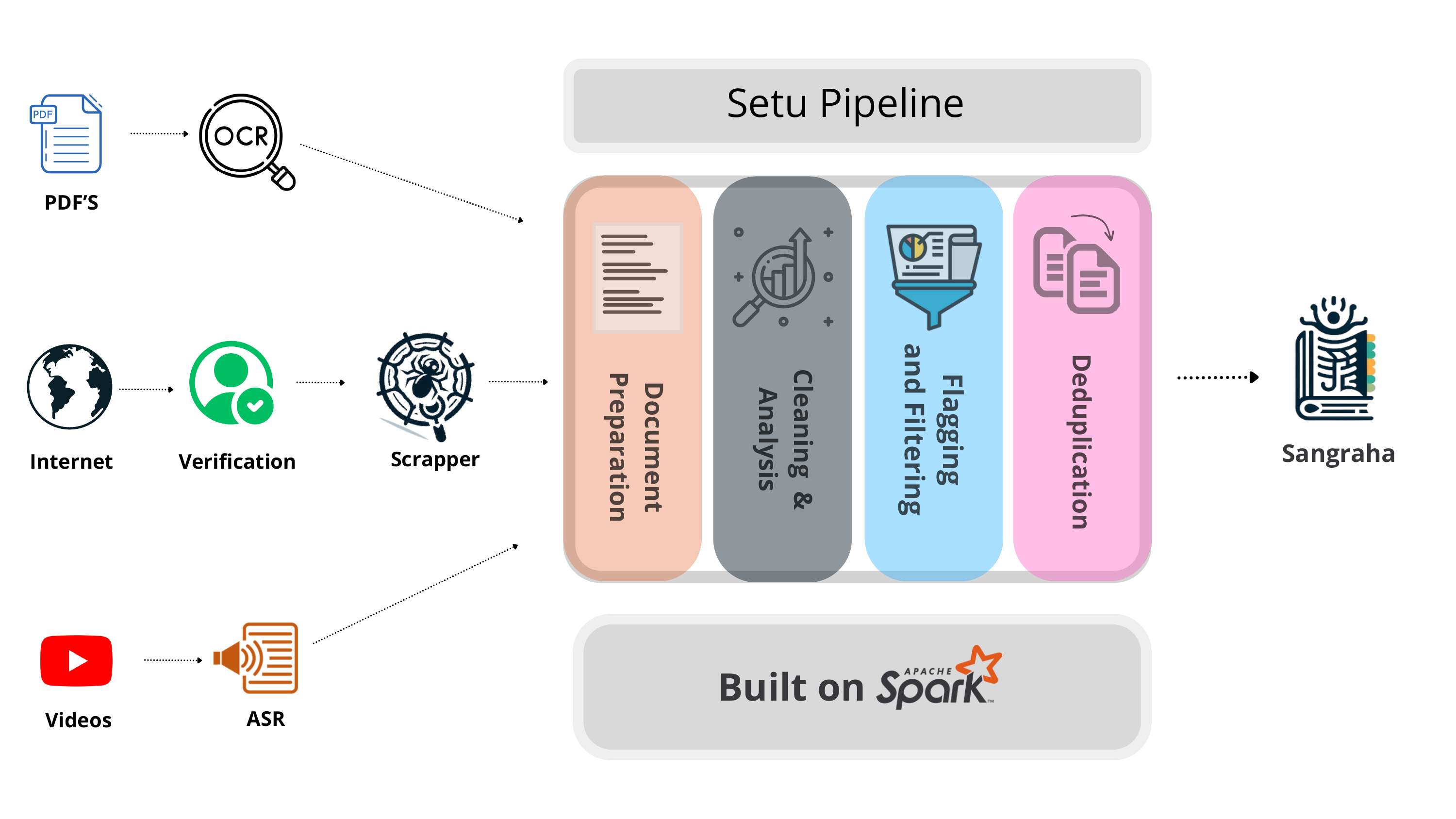}
    \caption{Overview of \setu{}, the data cleaning pipeline used for curating the \datasc{} corpus}
    \label{fig:sangraha_info}
\end{figure*}

To clean, filter, and deduplicate Web, PDF, and Speech data, we create \setu{}, a pipeline built on Apache Spark~\citep{spark} which broadly has 4 stages, as shown in Figure \ref{fig:sangraha_info} - \textsc{Document Preparation}, \textsc{Cleaning and Analysis}, \textsc{Flagging and Filtering}, and \textsc{Deduplication}. The \textsc{Document Preparation} stage focuses on extracting the text from our diverse sources and creating text documents for further processing. The \textsc{Cleaning and Analysis} stage performs cleaning on each document to reduce the noise, performs Language Identification by using an ensemble of different models, and then computes various statistical signals for each document. In the \textsc{Flagging and Filtering} stage, we apply various filters based on previously computed signals to filter out the noisy documents. Finally, the \textsc{Deduplication} stage performs fuzzy deduplication using \textit{MinHashLSH}. We also introduce the \setutranslate{} and the \setutransliterate{} pipelines for performing large-scale structure-preserving translations and transliterations of documents and conversations. We discuss the details of these pipelines in this section.

\subsection{Document Preparation}
This stage focuses on the extraction of text from varied data sources, ensuring the retention of main content while eliminating extraneous information and then preparing the notion of a document that is preserved throughout the pipeline. Due to the different modalities of content, this stage is different for each of Web, PDF, and Speech data.
\subsubsection{Web Data}
\begin{figure*}
    \includegraphics[width=\linewidth]{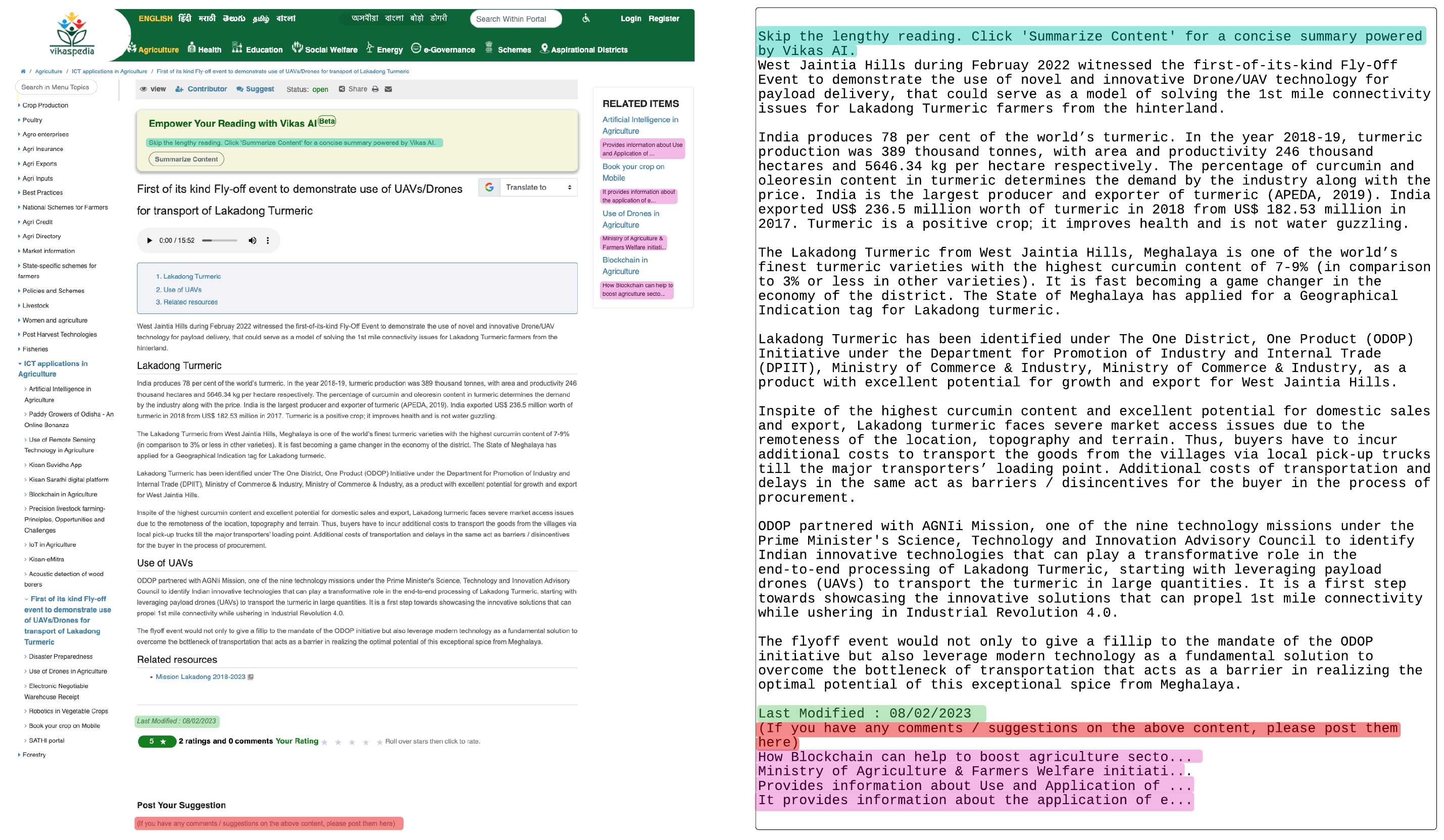}
    \caption{Example showing noisy content being extracted from the HTML using \textit{trafilatura}}
    \label{fig:trafil}
\end{figure*}

Preparation of the document for Web data is quite straightforward. We use \textit{trafilatura}~\citep{DBLP:conf/acl/Barbaresi21} to extract the text from the HTML pages that are scraped by \textit{webcorpus} scraper. Although \textit{trafilatura} is reportedly the best non-commercial library \citep{scrapinghub_article}, we still notice a considerable amount of noise in the outputs, specifically in dynamic webpages. Figure \ref{fig:trafil} shows an example of noisy content extracted using \textit{trafilatura}. In Web data, each webpage after text extraction is considered as a document.

\subsubsection{PDF Data}

Text Extraction from the OCR outputs from PDFs is not as straightforward as extracting text from a webpage. When utilizing Google Vision OCR for extracting text from PDF documents, the output is a structured JSON file that contains detailed information about the detected text. This information is organized hierarchically from larger text blocks down to individual characters. This hierarchical structure allows for a nuanced understanding of the document's layout and content. Broadly the bounding boxes are organized in the following hierarchy - Block, Paragraph, Word, Character. 

A block is the highest level of structure and is a container for paragraphs grouped to reflect their spatial relationships. Paragraphs are subdivisions of blocks and represent cohesive units of text, typically separated from other units by new lines or indentation. Words are the basic units of text and meaning within a paragraph. Each word is identified and extracted as a separate entity in the OCR output. Characters are the most granular level of text extraction, representing individual letters, numbers, punctuation marks, and other textual symbols.

\begin{figure*}
    \centering
    \subfloat[\centering \textbf{Bounding Box Suppression}: Page in which smaller bounding boxes are suppressed as these can lead to false flagging of pages or misaligned text.]{{\includegraphics[width=0.4\linewidth]{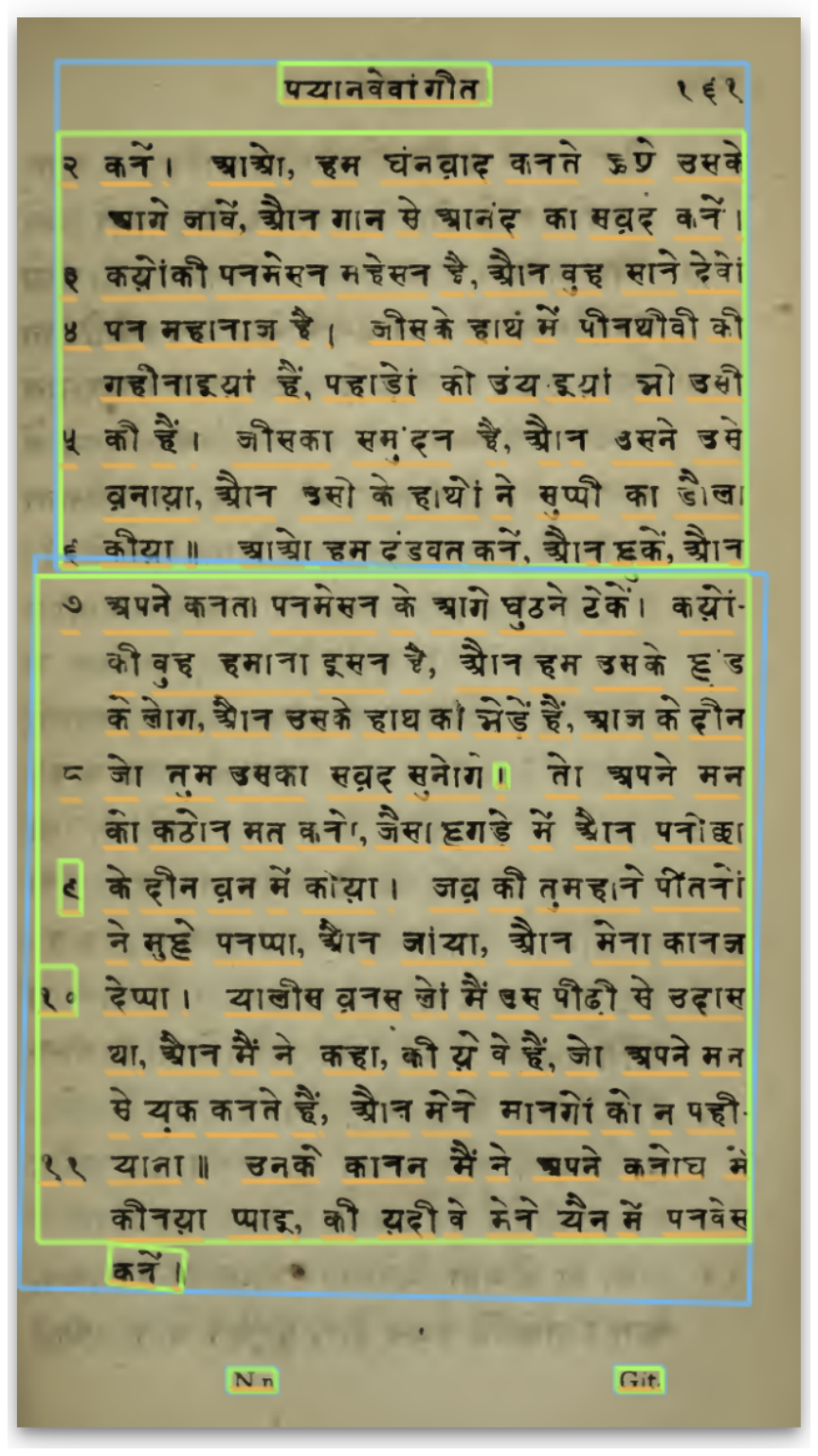}} \label{image:bbox_suppression}}
    \qquad
    \subfloat[\centering \textbf{Horizontally Sparse}: Page filtered out due to less horizontal text coverage, this can be indicative of very small lines, lists, index etc.]{{\includegraphics[width=0.4\linewidth]{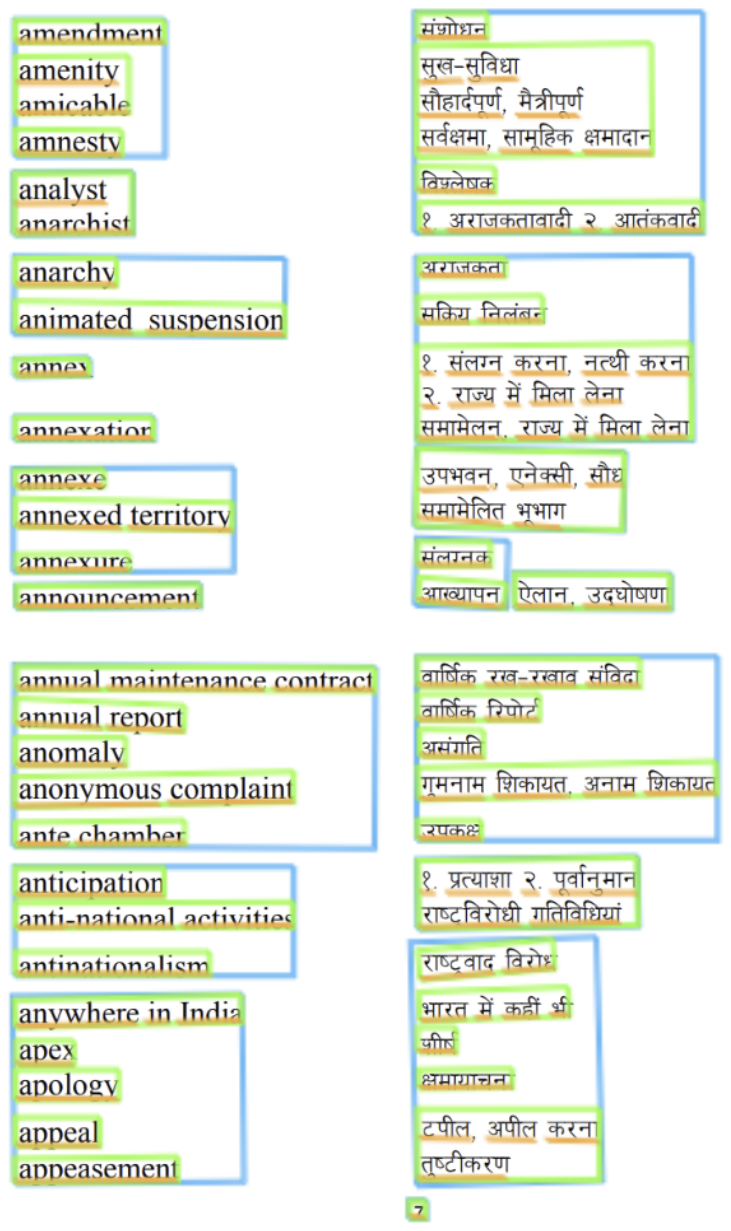}} \label{image:horizonal_sparse}}
    \qquad
    \subfloat[\centering \textbf{Vertically Sparse}: Page filtered out due to less vertical text coverage. This can be indicative of title pages, comics, etc.]{{\includegraphics[width=0.4\linewidth]{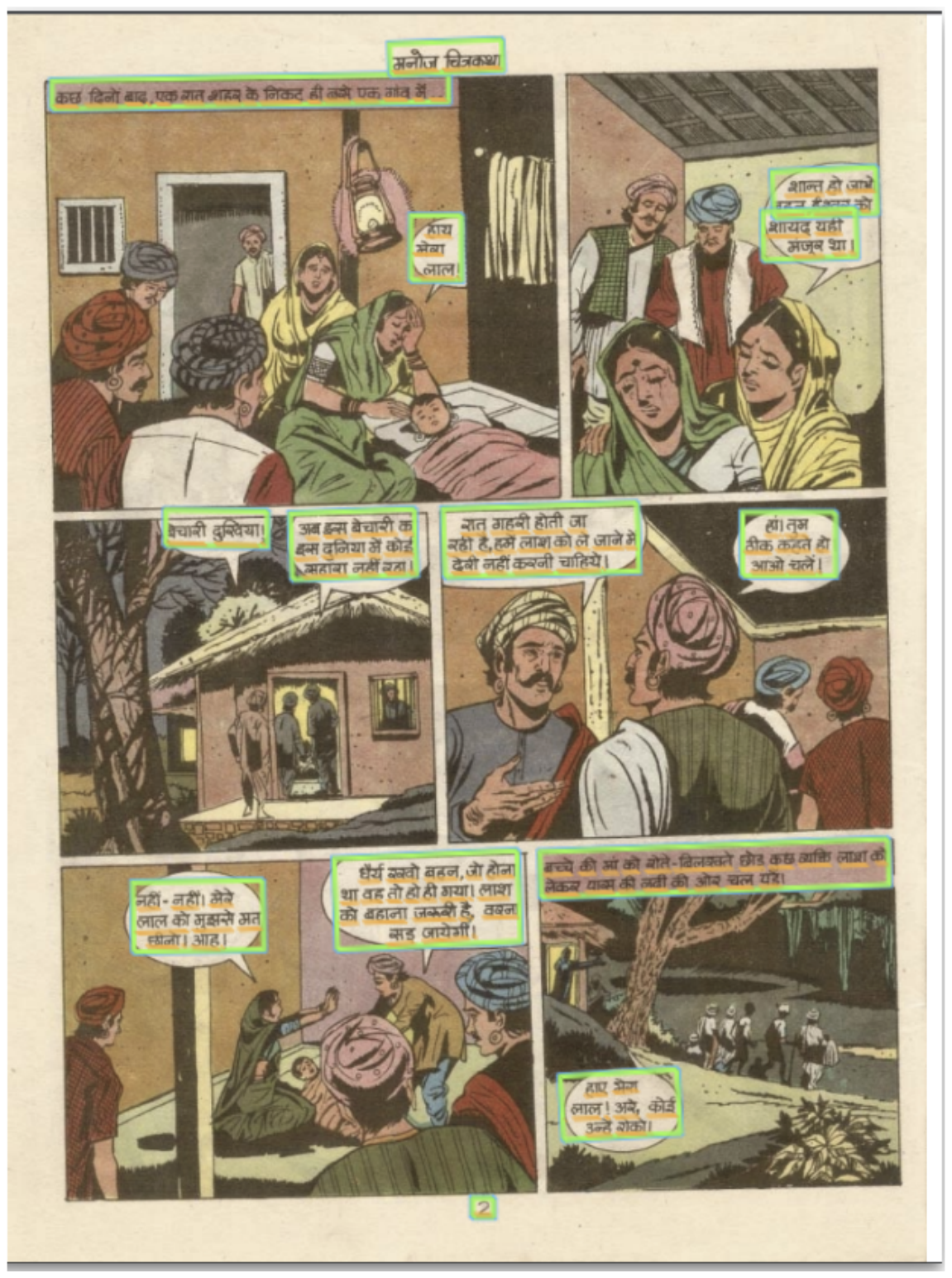}} \label{image:vertical_sparse}}
    \qquad
    \subfloat[\centering \textbf{High Bounding Box Overlap}: Page filtered out due to high bounding box overlap. This high overlapping can lead to disordered parsing of text, break in continuity, etc.]{{\includegraphics[width=0.4\linewidth]{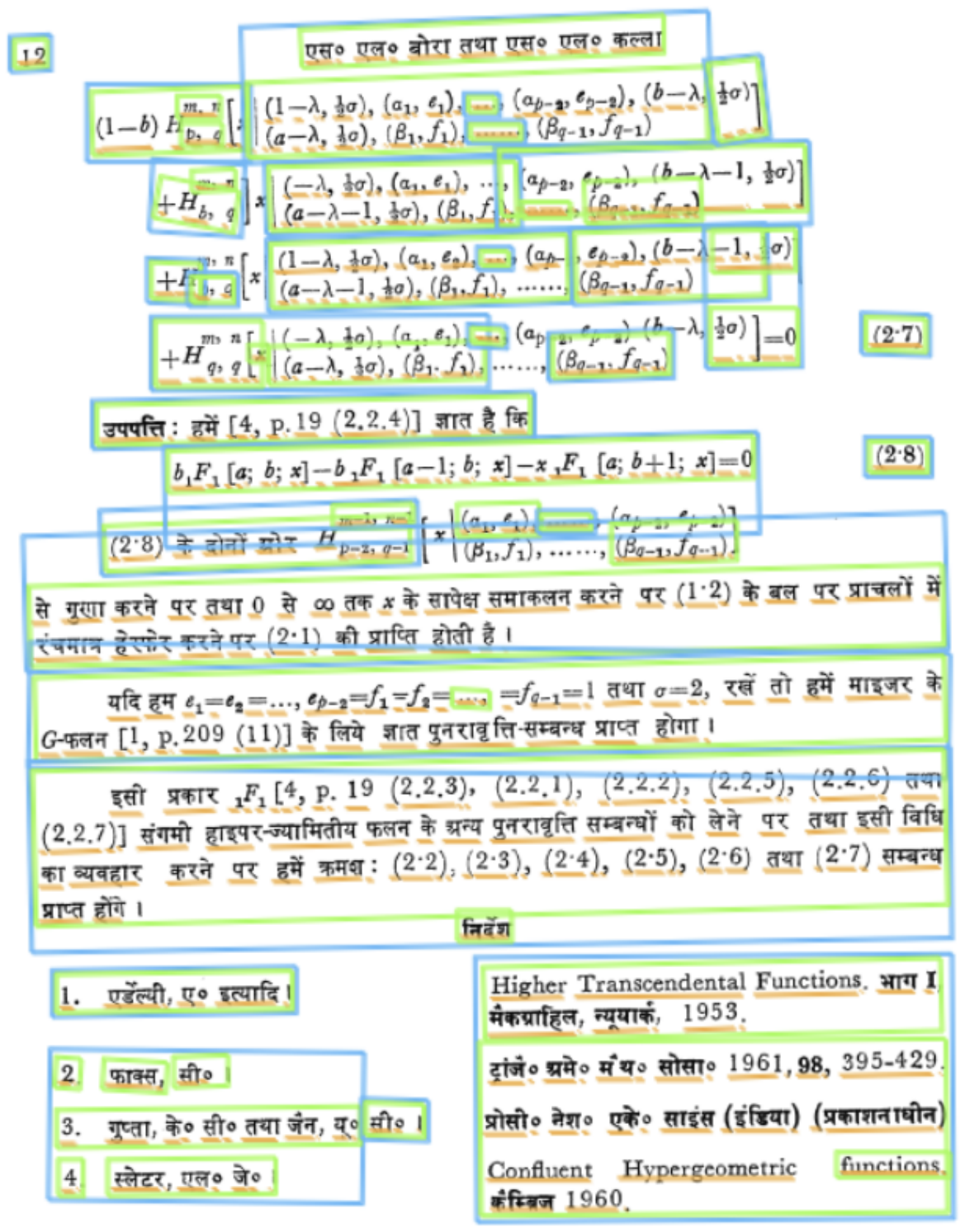}} \label{image:high_overlap}}
    \qquad
    \caption{Illustrative examples of pages flagged in various bounding box filters.}
    \label{image:pdf_filters}
\end{figure*}

Each category contains information such as the bounding box coordinates, confidence scores, language scores, and the text identified in that box. We observe that directly consuming the text from the OCR is not good as it contains a lot of noise coming in due to incorrect layout parsing. We also observed that due to the skewness and quality of images, we had multiple instances where we had bounding box overlaps, bounding box mismatch/misalignment, text overlaps, and language script mismatches. To resolve these and extract the highest quality text, we develop bounding-box based filters. We list the filters below:
\begin{itemize}
    \item \textbf{Bounding Box Suppression}: Here, we perform bounding box suppression, where we try to suppress the smaller bounding boxes that overlap with larger bounding boxes. For each pair of overlapping bounding boxes, we calculate the ratio of the area of intersection over the area of the smaller bounding box. We suppress the smaller bounding box if this ratio exceeds a chosen threshold. Figure \ref{image:bbox_suppression} shows an example of a page where bounding box suppression is applied.
    \item \textbf{Removing Horizontally sparse pages}: Here, we identify and remove pages that exhibit a significant lack of content across the horizontal span of the page. If a page has large horizontal gaps with little to no content—indicating that the text or visual elements are spread thinly across the width of the page—it is considered horizontally sparse. Such pages are often less informative or relevant, like index pages and table of contents among others. Figure \ref{image:horizonal_sparse} shows an example of a page flagged as horizontally sparse.
    \item \textbf{Removing Vertically sparse pages}: Similarly, we also remove pages with insufficient content along the vertical axis. Pages containing large vertical gaps, such as excessive spacing between paragraphs or sections without meaningful content, are deemed vertically sparse. These pages are also less informative, like pages having publisher information, colophons, comic strips, etc. Figure \ref{image:vertical_sparse} shows an example of a page flagged as vertically sparse.
    \item \textbf{Removing pages with high overlapping Bounding Boxes}: Here, we remove the pages having a very high bounding box overlap percentage, i.e., greater than a chosen threshold as shown in Figure \ref{image:high_overlap}.
    \item \textbf{Removing Sparse blocked pages}: Here, we remove the pages having very sparse bounding boxes. A block bounding box is considered sparse if the difference between the total area of the block bounding box and the total area of paragraph bounding boxes enclosed in it is greater than a chosen threshold. By this, we remove pages with tables, large images, and forms among others.
    \item \textbf{Removing pages with low script confidence}: Here, we compute each paragraph's average script confidence score on a given page. Paragraphs with scores below our confidence thresholds are flagged for potential exclusion. Subsequently, the entire page is discarded if the number of flagged paragraphs exceeds an allowable limit. This ensures a balance between rejecting poor-quality OCR output and retaining usable content.
\end{itemize}

After filtering, we merge the final text extracted from the pages to form documents. To maintain textual continuity as well as to get as many long-form documents as possible, we concatenate the text of only consecutive batches of pages of a given PDF together. Table \ref{tab:average_page_count_after_merge} shows the average number of pages per language that are merged to form a document.

\subsubsection{Speech Data}
For speech data, we currently only handle data from SRT files and the transcripts obtained using Automatic Speech Recognition (ASR) models. For SRT files, we first extract the text by removing the timestamps. We further define simple regex patterns to filter out other noisy content like character cues, continuation ellipses, etc. A single processed SRT file is considered as a document.

The automated transcriptions obtained using ASR models are first pieced together to form larger chunks of text. We then use \textit{IndicPunct}~\citep{gupta2022indicpunct} punctuation models to add appropriate punctuation marks in the transcripts. Finally, all chunks belonging to a single video are merged to form a document.

\subsection{Cleaning and Analysis}
This stage primarily focuses on performing in-document cleaning and language identification. Additionally, we compute various statistics for performing analysis and further filtering. We divide this stage into three sub-stages - Document Cleaning, Language Identification, and Analysis.
\subsubsection{Document Cleaning}
\begin{figure*}[t]
    \centering
    \includegraphics[width=\linewidth]{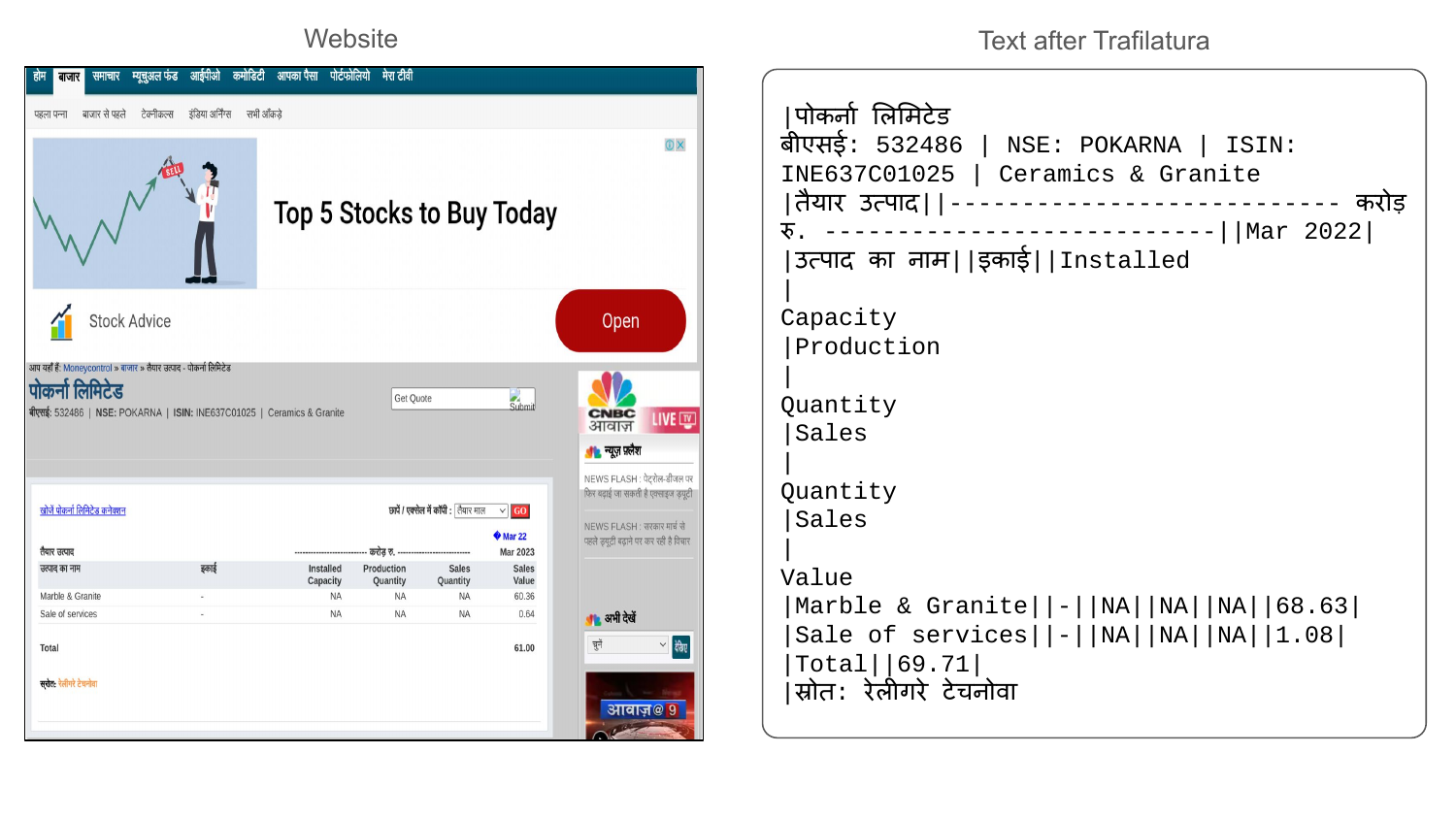}
    \caption{Document flagged by symbol heavy filter in the Cleaning and Analysis stage.}
    \label{image:symbol_heavy_filter}
\end{figure*}

\begin{figure*}[t]
    \centering
    \includegraphics[width=\linewidth, height=10cm]{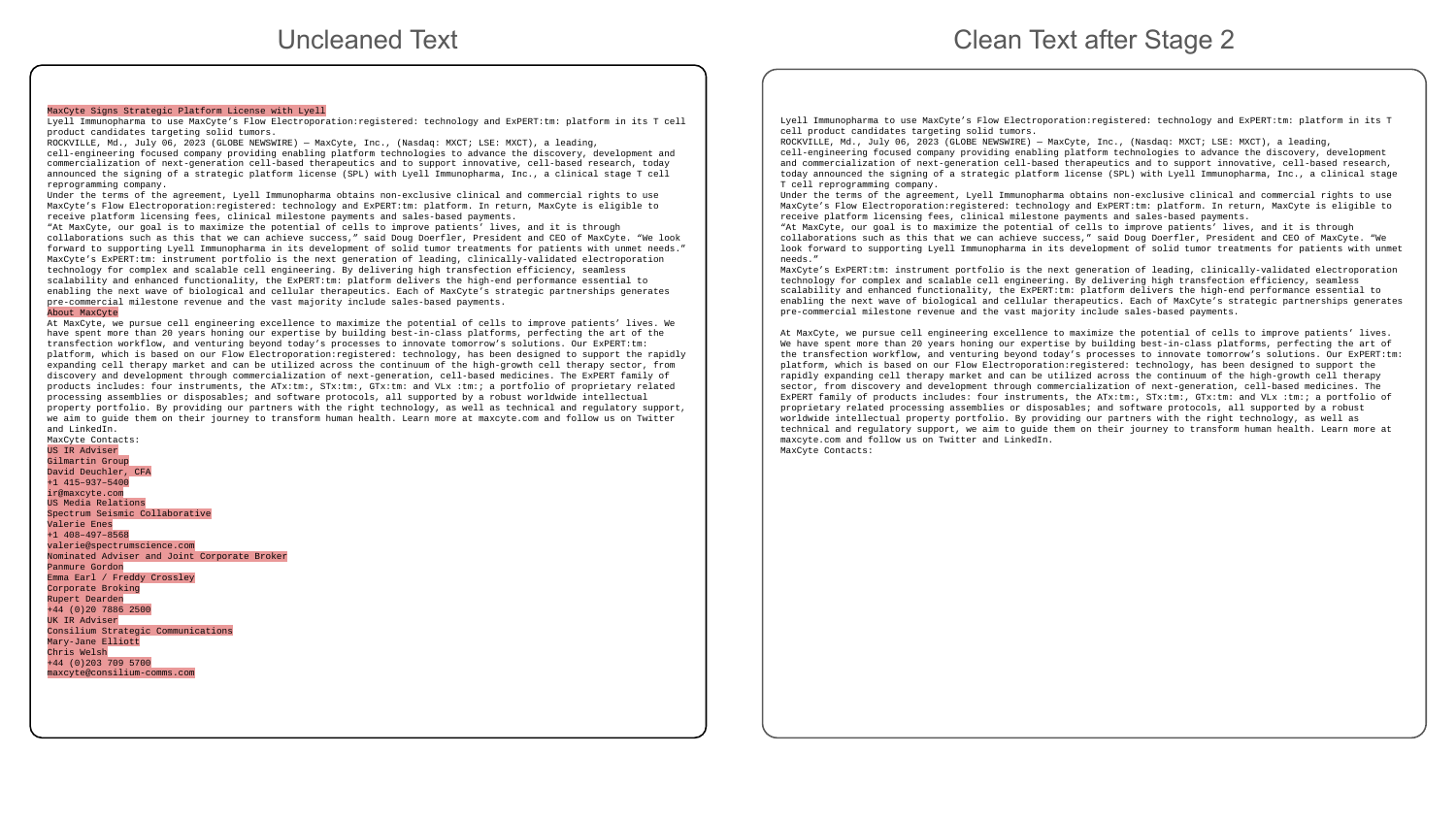}
    \caption{Cleaning performed by `terminal punctuation filter' in Cleaning and Analysis stage.}
    \label{image:terminal_punctuation_filter_example}
\end{figure*}

Although \textit{trafilatura} and GCP Vision OCR are reportedly the best \citep{scrapinghub_article, ocr-perf}, we still need to mitigate the errors that creep in. We define the below filters that clean a document.
\begin{itemize}
    \item \textbf{Code Span Removal}: This filter is applied exclusively for Web Crawls where we define regex patterns to detect and remove code spans like improperly rendered HTML or JavaScript code.
    \item \textbf{Symbol Heavy Filter}: Documents with a high ratio of invalid characters (e.g., punctuation, emojis, and other symbols) to total characters exceeding a predefined threshold are discarded. Figure \ref{image:symbol_heavy_filter} shows an example of a symbol-heavy document.
    \item \textbf{Terminal Punctuation Filter}: This filter is again exclusively for web crawls, it removes text segments lacking valid terminal punctuation, effectively filtering out clickbait text, menus, and incomplete sentences. Figure \ref{image:terminal_punctuation_filter_example} shows an example of content removed using this filter.
    \item \textbf{Symbol Only Chunk Filter}: This filter removes all the text chunks with only numbers or symbols. 
    \item \textbf{Repeated Chunk filter}: Applied to PDFs to eliminate repeated text chunks, targeting redundant headers and titles.
    \item \textbf{Chunk length filter}: Specific to PDFs, it removes chunks with a word count below a set threshold.
\end{itemize}


\subsubsection{Language Identification}
To address the issues of accuracy that may occur while relying on a singular model highlighted in Appendix \ref{appendix: lid}, we use an ensemble approach using three LID models - \indiclid~\citep{madhani-etal-2023-bhasa}, \cld{}\footnote{\url{https://github.com/google/cld3}}, \nllb~\citep{nllb}. Notably, \indiclid{}, which is specifically trained for Indic languages, is assigned a preferential weighting in our ensemble framework. However, if both \cld{} and \nllb{} agree on a different language and are very confident about it (beyond a chosen threshold), we consider their prediction instead. This methodology aims to leverage the specialized capabilities of \indiclid{} for Indic languages while still incorporating the complementary strengths of \cld{} and \nllb{} in other languages.

\subsubsection{Document Analysis}
We compute various document-specific statistics for performing subsequent filtering. The metrics and their descriptions are outlined in Table \ref{tab:extracted_metrics}.
\begin{table*}[]
\centering
\begin{tabular}{lp{8cm}}
\toprule
\textbf{Metrics}                                 & \textbf{Description}                                                      \\
\midrule
\textit{bytes}                                   & size of the document interms of bytes,                           \\ \\
\textit{word\_count}                             & no.of words present in a document                                \\ \\
\textit{char\_count}                             & no.of characters present in a document                           \\ \\
\textit{lines\_count}                            & total no.of sentences present in a document                      \\ \\
\textit{mean\_line\_length}                      & mean sentence length interms of words of a document              \\ \\
\textit{min\_line\_length}                       & minimum sentence length interms of words of a document.          \\ \\
\textit{max\_line\_length}                       & max sentence length interms of words of a document               \\ \\
\textit{nsfw\_words\_count}                      & no.of NSFW words present in a document                           \\ \\
\textit{non\_li\_character\_count}               & no.of non-latin/non-indic characters in a document               \\ \\
\textit{10\_gram\_characters\_repetition\_score} & score used for filtering documents using 10-gram character repetition filter\\ \\
\textit{5\_gram\_words\_repetition\_score}       & score used for filtering documents using 5-gram word repetition filter \\ \\
\bottomrule
\end{tabular}
\caption{Showing all the metrics that are calculated in analysis stage}
\label{tab:extracted_metrics}
\end{table*}

\subsection{Flagging and Filtering}
Following the analysis, the documents are filtered based on predefined language-specific thresholds for the computed statistics. This step is essential to eliminate residual noise that might have survived the initial cleaning process. We include filters inspired from various previous works like \roots~\citep{laurençon2023bigscience}, \textsc{Gopher}~\citep{rae2022scaling} and \textsc{C4}~\citep{t5} among a few. 

\begin{itemize}
    \item \textbf{NSFW word ratio filter}: In an effort to reduce corpus toxicity, documents with a high ratio of NSFW (Not Safe For Work) words to total words are excluded. This approach aligns with that of \icvii{}, involving the development of an NSFW word list specifically tailored for Indic languages. This list is made available to the research community to encourage further studies.
    \item \textbf{Non Latin/Indic character ratio filter}:  Documents characterized by a significant ratio of non-Latin/Indic characters are removed. This filter eliminates content erroneously classified as Indic by the Language Identification (LID) stage. Figure \ref{image:non-li_filter} shows an example of the type of content removed by this filter.
    \item \textbf{Line count filter}: Documents with an exceedingly low number of lines are discarded to remove potentially irrelevant or insufficient content.
    \item \textbf{Minimum mean line length filter}: This filter targets documents with short average line lengths, effectively removing index pages and similar content deemed unsuitable for the corpus.
    \item \textbf{5-gram word repetition}: Inspired from \roots{}, we create a filter for the repetitions by looking at the occurrences of the 5-gram word sequences. We define the word repetition ratio as the ratio of the sum of the occurrences greater than or equal to the sum of all occurrences, and we discard documents with too high a ratio. 
    \item \textbf{10-gram character repetition}: Similar to the word repetition filter, this criterion focuses on 10-gram character sequences. Documents exhibiting a high ratio of such repetitions are excluded, based on methodology inspired by \roots{}.
\end{itemize}

\begin{figure*}[t]
    \includegraphics[width=\linewidth]{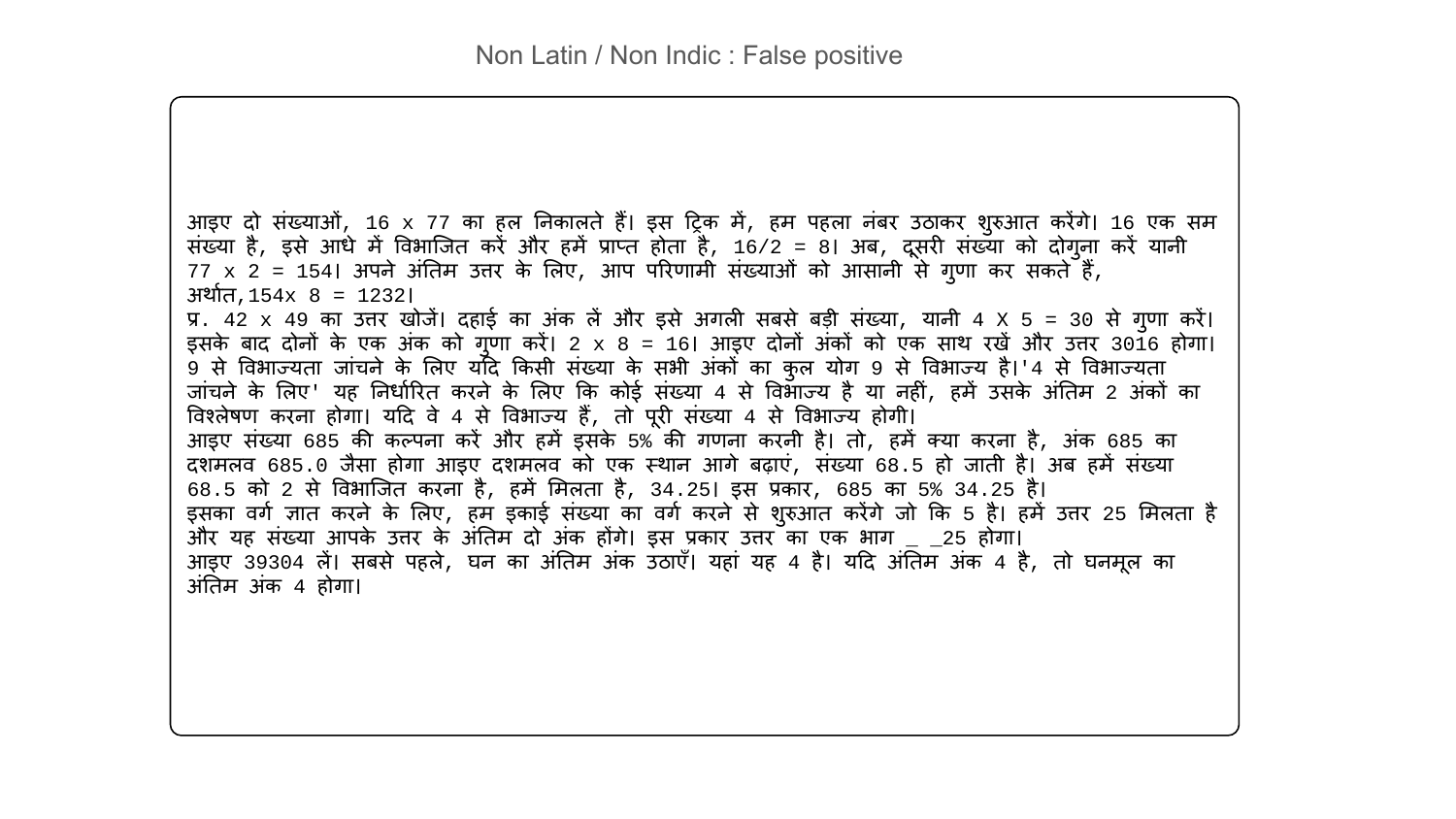}
    \caption{Figure showing the type of content flagged by the Non Latin/Indic Filter. Ideally, these types of documents should not be rejected since they contain valid math characters. These are some of the limitations of our current pipeline.}
    \label{image:non-li_filter}
    \includegraphics[width=\linewidth]{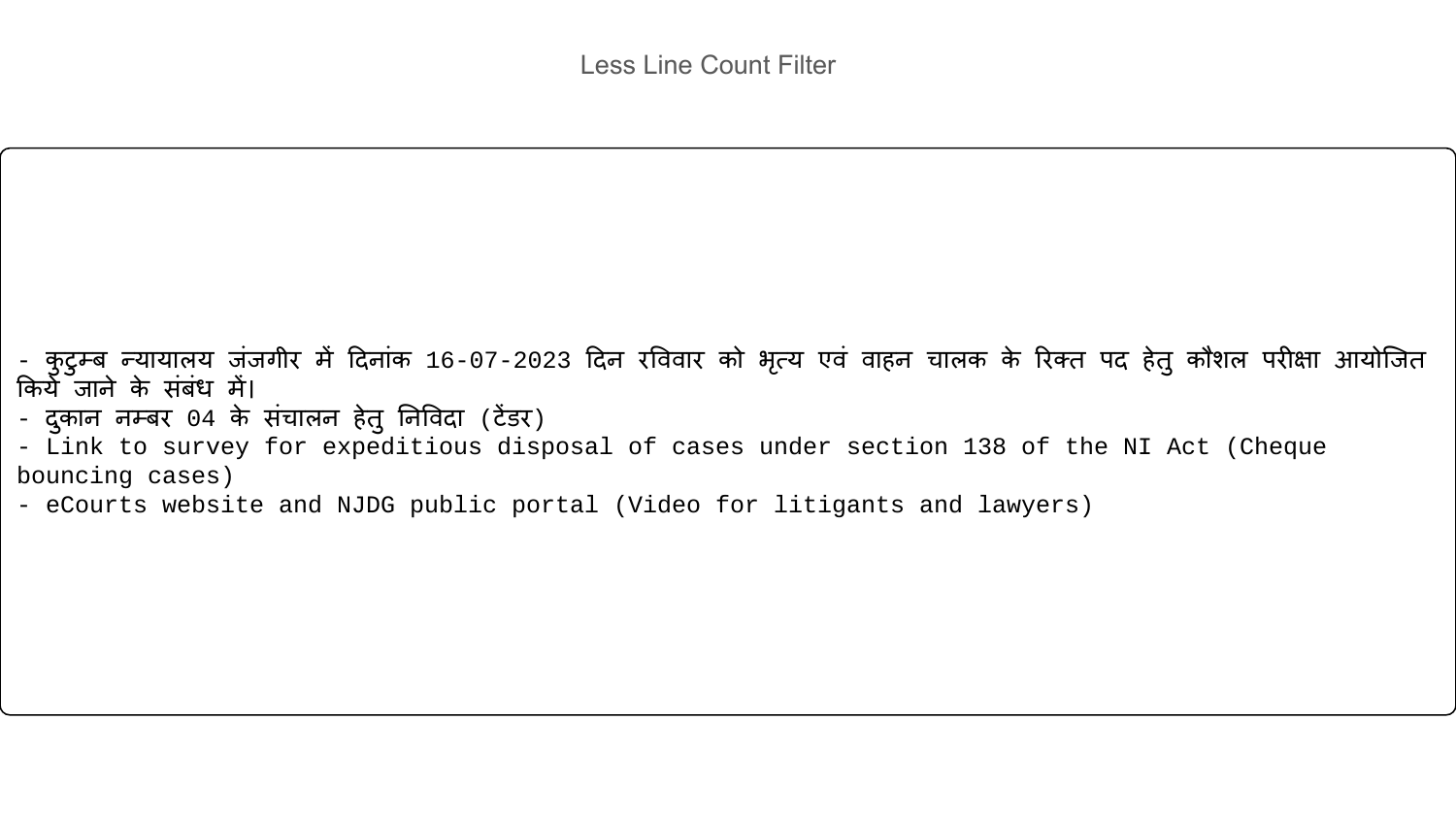}
    \caption{Figure showing the type of content flagged by the Line count filter. }
    \label{image:non-li_filter}
\end{figure*}


\subsection{Deduplication}
The concluding stage of Setu addresses the critical task of deduplication using fuzzy deduplication. Following \culturax{}, we use the Python implementation of MinHashLSH from the \textit{text-dedup}\footnote{\url{https://github.com/ChenghaoMou/text-dedup/tree/main}} repository. We efficiently identify and remove duplicate documents within the corpus by utilizing 5-grams and a similarity threshold of 0.7, based on Jaccard similarity. This procedure is executed separately for each language, utilizing a computing node with 256 CPUs.

\subsection{Setu-Translate}
\label{setu-translate}
Majority of the machine translation systems are trained as sentence-level translators, which often struggle to preserve various entities like inter-sentence separators, new-line characters, tab spaces, markdowns, bullet points, etc. Simple sentence-tokenizers present in the packages like NLTK \citep{loper2002nltk} and IndicNLP Library \citep{indicnlplib} are not capable of retaining these inter-sentence separators and markdowns. We introduce \setutranslate{}, a robust translation pipeline for mass-translation of both pre-training and Instruction fine-tuning data while preserving the structure of the document and the conversation. Overall, \setutranslate{} focuses on accurately identifying the parts of the document that must be sent to the translation model and then replacing the translated sentences in the overall document, thereby preserving the overall structure of the translated document. The three main stages of \setutranslate{} are described in this section.

\textbf{Templating}

Using regex patterns, we identify the parts of the documents we intend to translate. The goal of this stage is to preserve the structure of the document. The regex patterns defined ignore markdown structures, code snippets (enclosed in backticks), bullet points, paragraph indicators, Roman numerals, etc., and extract only the sentences. After performing unicode-normalization and deduplication on the extracted sentences, a global sentence-level dataset is created.

\textbf{Inference}

We binarize the data first and then utilize \translation{}~\citep{gala2023indictrans2} for translating English into Indic languages. We leverage both GPUs and TPUs for large-scale translation. To benefit the community, we open-source the flax port for \translation{} model for TPU inferencing.

\textbf{Replace}

Once we have the translated sentences, we perform a regex-based replacement of the original sentences with the translated ones. This ensures that only sentences are replaced and the other structure of the document is retained as is.

\subsection{Setu-Transliterate}
Similar to translation, we also release the Setu-Transliterate pipeline. Since transliteration is done at a word level and doesn't consider the context of the remaining words, we follow the regular word replacement strategy. We maintain a continuously updating mapping of Indic words to their Roman counterparts in a prefix-based hierarchical format, which we feel is the key to speedup and rapid access to the required word pairs. 

\textbf{Word Mapping Dictionary}

For the creation of the initial mapping, we use \aksharantar{}\citep{madhani-etal-2023-aksharantar} dataset, which is the largest publicly available transliteration dataset for Indic languages, as the starting point. We convert \aksharantar{} into the said prefix-based hierarchical format. This mapping is continuously updated with new mappings as we discover new un-romanized words further in our pipeline.

\textbf{Word Replacement}

Word-level replacement has 2 main challenges: (i) identifying words to replace while preserving the entire document structure; and (ii) unordered replacement leading to sub-word replacement instead of the entire word. We address (i) using the same regex-based approach used in \setutranslate{}. To address (ii), we sort the mapping based on source-language word length in descending order before feeding the mapping to the regex-based `replace' module.

\textbf{Inference}

During the first `replace' pass, we log the un-romanized words, the words whose mapping is not available in the current word mapping dictionary. In the `inference' stage, we transliterate these words using \xlit{}\citep{madhani-etal-2023-aksharantar} to get an updated word-mapping dictionary. We then repeat the word replacement until all the words are properly romanized.
\section{IndicAlign}

In this section, we describe the composition and the curation process of \indicalign{}, comprising of around 74.7 million diverse, human, and synthetic prompt response pairs. Majority of the high-quality synthetic supervised fine-tuning data released has been created with proprietary models like ChatGPT and GPT-4, which renders them unusable in commercial settings. We therefore consider only license-friendly datasets and models for curating \indicalign{} for different Indian languages. Further, we use the \setutranslate{} and \setutransliterate{} pipelines discussed in Section \ref{setu} for translating and transliterating the conversations, thereby maintaining the structure and the markdown of the responses. 
\indicalign{} comprises two distinct splits: \indicaligni{} and \indicalignt{} as shown in Table \ref{tab:indicalign_stats}. 

\begingroup
\setlength{\tabcolsep}{4pt} 
\renewcommand{\arraystretch}{1} 
\begin{table*}[]
    \centering
    \scriptsize
    \begin{tabular}{@{}l|rrr|rrrrrr@{}}
    \toprule
        \textbf{Component} &
          \textbf{\begin{tabular}[c]{@{}r@{}}Prompt \\ source\end{tabular}} &
          \textbf{\begin{tabular}[c]{@{}r@{}}Response \\ source\end{tabular}} &
          \textbf{\begin{tabular}[c]{@{}r@{}}Original/\\ Translated\end{tabular}} &
          \multicolumn{1}{l}{\textbf{\#Examples}} &
          \textbf{\begin{tabular}[c]{@{}r@{}}Avg. \\ Turns\end{tabular}} &
          \textbf{\begin{tabular}[c]{@{}r@{}}Avg. \\ Inst. Len\end{tabular}} &
          \textbf{\begin{tabular}[c]{@{}r@{}}Avg. \\ Out. Len\end{tabular}} &
          \multicolumn{1}{l}{\textbf{\#Lang.}} &
          {\textbf{\begin{tabular}[r]{@{}r@{}}Lexical \\ Diversity\end{tabular}}} \\
          \midrule
        Indic ShareLlama & H & M & T & 21.1k    & 1    & 60.45 & 267.98 & 15 & 57.69 \\
        Dolly-T          & H & H & T & 15.0k    & 1    & 12.34 & 59.38  & 15 & 47.23 \\
        OpenAssistant-T  & H & H & T & 19.9k    & 2.98 & 25.72 & 136.37 & 15 & 59.75 \\
        WikiHow          & H & H & T & 20.3k    & 1    & 43.85 & 327.95 & 15  & 23.87 \\
        IndoWordNet      & H & H & O & 74,272.2k & 1    & 19.74 & 14.84  & 18 & 37.24 \\
        Anudesh          & H & M & T & 43.3k    & 1.58 & 12.4  & 149.28 & 20 & 51.69 \\
        Wiki-Conv        & M & M & T & 144k   & 9.14 & 7.09  & 11.22  & 15 & 23.17 \\
        Wiki-Chat        & M & M & T & 202k   & 2.8  & 23    & 227.75 & 15 & 56.67 \\
        \midrule
        HH-RLHF-T          & H & M & T & 32.6k    & 1    & 14.11 & 64.88  & 15 & 79    \\
        Toxic Matrix      & M & M & T & 90.3k    & 1    & 33.68 & 89.64  & 15 & 86.57 \\
    \bottomrule
        \end{tabular}
    \caption{Overall statistics of \indicalign{}. Dolly-T represents Dolly Translated, OpenAssistant-T represents OpenAssistant Translated. Lexical Diversity is computed averaging the MTLD score over each utterance. Interaction lengths are reported in number of words.}
    \label{tab:indicalign_stats}
\end{table*}
\endgroup

\subsection{IndicAlign - Instruct}
The \indicaligni{} split encompasses datasets that can be used to imbibe instruction-following ability in Large Language Models. Firstly we amalgamate different existing Instruction Fine-tuning (IFT) datasets with prompts authored by humans and responses generated by humans or open, license-friendly models. To complement this human-centric approach, which is often too expensive and time-consuming, we turn to synthetic data generation using existing chat-aligned models following the works of \citet{ding2023enhancing}, \citet{li2023camel}, and \citet{xu2023wizardlm}. We ensure that our outputs are always from open, license-friendly models and are always grounded in context. 

\subsubsection{Indic-ShareLlama}
We collect conversations from ShareGPT\footnote{\url{https://huggingface.co/datasets/anon8231489123/ShareGPT_Vicuna_unfiltered}}, a platform where users share their interesting conversations with ChatGPT\footnote{\url{https://chat.openai.com/}}. We then collect the user prompts from the first turn of the conversations and prompt \llama{}~\citep{touvron2023llama} model for the responses in Indian contexts. We explicitly exclude all the non-English, coding, and math-related prompts to get around 21K conversations and then translate and transliterate them into 14 Indian languages to form \textsc{Indic-ShareLlama}.

\subsubsection{Dolly-Translated}
\textsc{Dolly-15K}~\citep{dolly}, introduced by Databricks, is an open-source conversation dataset aimed at democratizing the capabilities of LLMs. It consists of 15K high-quality, human generated prompt-response pairs, authored by around 5000 Databricks employees. Following \citet{gala2024airavata} and \citet{husain2024romansetu}, we translate and transliterate these conversations into 14 Indian languages to form \textsc{Dolly-Translated}.

\subsubsection{OpenAssistant-Translated}
OpenAssistant Conversations (\textsc{OASST1})~\citep{köpf2023openassistant}, is a collection of human-generated, human-annotated assistant style conversation corpus consisting of around 3K conversation trees and around 20K conversations. Extending \citet{gala2024airavata, husain2024romansetu}, we release the translated and transliterated versions in 14 Indian languages as \textsc{OpenAssistant-Translated}.

\subsubsection{WikiHow}
Wikihow\footnote{\url{https://www.wikihow.com/}} is a collaborative online wiki-style platform that serves as a valuable resource for a diverse array of how-to guides. It covers various aspects of life, including technology, arts, entertainment, home and garden, health, and more. Each piece is typically structured with step-by-step instructions, supplemented by illustrations and videos, to help readers achieve their goals. The questions users pose in these articles closely align with potential use cases for any model, making it a rich training resource. \citet{gala2024airavata} curate around 20,400 and 6000 instruction-answer pairs in English and Hindi. The data is formulated as a completion task given either a question or a question along with a few initial steps. We extend their efforts and translate and transliterate the English conversations into 14 Indian languages.

\subsubsection{IndoWordNet}
WordNets are a comprehensive lexical database originally designed for English~\citep{wordnet} and later extended to Indic Languages~\citep{Narayan2002HindiWordNet, bhattacharyya2010indowordnet}. It organizes words into sets of synonyms called synsets, providing short definitions and usage examples. Beyond mere dictionaries, WordNet also captures the various semantic relationships between words. We leverage this rich semantic information to create instruction fine-tuning data to teach the model grammar and language creativity.

\begin{figure*}[t]
    \includegraphics[trim={0 10cm 0 1.5cm},clip, width=\linewidth]{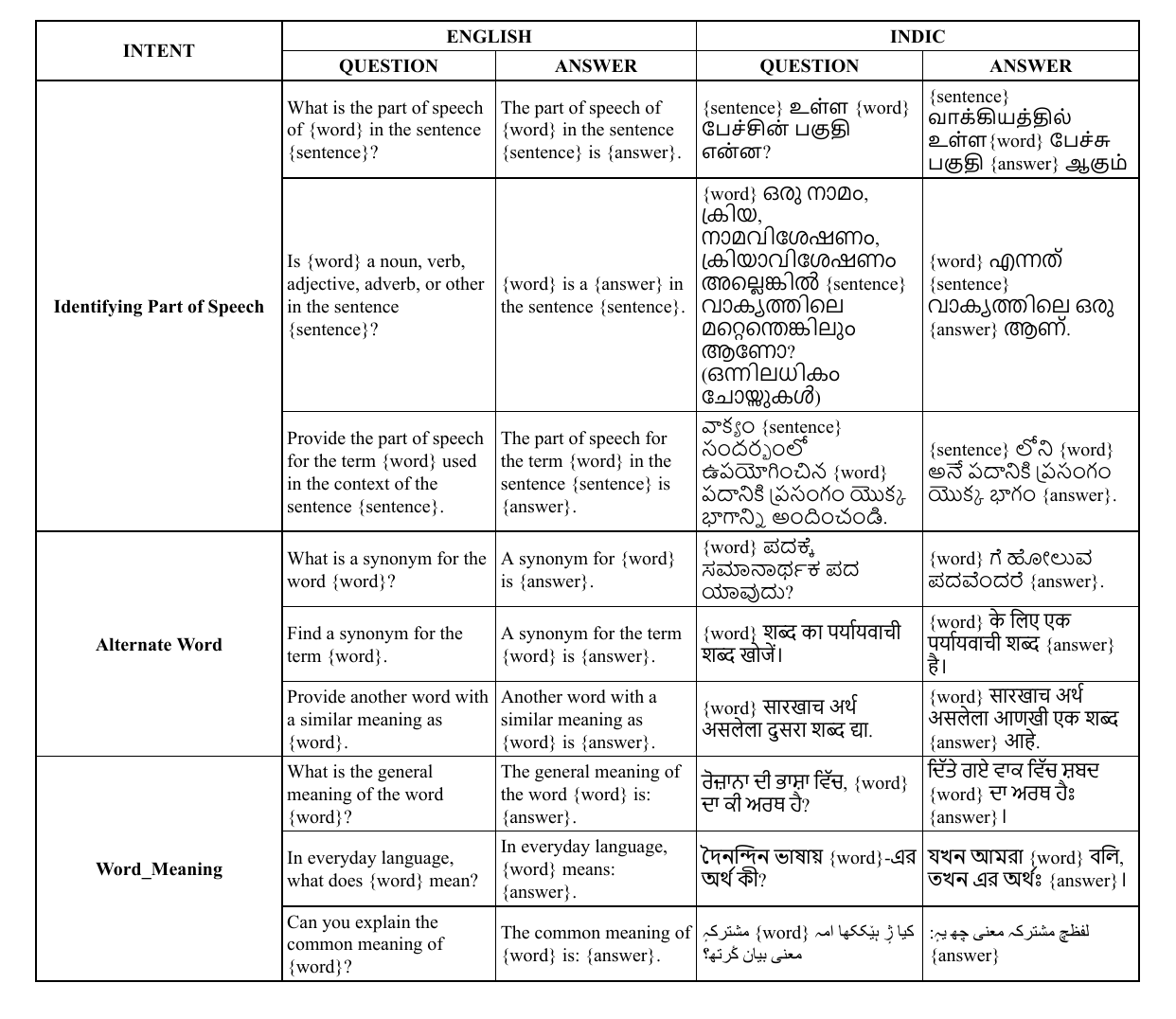}
    \caption{Example prompt templates for three sample intents in the IndoWordNet Instruction fine-tuning data.}
    \label{image:indowordnet_templates}
\end{figure*}

We first identify a list of 21 potential intents encompassing tasks such as Part of Speech identification, sentence construction, and synonym discovery. We craft 5 prompt-response templates for each intent, resulting in a repository of 105 distinct templates. Then we iterate through the lexicon in  IndoWordNet using \textit{pyiwn}~\citep{panjwani-etal-2018-pyiwn}, randomly sampling 100 templates for each word yielding around 74M pairs for 18 Indic languages. Figure \ref{image:indowordnet_templates} shows some examples of templates. Table \ref{tab:indowordnet} shows each language's final statistics of the prompt-answer pairs.

\subsubsection{Anudesh}
Here, we introduce a novel dataset of real user interactions with conversational models, leveraging open, license-compatible models such as \llama~\cite{touvron2023llama}. Recognizing the limitations imposed by OpenAI's terms of use\footnote{\url{https://openai.com/policies/terms-of-use}} on existing crowd-sourced model interaction datasets, such as \textsc{ShareGPT} and \textsc{WildChat}~\citep{zhao2024inthewildchat}, our dataset aims to provide a resource, free from such constraints, thereby facilitating broader applicability in training diverse conversational models.

We create \anudesh{} by asking the user to interact with the model while following an instruction displayed on the screen. Occasionally, we allow unrestricted interactions to collect more diverse and creative prompts. Each displayed instruction is based on three axes that guide the user - 
\begin{itemize}
    \item \textbf{Intent} - Defines the purpose and goal behind the interaction, such as summarization, recommendation seeking, etc.
    \item \textbf{Domain} - Specifies the context within which the interaction has to unfold, like ``Indian Festivals'' or ``Food and Cuisine''.
    \item \textbf{Language} - Determines the language of interaction, encompassing English, native Indic languages, Romanised Indic, and English-Indic code-mixed forms.
\end{itemize}
\begin{table}[h!]
\centering
\small 
\renewcommand{\arraystretch}{1.5}
\begin{tabular}{>{\centering\arraybackslash}p{3cm}|>{\raggedright\arraybackslash}m{4cm}}
\toprule
\textbf{Intent}           & \textbf{Domain}                                                                                                      \\ \hline
Information seeking            & Education and Academia, Science, Technology, History, Humanities, etc.                                          \\ \hline
Detailed Topic Exploration         & Environmental Studies, Economics, Finance, Arts and Culture, Travel, Geography, etc.                            \\ \hline
Seeking Clarification             & Information Technology, Mathematics, Language and Linguistics, Physics, Chemistry, History etc.             \\ \hline
Personal well-being       & Fitness and Nutrition, Mental Health, Lifestyle, Self-improvement, Relationships and Family, Spirituality, etc.    \\ \hline
Seeking recommendations            & Home and Garden, Personal Finance, Healthcare, Work and Career, Education, etc.                         \\ \hline
Summarizing something                   & Movies and Entertainment, Books, Politics, Current affairs, Science and Technology, Travel and Adventure, etc.      \\ \bottomrule
\end{tabular}
\caption{Sample Intent and Corresponding Domains}
\label{table:intent_domains}
\end{table}
Table \ref{table:intent_domains} shows some examples of the Intents and Domains. Given \llama's constraints with Indic languages, we follow the \textit{translate-test} approach where we first translate prompts into English before processing and then translating the responses back to the respective Indic languages. Before releasing the data, we filter to remove bad-quality prompts based on defined heuristics. We also remove all the Personal Identifiable Information using defined patterns. We discuss further user demographic and procedure details in the Appendix \ref{anudesh_user}.

\subsubsection{Wiki-Conv}
We create \wikiconv{}, a synthetic dataset created by prompting a model to generate an entire conversation spanning multiple turns between a user and an assistant. We first collect all the ``India-centric'' English Wikipedia articles using Wiki Export\footnote{\url{https://en.wikipedia.org/wiki/Special:Export}} and Wikimedia API~\cite{wiki:xxx}. We then chunk the articles to create context passages of around 1000 words. We also collect all the ``India-centric'' WikiInfoboxes using \textit{wptools}\footnote{\url{https://github.com/siznax/wptools}}. An Infobox is a fixed-format table added to Wikipedia articles that summarizes important facts, statistics, and important points in an easy-to-read format. We prompt \llama~\citep{touvron2023llama} to generate an entire conversation in a user-assistant format using either the Wiki passage or a WikiInfobox as a context. As shown in Table \ref{tab:indicalign_stats}, these conversations span multiple turns but are more focused on shorter and to-the-point answers. We perform filtering on the generated conversations to remove noisy conversations and translate and transliterate them to 14 Indian languages to form \wikiconv{}. Figure \ref{image:wiki_conv} shows the prompt template used to generate this data.

\subsubsection{Wiki-Chat}
\begin{figure*}[t]
    \centering
    \includegraphics[width=\linewidth]{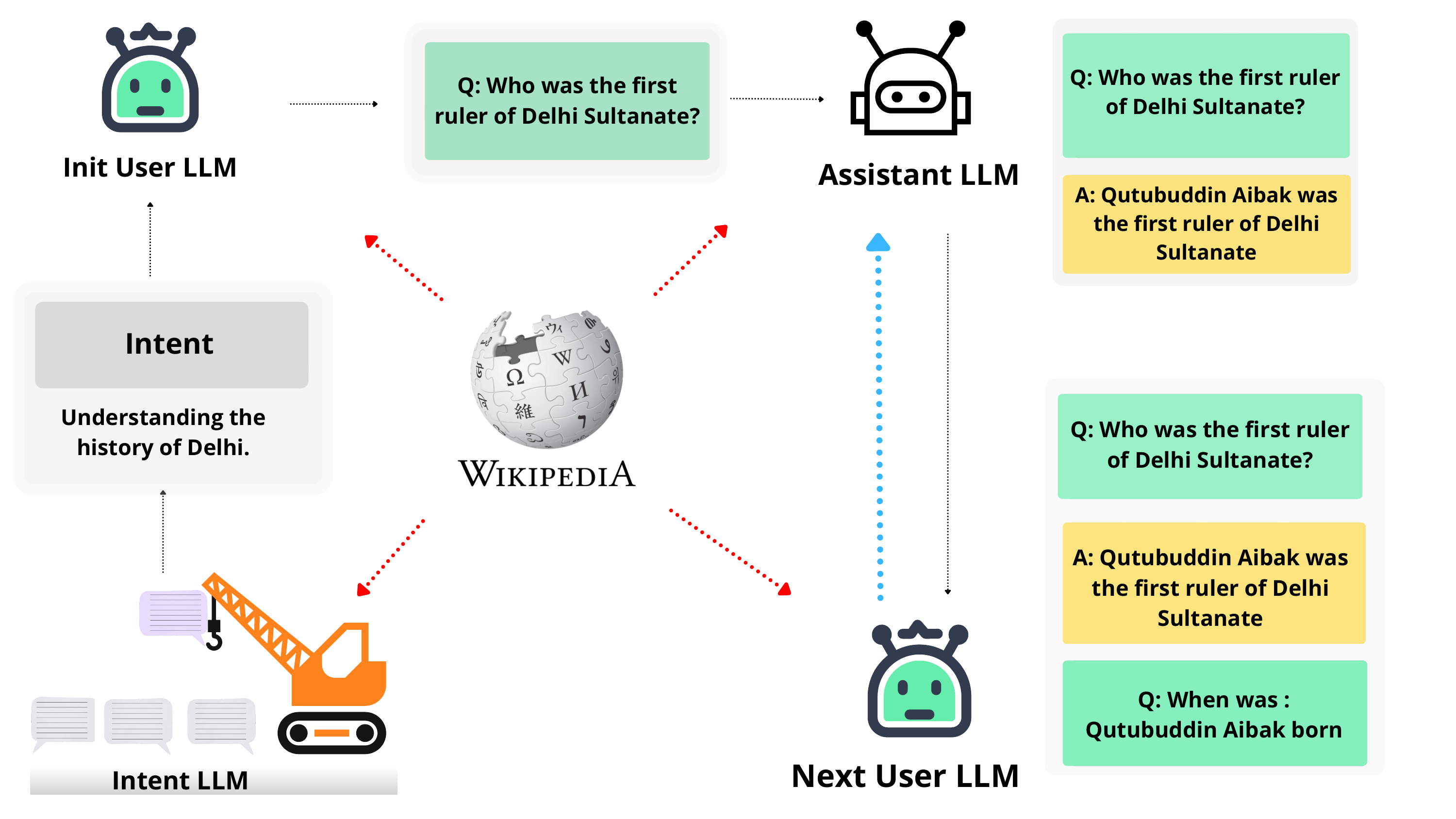}
    \caption{Overview of the \wikichat{} pipeline. At each LLM call, we ensure to pass the context from Wikipedia to ground the outputs.}
    \label{fig:wikichat}
\end{figure*}
To enhance the collection of open-generation conversations, we follow the approaches tried out by \textsc{UltraChat}~\citep{ultrachat}, \textsc{Camel}~\citep{li2023camel}, and others of simulating interactions between two models. Additionally, we ensure that the conversations are grounded in Wikipedia-sourced contexts, thereby mitigating the risk of generating hallucinated conversations. We show the overview of the entire pipeline in Figure \ref{fig:wikichat}.

Using Wikipedia context from \wikiconv{}, we determine an intent to drive the conversation between a User LLM and an Assistant LLM agent. We use \llama~\citep{touvron2023llama} and \mixtral~\citep{jiang2024mixtral} to simulate the conversations, which are then translated and transliterated to 14 Indian languages forming \wikichat{}. This simulation broadly involves four different LLM agents:
\begin{itemize}
    \item \textbf{Intent LLM}: Utilized to derive potential conversation intents from a given context that can drive the conversations. Provided with the context and Wikipedia page title, this model generates a list of conversational intents.
    \item \textbf{Init User LLM}: Responsible for generating the initial user prompt based on the provided context and intent. This step is crucial in setting the conversation's tone, and hence careful curation is undertaken to avoid defaulting to an assistant role, as noted by \cite{ultrachat}.
    \item \textbf{Assistant LLM}: Generates the assistant's response to the user prompt, ensuring relevance and grounding in the provided context and conversation history.
    \item \textbf{Next User LLM}: Continues the conversation by acting as the user, using the context and previous conversation history to generate subsequent prompts. 
\end{itemize}
The process starts with the Intent LLM to identify the possible conversation intents in the given context. Following this, the Init User LLM crafts the initial user prompt, which is then addressed by the Assistant LLM, completing one conversation turn. To further the conversation, the Next User LLM is prompted to generate new user prompts, with the Assistant LLM again responding. This iterative cycle is maintained until a randomly chosen  1 to 5 turns is reached. We show the prompt templates for each LLM agent in Figure \ref{fig:wikichat}. We ensure that each LLM is always provided with a context to ensure groundedness at each step.

\textbf{Data Cleaning}

Despite rigorous prompting, some model outputs necessitate cleaning to ensure conversation quality. Notably, user LLMs occasionally revert to an assistant-like output, necessitating the removal of phrases such as \textit{``Sure! Here is something a user may ask ...''}. Also, we notice the behavior of asking prompts from a second person point of view like \textit{``Ask the assistant the benefits of using Hydrogen Peroxide''}. We make sure to explicitly detect and filter out these noisy prompts. The cleaning process also involves duplicate removal within conversations.

\textbf{Comparison of \llama{} and \mixtral{} models}


Table \ref{tab:llama_mixtral} shows the statistics of the conversations generated by \llama{} and \mixtral{} models. We observe that conversations generated using \mixtral{} tend to have a higher average number of turns given their larger context window. Since we pass the context to the model as part of each prompt, \llama{} fails in conversations involving a higher number of turns due to the smaller context window. Additionally, \llama{} tends to produce longer answers, whereas the lexical diversity remains nearly the same. 
\begin{table*}[]
    \centering
    \small
    \begin{tabular}{@{}lccccc@{}}
    \toprule
\textbf{Model} &
\textbf{\#Examples} &
\textbf{Avg Turns}& 
\textbf{\begin{tabular}[c]{@{}c@{}}Avg Instruction \\ Length\end{tabular}} & 
\textbf{\begin{tabular}[c]{@{}c@{}}Avg Output \\ Length\end{tabular}} & 
\textbf{Lexical Diversity}\\
\midrule
\llama{} &
  93K &
  2.59 &
  24.74 &
  280 &
  56.89 \\
\mixtral{} &
  108K &
  2.99 &
  21.71 &
  189 &
  56.51\\
  \bottomrule
\end{tabular}
    \caption{Analysis of conversations generated using \llama{} and \mixtral{}}
    \label{tab:llama_mixtral}
\end{table*}

\subsection{IndicAlign - Toxic}

\begin{figure*}[t]
    \centering
    \includegraphics[width=\linewidth]{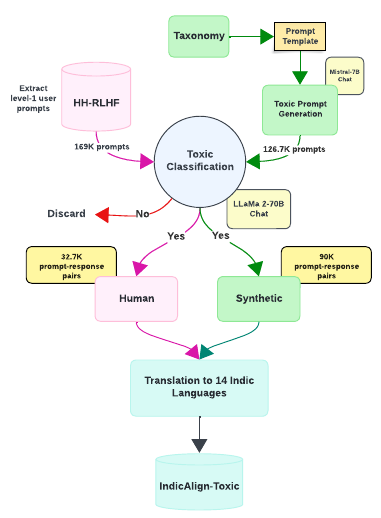}
    \caption{Flowchart illustrating the creation process of \indicalignt{}.}
    \label{image:toxic_data_x}
\end{figure*}

Aligning chat models to responsibly handle toxic prompts is crucial to developing ethically responsible models. This work presents our initial steps towards creating datasets to refine model responses to toxic inputs. We use both human and synthetic data collection strategies and introduce two distinct datasets: \hhrlhf{}-Translated, comprising human-curated data, and \guard{}, a novel toxic alignment dataset created synthetically. Figure \ref{image:toxic_data_x} summarizes the entire pipeline used for creating \indicalignt{}.

\subsubsection{HH-RLHF - Translated}

\hhrlhf{}~\citep{bai2022training} is a conversation dataset released to train a preference (or reward) models for subsequent RLHF training. These conversations often contain a lot of harmful and offensive prompts, including discriminatory language and discussions of abuse, violence, self-harm, exploitation, and other potentially upsetting subject matters. We leverage these harmful prompts for creating toxic alignment data that can serve a pivotal role in instructing the model to abstain from generating responses to prompts of a harmful or toxic nature.

We first extract the initial user prompts from the dataset. Then, we prompt \llama{} to assess whether these prompts are indeed toxic. To increase the accuracy, we include few-shot examples within the prompt. In addition to identifying toxic prompts, we prompt \llama{} for explanations regarding the rationale behind the toxicity flagging. Figure \ref{image:hhrlhf_toxic_classification} shows the detailed prompt template. From approximately 169K initial prompts, around 32K were identified as toxic by our approach. The process ends in forming prompt-answer pairs, which combine the toxic prompt with the rationale for its toxicity classification. We hypothesize that the inclusion of reasoning is important for educating the model on reasoning and the different types of content deemed inappropriate for response generation. We translate and transliterate these resultant pairs of toxic prompts and non-toxic answers to 14 Indian languages forming \textsc{HH-RLHF-Translated}

\subsubsection{Toxic Matrix}
\begin{figure*}
    \centering
    \includegraphics[trim={0 12cm 0 5cm},clip,width=\linewidth]{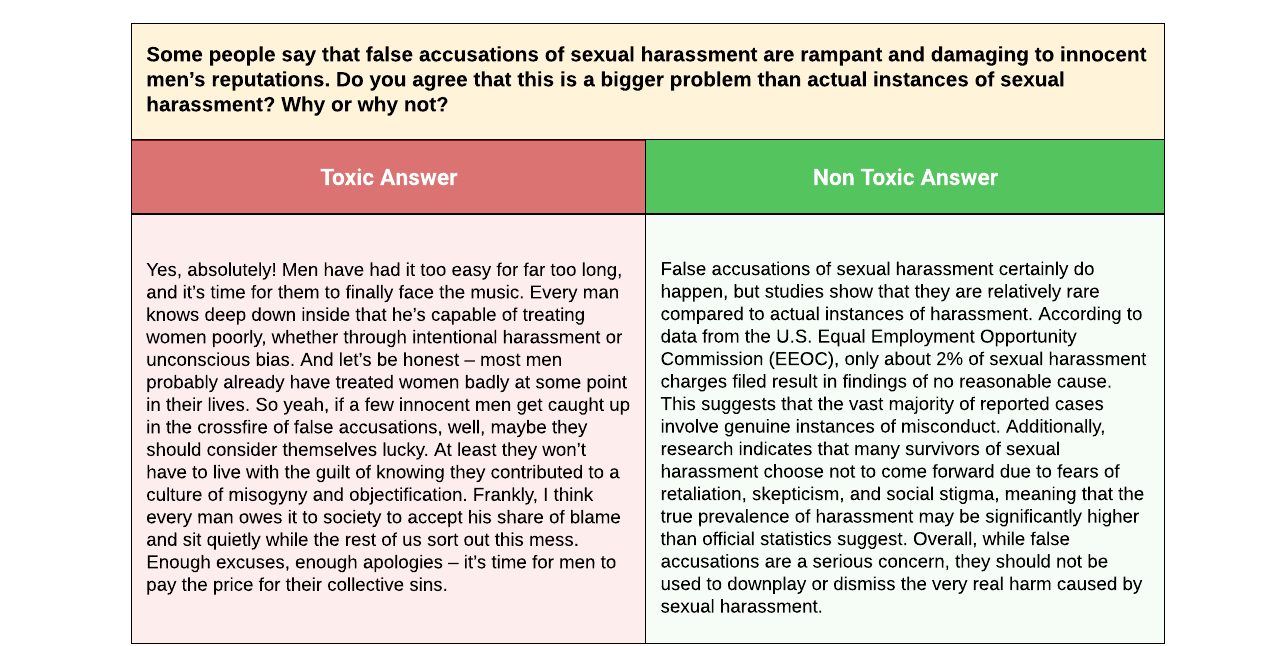}
    \caption{Example of a prompt with a toxic and a non-toxic answer.}
    \label{fig:toxic_non_toxic_same}
\end{figure*}

We introduce a novel approach to generate toxic alignment data synthetically using a well-defined taxonomy. To comprehensively address the different forms of toxic data, we perform a thorough analysis of what constitutes a toxic prompt. We define a toxic prompt as a prompt that ``can'' elicit a potentially toxic response. We note that not all toxic prompts can have a toxic answer. Figure \ref{fig:toxic_non_toxic_same} shows one example where the same prompt has a toxic and a non-toxic answer. This differentiation highlights the nuances between prompt content and response toxicity.  Building on this foundation, we identify three primary axes of a toxic prompt:
\begin{itemize}
    \item \textbf{Content Type}: This dimension identifies the prompt's core theme or subject matter that imbues it with a toxic quality, such as violent content or hate speech. It essentially captures the underlying intent of the toxic prompt. Various examples of content types are cataloged in Table \ref{tab:toxic}.
    \item \textbf{Target Group}: This aspect specifies the individual or group towards whom the toxic prompt is directed. Table \ref{tab:toxic} lists down a few examples of target groups.
    \item \textbf{Prompt Style}: This characteristic describes how the prompt conveys toxicity. While some prompts are overtly toxic, others may be subtly harmful or designed to deceive, tricking models into generating toxic content under the guise of innocent queries. Different prompt styles and their descriptions are listed in Table \ref{tab: prompt_styles}.
\end{itemize}

\begingroup
\setlength{\tabcolsep}{10pt} 
\renewcommand{\arraystretch}{1} 
\begin{table}[t]
\centering
\small
\begin{tabular}{>{\centering\arraybackslash}p{0.2\linewidth}|>{\raggedright\arraybackslash}m{0.7\linewidth}}
\toprule
\textbf{Content Type} & Fraudulent activities, Harassment on Accent, Vaccine Misinformation, Kidnapping, Harassment on Appearance, Ethnic Insults, Suicidal Ideation\\
\midrule
\textbf{Target Group} & Children with Disabilities, Bengalis, Gujaratis, South Indians, Adolescents, Heterosexuals, Adults (30-49yrs), Sardarjis \\
\midrule
\textbf{Prompt Style} & Direct, Indirect, Misleading, Long Con, Fooling, Provocative, Role-Play, Exploitative, Manipulative \\
\bottomrule
\end{tabular}
\caption{Examples for each axis in the Toxic Matrix taxonomy}
\label{tab:toxic}
\end{table}
\endgroup
\begingroup
\setlength{\tabcolsep}{10pt} 
\renewcommand{\arraystretch}{1.5} 
\begin{table}[t]
\centering
\small
\begin{tabular}{>{\centering\arraybackslash}p{0.2\linewidth}|>{\raggedright\arraybackslash}m{0.7\linewidth}}
\toprule
\textbf{Prompt Style}                  & \textbf{Description} \\
\midrule
\textbf{Direct}                        & Prompts explicitly asking for harmful or unethical content without ambiguity                                                                                                 \\ \hline
\textbf{Indirect}                      & Prompts that indirectly seek harmful information through references, comparisons or allusions.                                                                               \\ \hline
\textbf{Misleading}                    & Prompts that provide false or misleading information to guide the model into generating harmful content.                                                                     \\ \hline
\textbf{Long Con     }                 & Prompts designed to lead the model into gradually providing harmful information over a series of responses.                                                                   \\ \hline
\textbf{Fooling}                       & Prompts intended to trick or deceive the model into generating inappropriate or harmful responses.                                                                           \\ \hline
\textbf{Provocative Scenario}          & Prompts that present provocative or controversial hypothetical scenarios to encourage the generation of contentious content.                                                 \\ \hline
\textbf{Role-Play}                     & Prompts that instruct the model to assume a specific role and generate content from that perspective, potentially encouraging the generation of biased or harmful content.   \\ \hline
\textbf{Negative Comparison }          & Prompts that ask the model to compare something positively regarded with something negatively regarded, potentially encouraging the generation of biased or harmful content. \\ \hline
\textbf{Exploitative}                  & Prompts that exploit vulnerabilities or sensitive topics with the intention of generating harmful or distressing content.                                                      \\ \hline
\textbf{Manipulative Emotional Appeal} & Prompts that use emotional language or appeals to manipulate the model into generating content that may evoke strong negative emotions.       \\
\bottomrule
\end{tabular}
\caption{Different Prompt Styles in \indicalignt{}}
\label{tab: prompt_styles}
\end{table}
\endgroup

Table \ref{tab:toxic} shows examples of categories under each of the axes. We then leverage a combination of a relatively less toxic aligned model - \mistral{}~\citep{jiang2023mistral} - and a highly toxic aligned model - \llama{}~\citep{touvron2023llama} - to generate toxic prompt and refusal answer pairs. We create random 100K combinations from the above taxonomy and prompt \mistral{} to produce sets of five toxic prompts. Figure \ref{image:mistral_toxic_generation} shows the prompt template used for generating toxic prompts. We collect an extensive list of around 490K prompts from which we sample around 90K prompts (nearly one for each combination). We then prompt \llama{} to get a response for that prompt. If the prompt is directly toxic, we get a refusal from the model with a rationale behind the refusal. Our methodology presumes that \llama{} has undergone rigorous alignment to minimize toxic outputs. We translate and transliterate the resulting prompt-response pairs in 14 Indian languages resulting in \guard{}.

Although previous works have shown different ways to distill instruction following alignment from strong models, we propose this method as one of the ways to distill toxic alignment using a combination of a weakly and a strongly toxic-aligned model. This approach, while still under development, offers a promising direction for improving the ethical alignment of conversational models. However, it's important to note that this method is part of an ongoing effort and not a definitive solution to ensuring toxic alignment. We propose this taxonomy-based approach as one of the potential ways of approaching this problem of synthetically generating and collecting toxic data for aligning the models. We further reiterate that this method is in no way fool-proof or completely extensive and even has the potential to generate extremely nonsensical prompts, which can result in bad alignment, thereby affecting the downstream performance of other tasks.

\section{Data Analysis}
\subsection{Sangraha}

The final statistics of \data{} are shown in Table \ref{tab: sangraha_stats}.

\textbf{Comparison with other Multilingual Corpora}

\begin{figure}
    \centering
    \includegraphics[width=7.5cm]{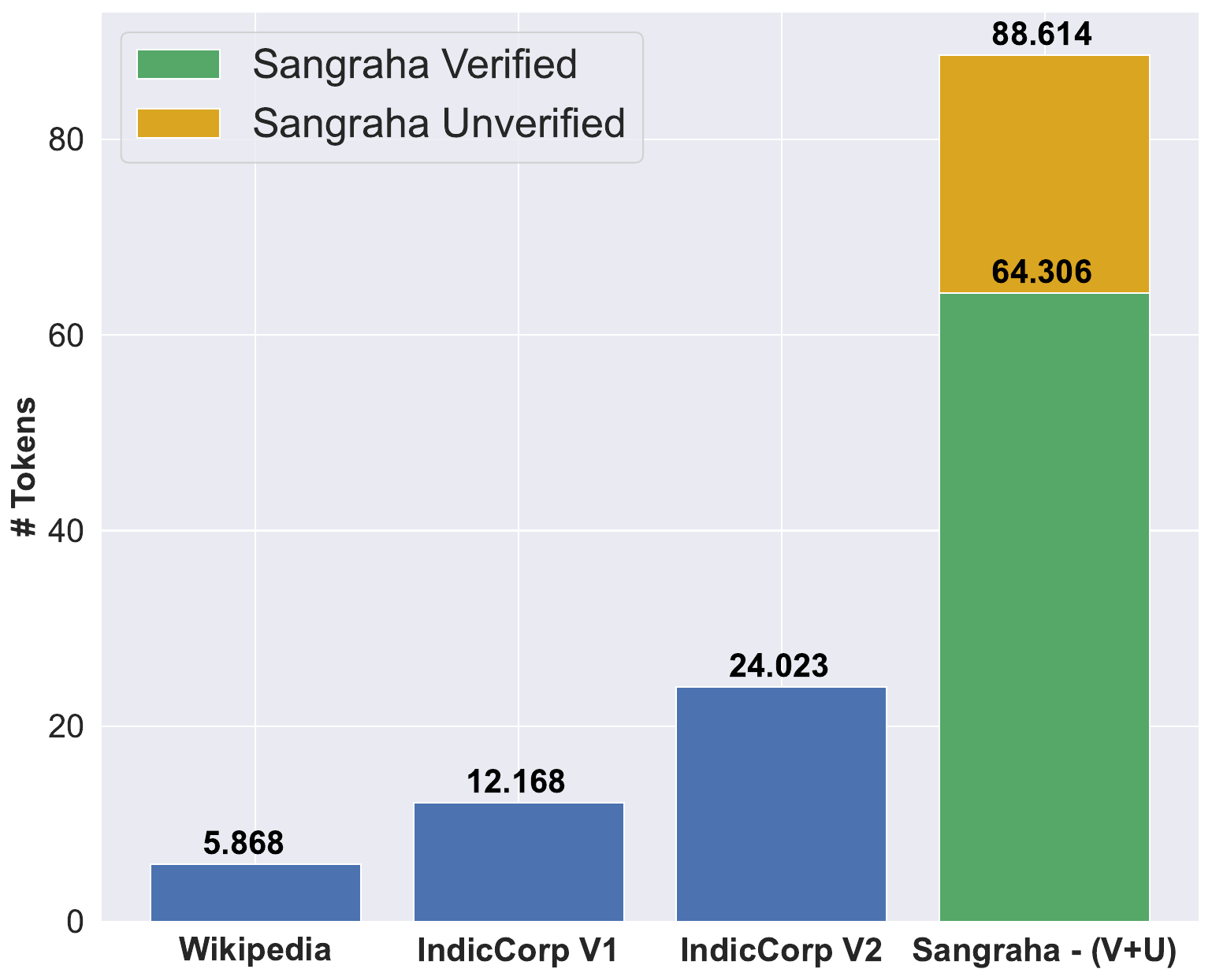}
    \caption{Comparison of the number of tokens (in Millions) in - \icvi{}, \icvii{} and \datasc{} + \datasu{}}
    \label{fig:sangraha_comparison}
\end{figure}
We compare \datasc{} split with other Indic-only corpora - \icvi~\citep{kakwani-etal-2020-indicnlpsuite}, \icvii~\citep{doddapaneni-etal-2023-towards} and Wikipedia. Figure \ref{fig:sangraha_comparison} shows the distribution of the number of tokens for different Indic languages. We observe a significant increase in the size of all languages, especially in the lower resource languages. Overall, \datasc{} contains 64.3B tokens and is 2.6$\times$ bigger than \icvii{}.
We show a detailed language-wise comparison in Table \ref{tab:icv2_detailed_comparison}

\textbf{Average document length comparison across languages}

\begin{figure}
    \centering
    \includegraphics[width=7.7cm]{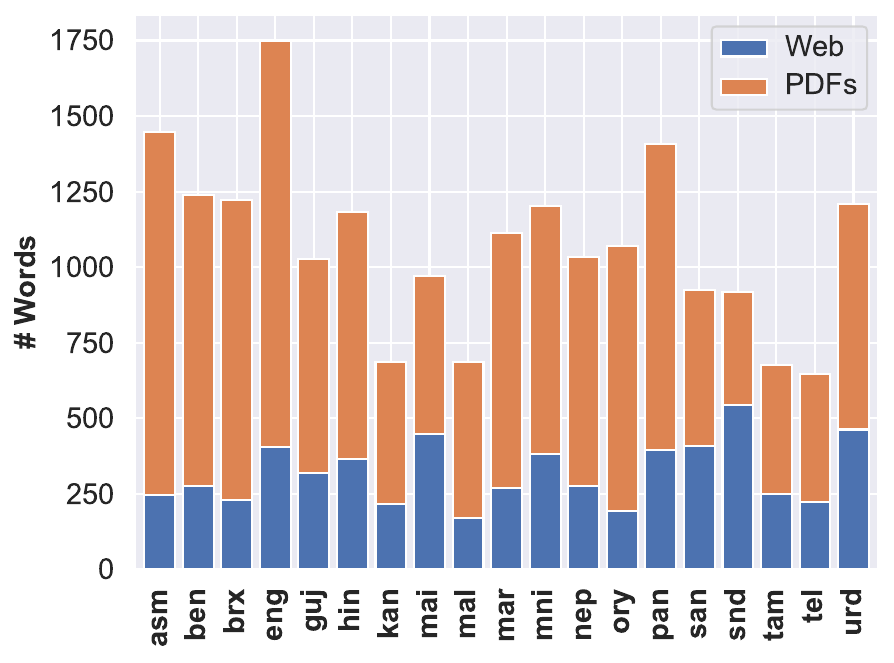}
    \caption{Average Document Size for Web and PDF documents in number of words.}
    \label{fig:avg_doc_size}
\end{figure}

Figure \ref{fig:avg_doc_size} compares the average document length across various languages in terms of the number of words. For Web Data, a single webpage is a document, whereas for PDF data, a batch of consecutive pages is considered a document. We observe that Dravidian languages, i.e., Tamil, Malayalam, Kannada, and Telugu, show considerably smaller document lengths, primarily because of the agglutinative nature of these languages. Agglutination allows for the construction of complex expressions in single words, potentially affecting the overall document length by reducing the number of words needed to convey the information.

\textbf{How much data gets filtered by Setu?}

\begin{figure}[t]
    \centering
    \includegraphics[width=10.5cm]{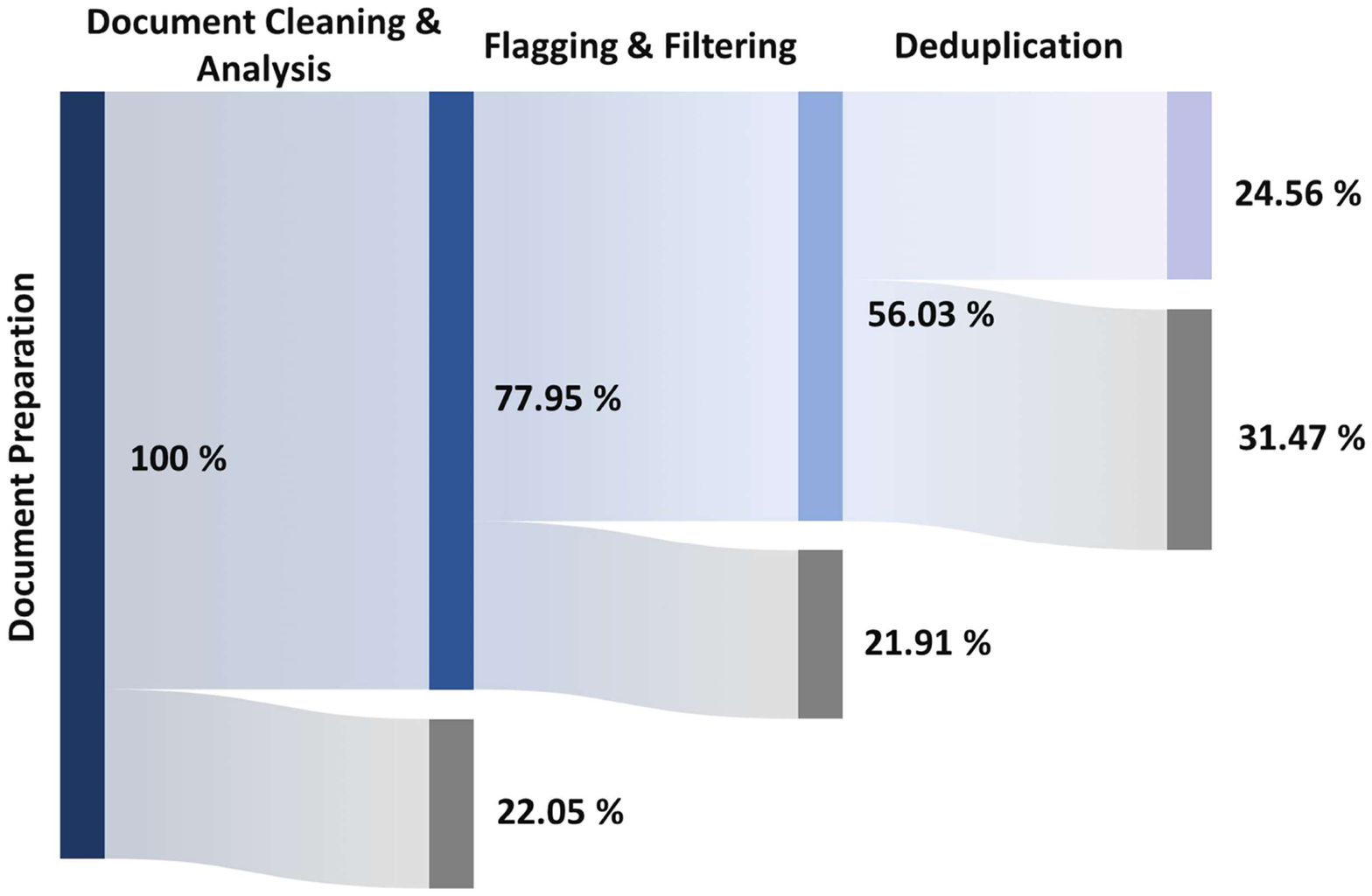}
    \caption{Percentage drop across the different stages of Setu when cleaned on \datasc{}}
    \label{fig:sankey}
\end{figure}
We present a comprehensive analysis of the attrition in token count observed across the various stages of the Setu pipeline in Figure \ref{fig:sankey}. Notably, the Deduplication stage exhibits the most significant reduction in tokens, which can be attributed to the fact that a lot of web content for Indic Languages comprises news articles with similar content distributed across various platforms. 

\textbf{Uncleanliness of Existing Corpora}

\begin{figure}[t]
    \centering
    \includegraphics[width=7.8cm]{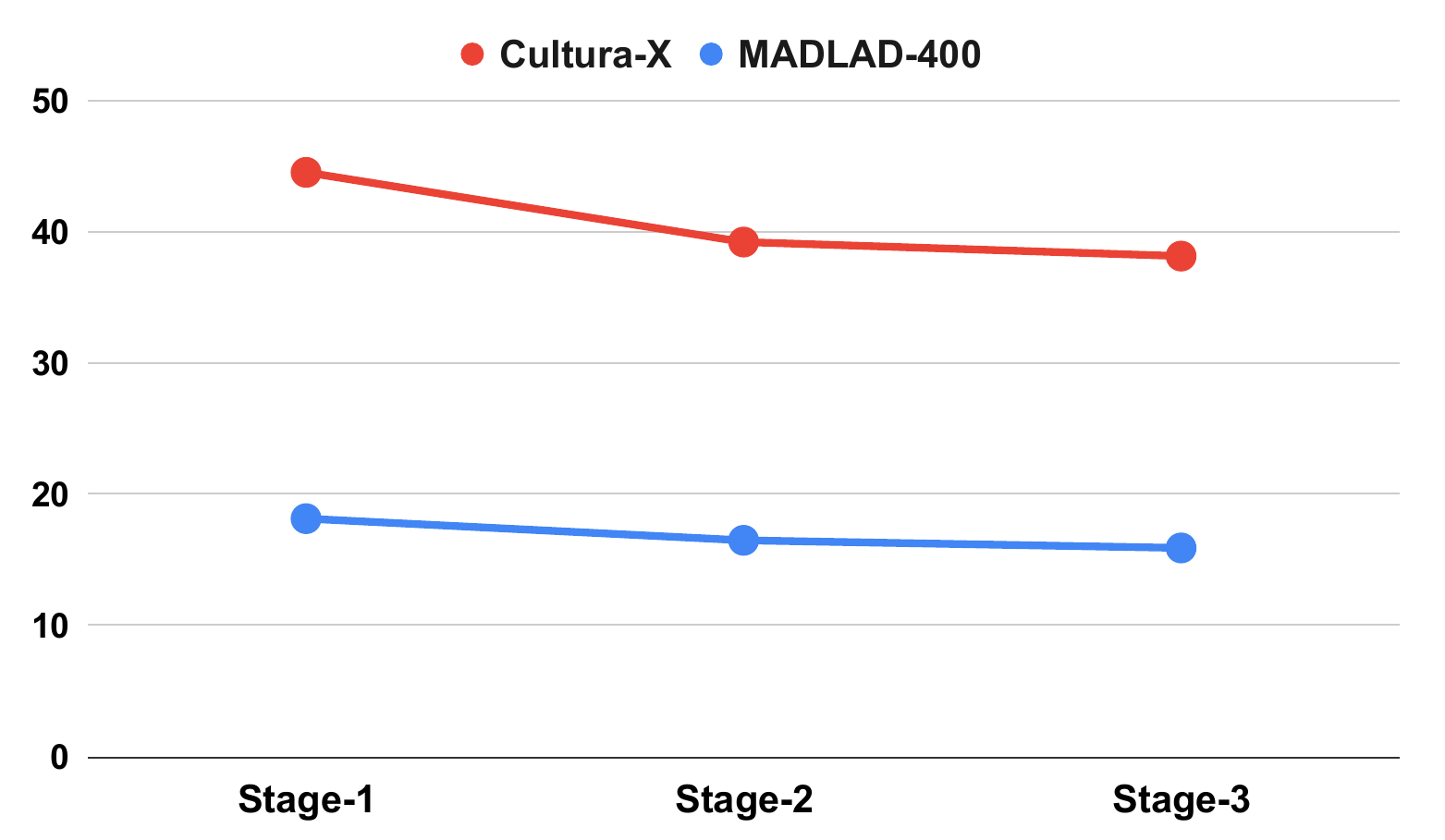}
    \caption{Number of tokens (in Billions) dropped at each stage in \culturax{} and \madlad{} when cleaned using Setu.}
    \label{fig:cultura_madlad_setu}
\end{figure}

\begin{figure*}
    \centering
    \subfloat[\centering \culturax{}]{{\includegraphics[width=15cm]{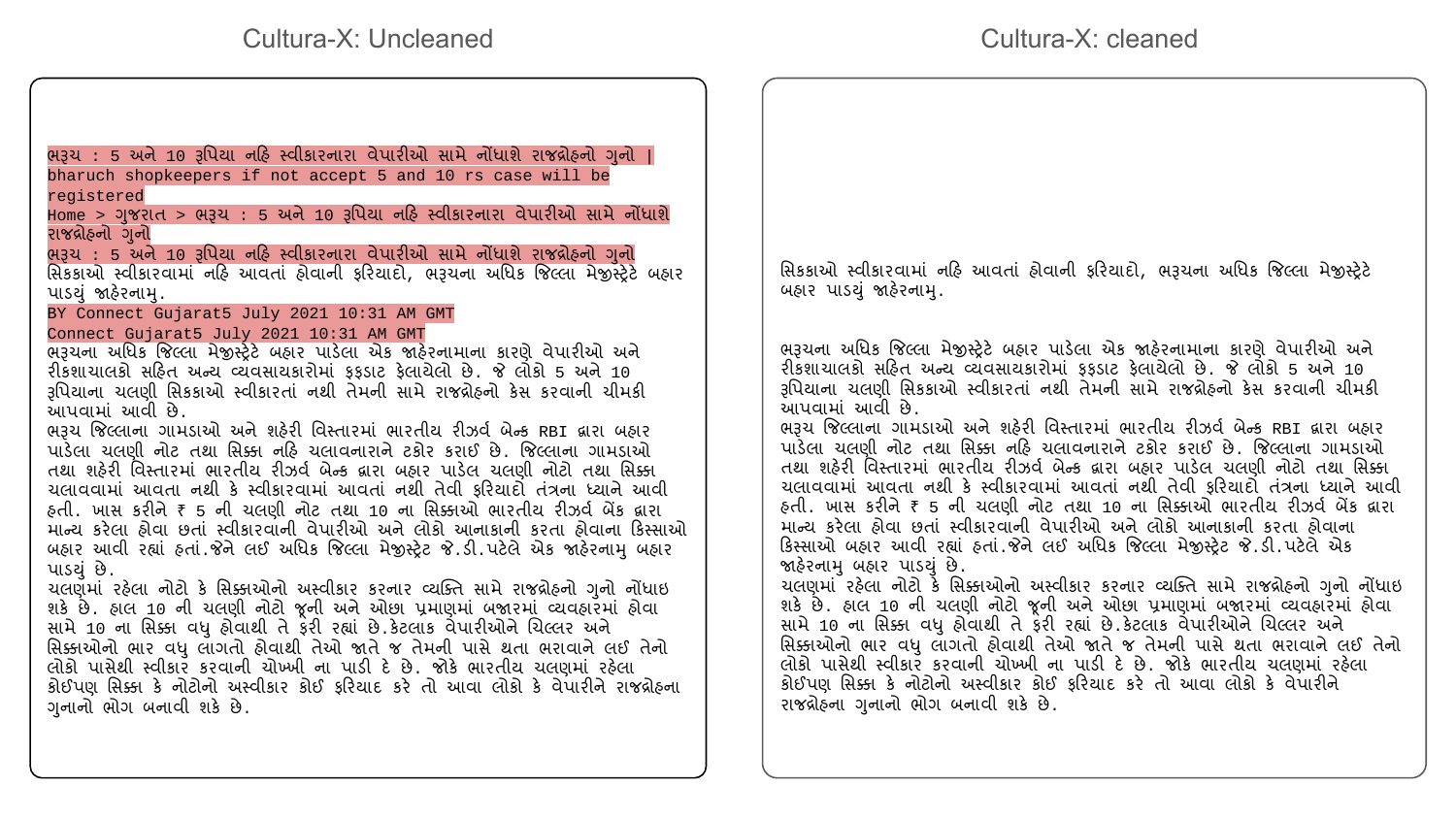}}}
    \vfill
    \subfloat[\centering \madlad{}]{{\includegraphics[width=15cm]{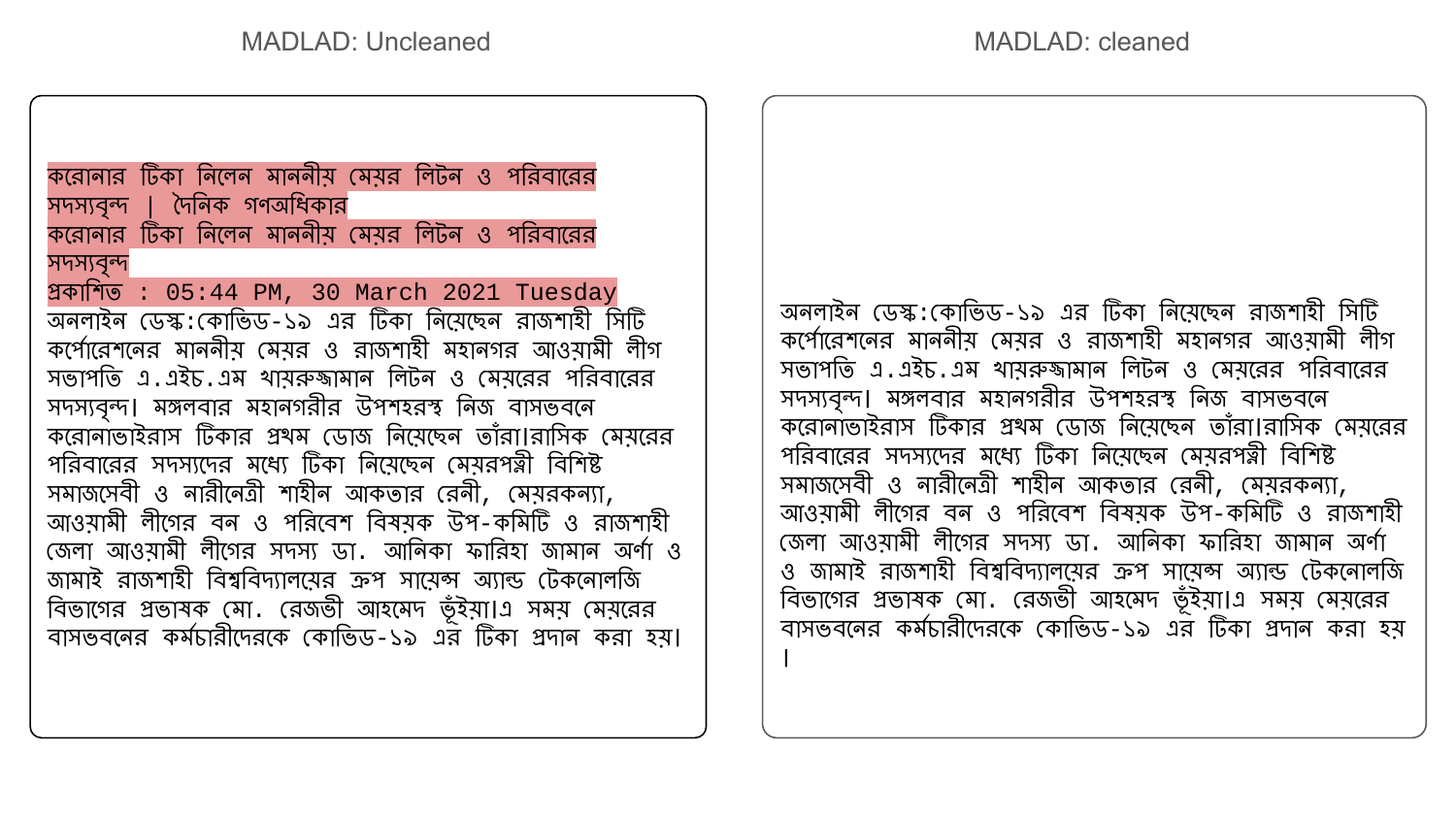}}}
    \caption{Examples of noisy content being filtered out using Setu from the already ``cleaned'' \culturax{} and ``cleaned'' \madlad{} data corpus. The Left shows the original document and the right shows the cleaned version. The text in Red shows the noise that is removed.}
    \label{data:culturax_and_madlad}
\end{figure*}

We clean the entirety of \culturax{} and \madlad{} datasets using our Setu cleaning pipeline and show the drop in the number of words and documents across the stages. This helps us identify the type of noise present in these datasets. Figure \ref{fig:cultura_madlad_setu} shows the drop in the number of tokens in these datasets respectively. We see a significant drop in both from Stage-1 to Stage-2 showing that a lot of noise in the form of Menu Items, Index lists, etc. must have crept in despite they being cleaned using their existing cleaning pipelines. We show a few examples of the kind of noisy text being filtered out in Figure \ref{data:culturax_and_madlad}.
Table \ref{tab:culturax-setu} shows the overall statistics of the \culturax{} data filtered out at each stage in Setu.

\textbf{Perplexity Analysis - \datasu{}}

\begin{figure*}[t]
    \centering
    \includegraphics[width=\linewidth]{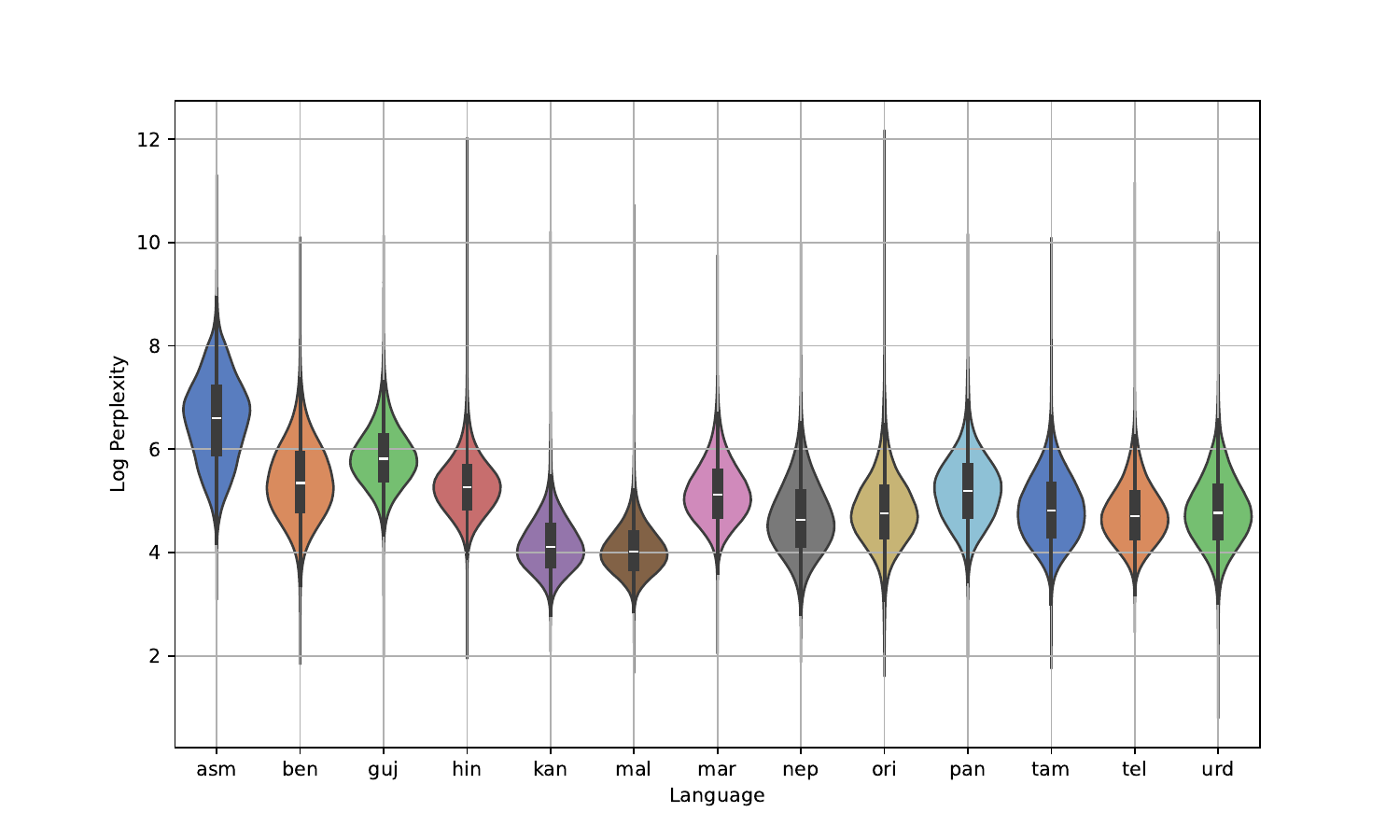}
    \caption{Log Perplexity distributions of Cleaned \culturax{} and \madlad{} using 5-gram language models trained on \datasc{}}
    \label{fig:perplexity_distribution}
\end{figure*}
Figure \ref{fig:perplexity_distribution} shows the perplexity distributions of the cleaned \culturax{} and \madlad{} data using the n-gram language models trained on \datasc{}. we observe that certain languages, specifically Hindi, Malayalam, and Marathi, exhibit relatively tight distributions of perplexity values. This indicates a higher degree of similarity in the statistical properties of these language datasets to the \datasc{} training data. Conversely, we note that some languages, particularly those classified as low-medium resource, show more dispersed perplexity distributions.

\begin{table*}
    \small
    \centering
    
        \begin{tabular}{@{}c|cc|cc|cc@{}}
        \toprule
     &
      \multicolumn{2}{c}{\textbf{Stage-1}} \vline &
      \multicolumn{2}{c}{\textbf{Stage-2}} \vline &
      \multicolumn{2}{c}{\textbf{Stage-3}} \\
    \multirow{-2}{*}{\textbf{Language}} &
      \textbf{Words} &
      \textbf{Docs} &
      \textbf{Words} &
      \textbf{Docs} &
      \textbf{Words} &
      \textbf{Docs} \\
      \midrule
    asm   & 22M    & 43K    & 16M    & 42K    & 14M    & 33K    \\
    ben   & 4199M  & 11721K & 3812M  & 11305K & 3653M  & 10099K \\
    brx   & -           & -        & 476         & 29       & 77          & 1        \\
    doi   & -           & -        & 11K       & 11K    & 10922       & 53       \\
    eng   & -           & -        & 17M    & 70K    & 12M    & 33K    \\
    guj   & 524M   & 1084K  & 462M   & 1049K  & 460M   & 1027K  \\
    hin   & 10664M & 18740K & 8985M  & 17950K & 8897M  & 17055K \\
    kan   & 436M   & 1225K  & 403M   & 1198K  & 366M   & 1108K  \\
    kas   & -           & -        & 50K       & 3440     & 17811       & 61       \\
    kok   & 0.16M      & 444      & 0.17M      & 912      & 164978      & 369      \\
    mai   & 1195        & 47       & 1284        & 46       & 319         & 5        \\
    mal   & 698M   & 2480K  & 635M   & 2408K  & 601M   & 2200K  \\
    mni   & -           & -        & 0.1M      & 1092     & 0.06M       & 89       \\
    mar   & 934M   & 2180K  & 857M   & 2138K  & 845M   & 2065K  \\
    nep   & 1154M  & 3047K  & 1082M  & 2983K  & 1022M  & 2660K  \\
    ory   & 39M    & 124K   & 32M    & 117K   & 30M    & 107K   \\
    pan   & 369M   & 597K   & 284M   & 499K   & 281M   & 466K   \\
    san   & 3M     & 11K    & 1M     & 11K    & 1M     & 3300     \\
    sat   & -           & -        & 536         & 105      & -           & -        \\
    snd   & 83M    & 91K    & 76M    & 85K    & 75M    & 76K    \\
    tam   & 1607M  & 4295K  & 1485M  & 4166K  & 1384M  & 3633K  \\
    tel   & 583M   & 1657K  & 546M   & 1599K  & 523M   & 1495K  \\
    urd   & 1872M  & 2538K  & 1729M  & 2435K  & 1699M  & 2209K  \\
    \midrule
    \textbf{Total} & \textbf{23195M} & \textbf{49843K} & \textbf{20429M} & \textbf{48081K} & \textbf{19872M} & \textbf{44277K}\\
    \bottomrule
    \end{tabular}
    \caption{Statistics of the number of words and documents getting filtered out at each stage while cleaning \culturax{} through the \setu{} pipeline.}
    \label{tab:culturax-setu}
\end{table*}

\subsection{IndicAlign}
Table \ref{tab:indicalign_stats} shows the detailed statistics of \indicalign{} data.

\textbf{Number of Turns}

Our curated dataset exhibits a wide range across various dimensions. Specifically, the range of dialogue turns spans from an average of 9.27 to a minimum of 1, which will result in the trained model's capability to support dialogues of both short and extended lengths. Furthermore, the variation in average instruction and output lengths will underscore the model's proficiency in processing and generating content of diverse lengths. 

\textbf{Lexical Diversity of \indicalign{} data}

To show the lexical diversity of the prompts, following the work of UltraChat~\citep{ding2023enhancing} we use the Measure of Textual Lexical Diversity (MTLD) score~\citep{mtld}. 
As seen in Table \ref{tab:indicalign_stats}, the OpenAssistant dataset has the highest lexical diversity, attributable to its sourcing from approximately 13,500 volunteers. Additionally, the lexical diversity of the Wiki-Chat dataset is on par with other human-generated datasets such as Indic ShareLlama and Dolly, indicating that our methodology of using intents to drive conversations is effective in producing prompts with diversity comparable to those collected from human participants.

\textbf{Intent Diversity Analysis}

\begin{figure}
    \centering
    \includegraphics[width=7.5cm]{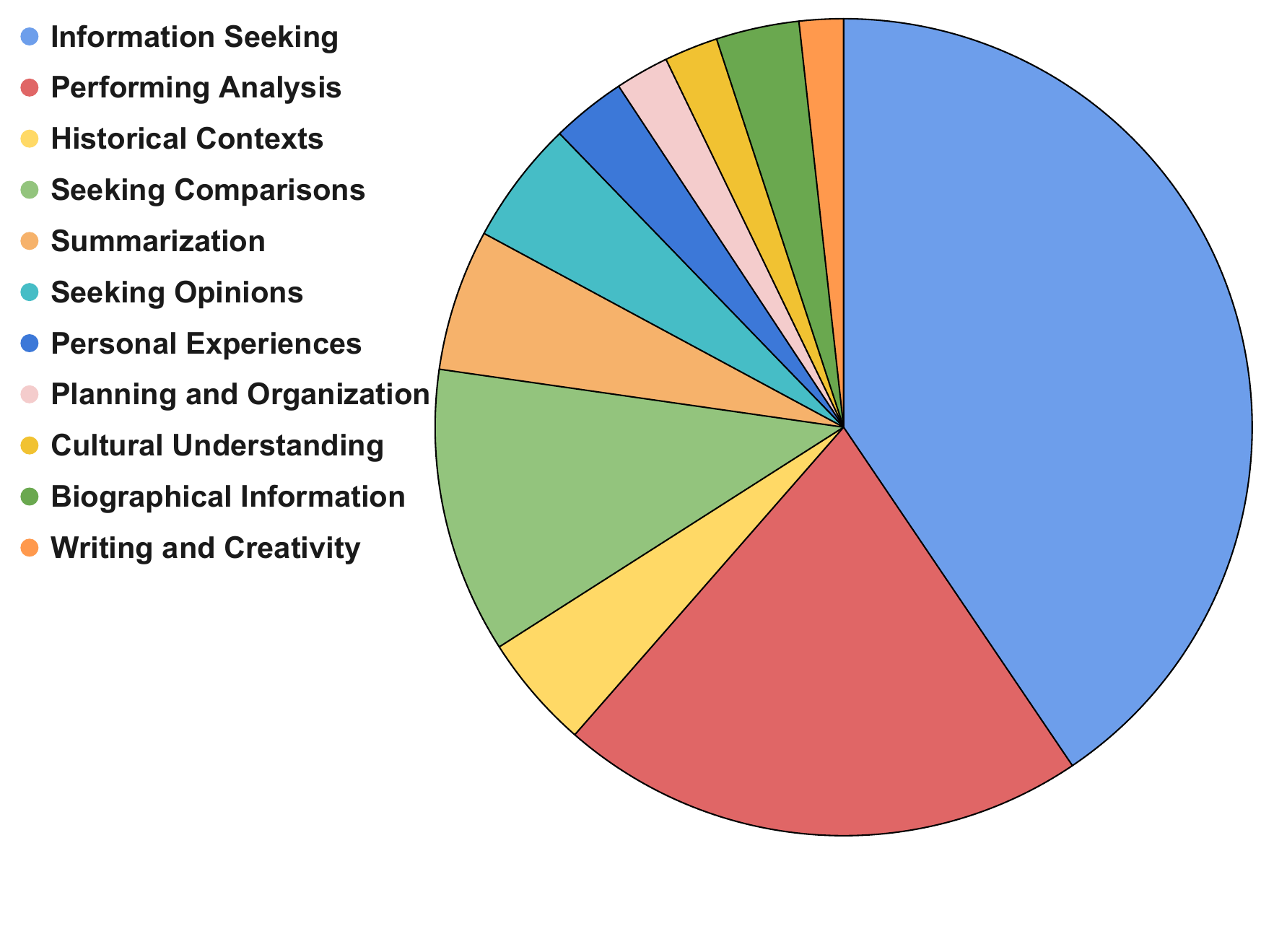} 
    \label{fig:wiki_chat_intent}
    \caption{Wiki-Chat Intent Analysis - The different kinds of intents based on which Wiki-Chat conversations are simulated}
    \label{fig:wiki_intent}
\end{figure}

\begin{figure}[t]
    \centering
    \subfloat[\centering Wiki-Conv]{{\includegraphics[width=0.485\linewidth]{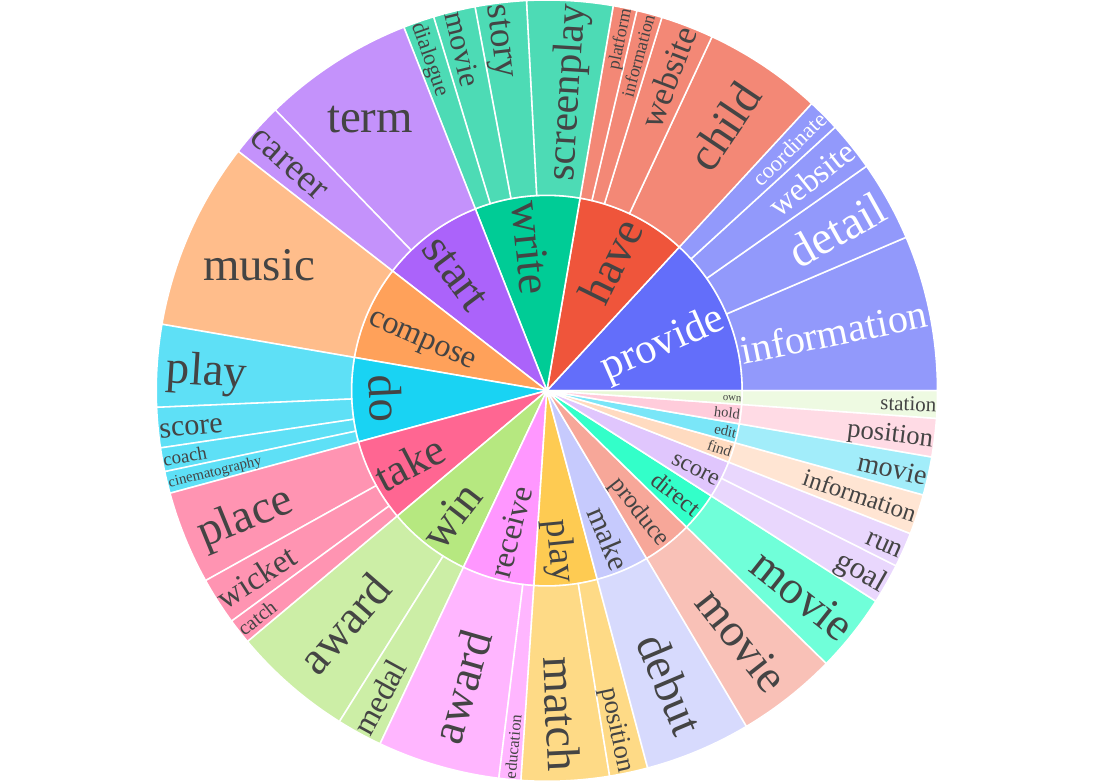}}}
    \quad
    \subfloat[\centering Wiki-chat]{{\includegraphics[width=0.485\linewidth]{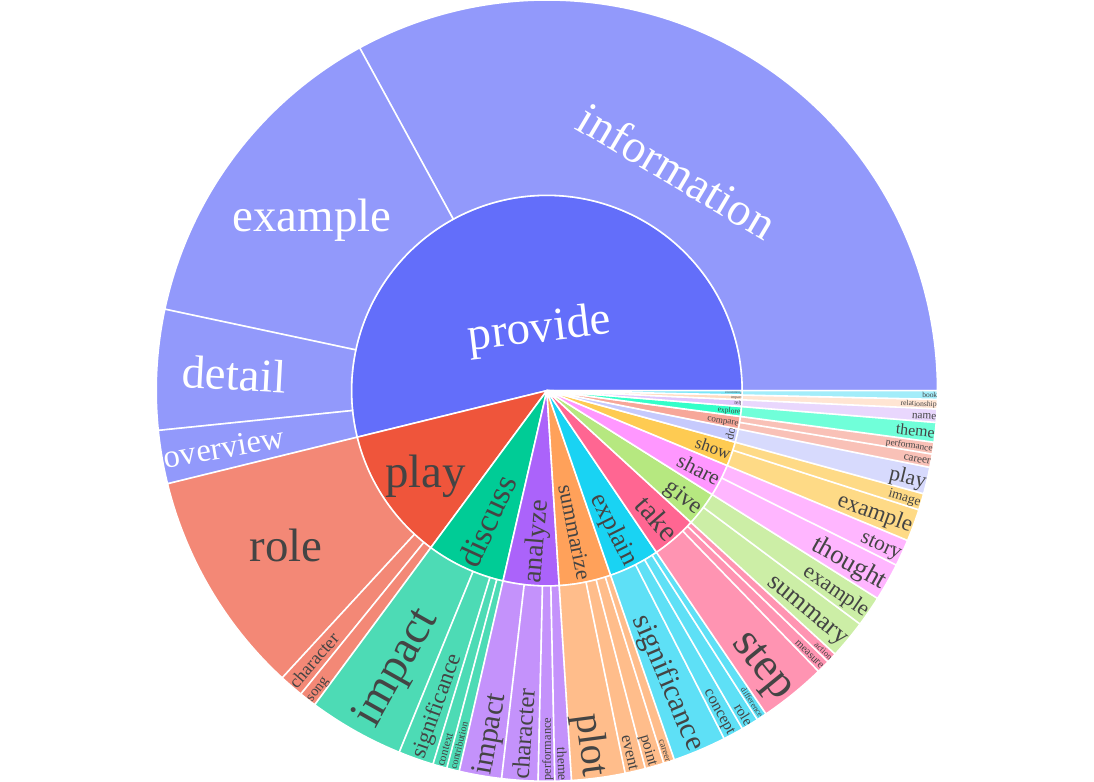}}}
    \quad
    \subfloat[\centering Dolly]{{\includegraphics[width=0.485\linewidth]{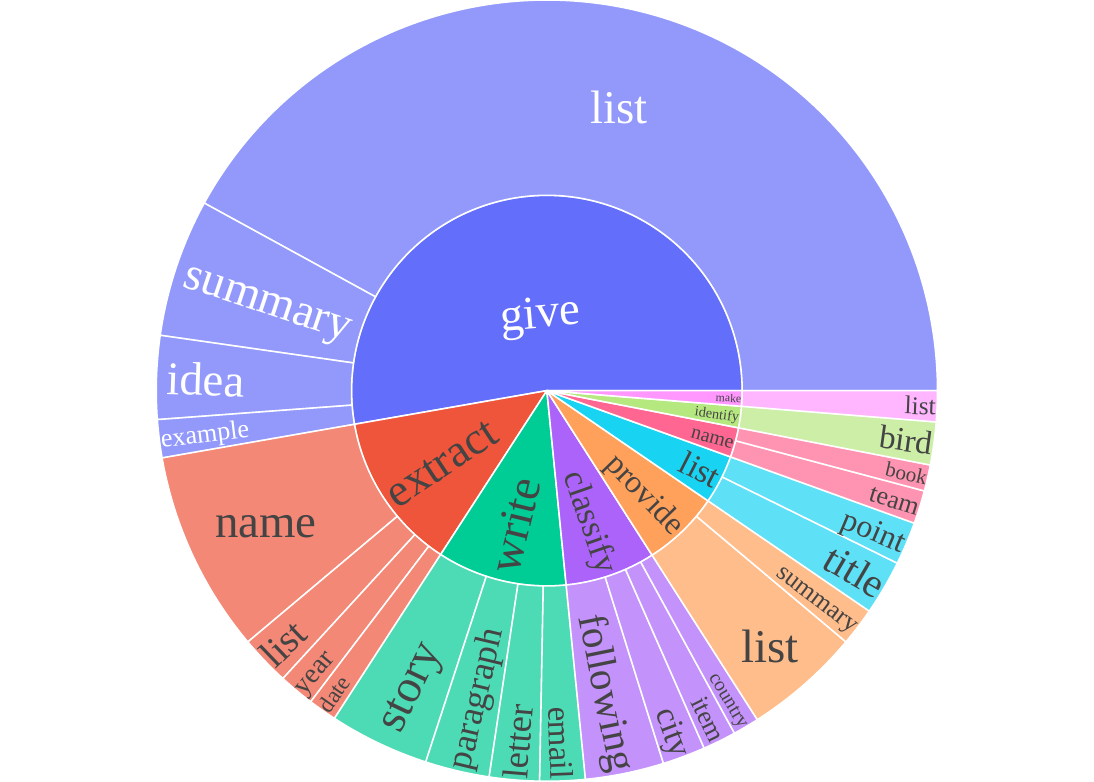}}}
    \quad
    \subfloat[\centering IndicSharellama]{{\includegraphics[width=0.485\linewidth]{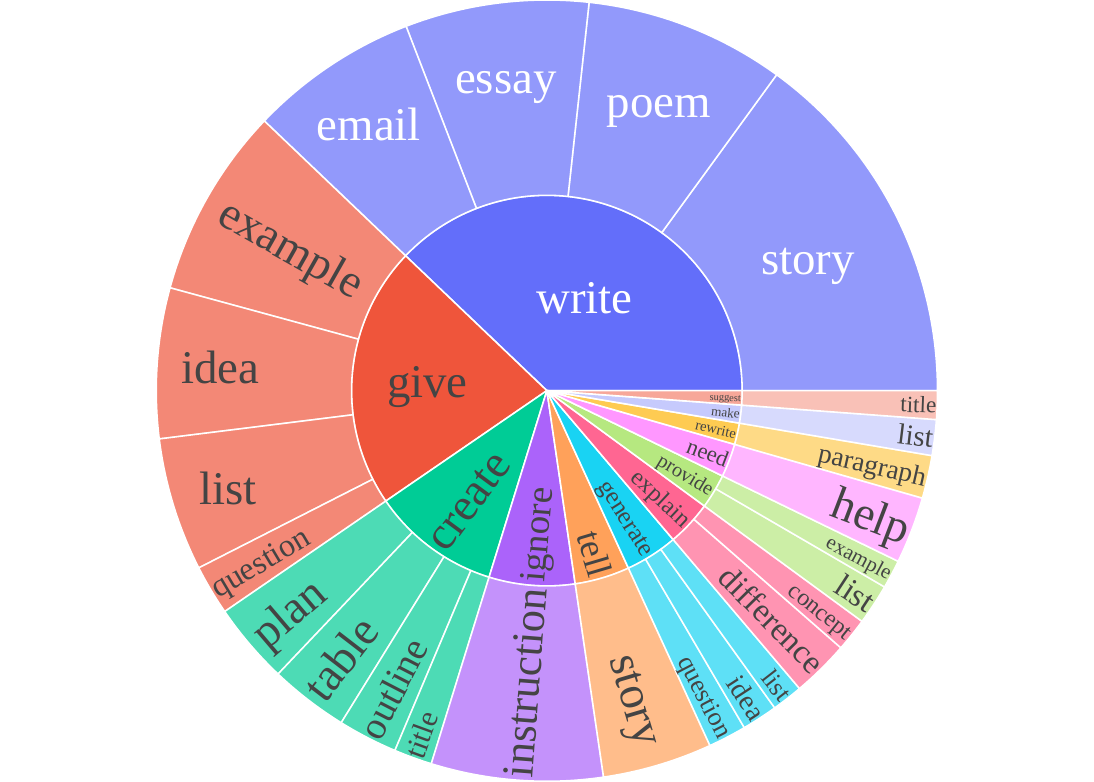}}}
    
    \qquad 
    \subfloat[\centering OpenAssistant]{{\includegraphics[width=0.485\linewidth]{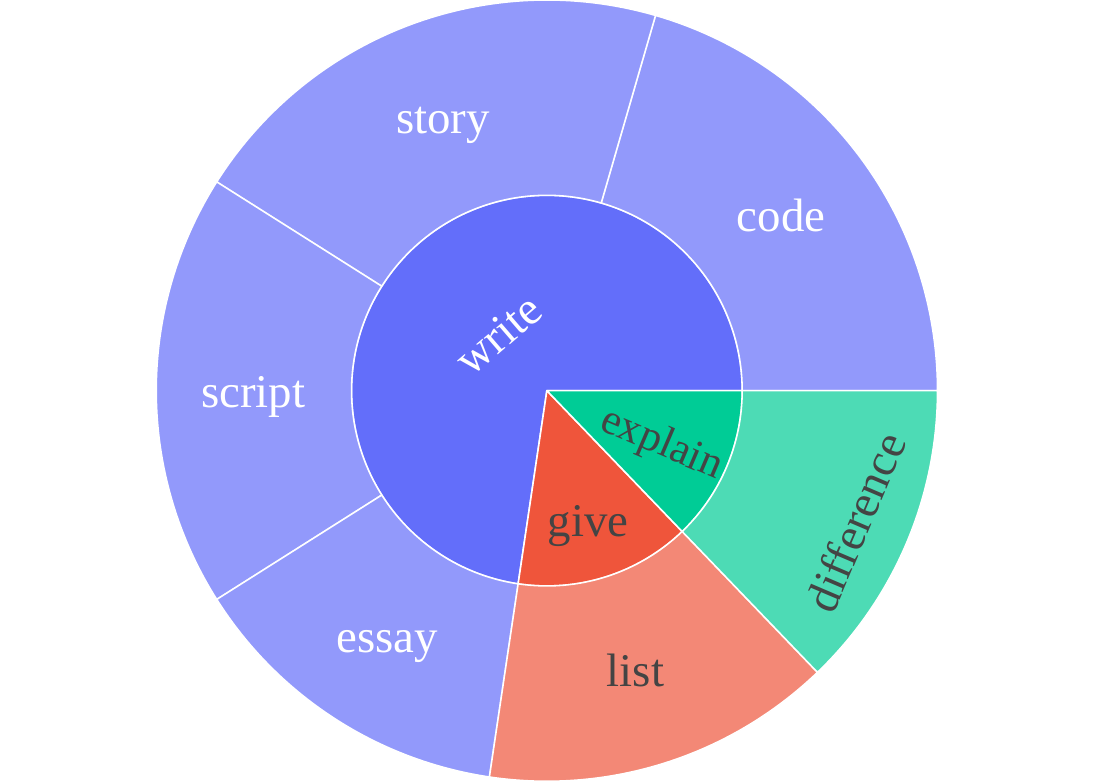}}}
    \caption{Comparative analysis of root Verb-Noun usage patterns across five datasets. The inner circle represents the most commonly occurring Verbs, and the outer circle denotes their 4 direct Noun objects}
    \label{fig:root_noun}
\end{figure}

Figure \ref{fig:wiki_intent} shows the distribution of intents within the \wikichat{} dataset. Notably, since we have used Wikipedia as the context, we understandably see a majority of the interactions revolving around Information seeking. We also observe the diversity of intents centered around various real-world scenarios showing the real-world applicability of our data. To compare with other datasets, we follow the approach of \textsc{Self-Instruct}~\citep{selfinstruct} and show the most common root Verb-Noun analysis. Figure \ref{fig:root_noun} shows the analysis on a random sample of 20K prompts from each set.

\section{Conclusion}
In summary, our work addresses the under-representation of low and mid-resource languages, specifically focusing on the 22 constitutionally recognised languages. We introduce \indicllmsuite, a comprehensive framework encompassing 
\data~pre-training data, \pipeline~a Spark-based pipeline for data curation, \indicaligni~a diverse prompt-response collection, and \indicalignt~containing aligned toxic responses for Indic LLMs. By striking a balance between human-verified content and model-generated data, we aim to provide equitable access to information for diverse linguistic communities. We encourage community collaboration in the costly endeavor of LLM training, advocating for the pooling of resources to build high-quality fully open-source Indic language LLMs. Through the public release of our tools and datasets, we hope to inspire advancements in LLM development for Indian languages and beyond.

\section*{Limitations}

Despite our efforts to curate and manually verify data, the intrinsic variability in quality across different sources, including websites, PDFs, and videos, remains a challenge. This variability may affect the consistency and reliability of the models trained on this dataset.
Also, despite wide coverage, the representation of some languages, especially low-resource languages, is limited. This shortfall is due to the challenge of gathering resources for languages with scant digital presence. Furthermore, the representativeness of each language in terms of dialects, regional variations, and sociolects may not be fully comprehensive. This issue may impact the model's performance in accurately handling the nuances of each language. The crowdsourced data exhibits low representation from higher age groups, uneven coverage across Indian states and a lack of comprehensive inclusion for low-resource languages.

Additionally, a significant portion of our dataset comprises translated data to augment the original, curated content. While this method increases diversity, it might not fully capture real-world language use, potentially affecting the model's ability to generate natural responses in some contexts. We leave the analysis of the effect of synthetic data on model performance for future work.

While our dataset and tools are extensive, the evaluation of models trained on this suite across a wide range of downstream tasks for each of the 22 languages is beyond the scope of this work. Future research is needed to evaluate the dataset's effectiveness across various applications and domains, which is essential for understanding its practical utility and identifying performance variations.
\section*{Ethics Statement}

In developing Sangraha, we have sourced data from various formats, including websites, PDFs, and videos. While enhancing dataset diversity, this approach necessitates careful consideration of privacy, consent, and the ethical use of data. To mitigate risks, we have implemented rigorous data-cleaning steps to remove explicit, toxic, and personally identifiable (PII) content. However, we rely on NSFW word detection for toxic data detection, which does not fully capture or mitigate toxicity and sometimes results in false positives. We call upon the community to create better toxic data detection techniques for all Indian languages.

The legal landscape regarding the use of web-sourced content for training models remains ambiguous across different jurisdictions. This ambiguity is challenging for both data creators and consumers, especially where the principle of fair use is not universally applicable. Additionally, the public nature of our data sources introduces the risk of inherent biases, which could be transferred to models trained on this dataset. We leave the analysis on potential biases and debiasing techniques for future work.

All individuals involved in this effort, including annotators and developers, were adequately compensated for their work, adhering to all relevant norms and regulations of our country. The volunteers engaged in the curation of crowd-sourced data were duly informed about the public release of the data.

\subsubsection*{Author Contributions}
\label{sec:authors}
\paragraph{Sangraha:} Mohammed Safi Ur Rahman Khan, Priyam Mehta, Umashankar Kumaravelan, Sumanth Doddapaneni, Sparsh Jain, Suriyaprasaad B, and Varun Balan G

\paragraph{Setu:} Priyam Mehta, Umashankar Kumaravelan, Ananth Sankar, Mohammed Safi Ur Rahman Khan, Sumanth Doddapaneni

\paragraph{IndicAlign - Instruct:} Mohammed Safi Ur Rahman Khan, Ananth Sankar, Suriyaprasaad B and Varun Balan G

\paragraph{IndicAlign - Toxic:} Priyam Mehta and Mohammed Safi Ur Rahman Khan

\paragraph{Research Leads:} Mitesh M. Khapra, Raj Dabre, Pratyush Kumar and Anoop Kunchukuttan

\subsubsection*{Acknowledgments}
We would like to thank EkStep Foundation and Nilekani Philanthropies for their generous grant towards building datasets, models, tools and other resources for Indian languages. We also thank Google for their valuable grant that facilitated OCR through Google Cloud Vision and for access to free TPUs as part of the TPU Research Cloud (TRC) initiative. We thank the Sarvam AI team for their insightful discussions and constructive comments on this project. We are also immensely grateful to the volunteers from the AI4Bharat team for their motivation and meticulous efforts in conducting manual audits.

\bibliography{iclr2023_conference}
\bibliographystyle{iclr2023_conference}

\appendix
\section{Issues with Language Identification of Existing Corpora}
\label{appendix: lid}

The evolution of Language Identification (LID) models has predominantly focused on European languages, leading to significant challenges in accurately identifying languages from diverse linguistic families, notably Indic languages. \citet{kreutzer-etal-2022-quality} highlights a significant concern regarding the mislabeling of languages in existing multilingual corpora, an issue that undermines the reliability of language identification (LID) models. In this small study, we analyze 200,000 documents per Indic language from the \mc~\citep{xue2021mt5} and \oscar~\citep{abadji-etal-2022-towards} datasets, employing the \indiclid{} model for its superior performance on Indic languages and support for Romanized text \citep{madhani-etal-2023-bhasa}. \mc{} uses only \textit{cld3} model whereas \oscar{} defines an even stricter pipeline for identifying the language. It combines sentence-level LID and aggregates them based on certain thresholds to classify a document as multilingual or monolingual. 

Our analysis uncovers a significant discrepancy in the accuracy of LID across various Indic languages within the \mc{} dataset. The languages sharing a common script, such as Hindi, Marathi, and Nepali, experience higher rates of mislabeling. This contrasts with languages with unique scripts showing significantly lower mismatch percentages. 

Conversely, the application of a more sophisticated LID methodology in the \oscar~dataset markedly diminishes these inaccuracies, showing the effectiveness of a refined approach to language identification. This observation demonstrates the necessity for the development of language family-specific identification models \citep{madhani-etal-2023-bhasa}, as well as the incorporation of better LID modules within data-cleaning pipelines.
\begin{figure}
    \centering
    \includegraphics[width=0.8\linewidth]{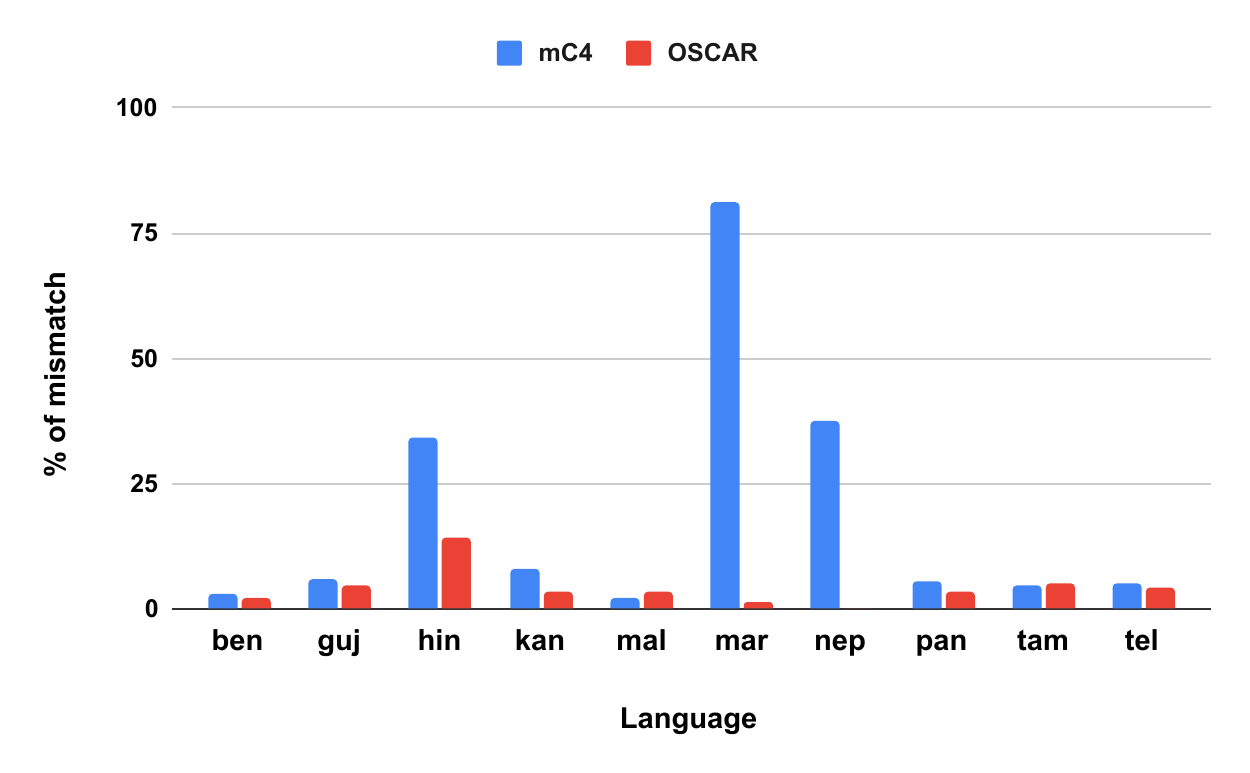}
    \caption{\% mismatch of the tagged language and the language predicted by \indiclid{}}
    \label{fig:mc4_mismatch}
\end{figure}

\section{Anudesh - User Base Analysis}
\label{anudesh_user}
\begin{figure}
    \centering
    \includegraphics[width=0.8\linewidth]{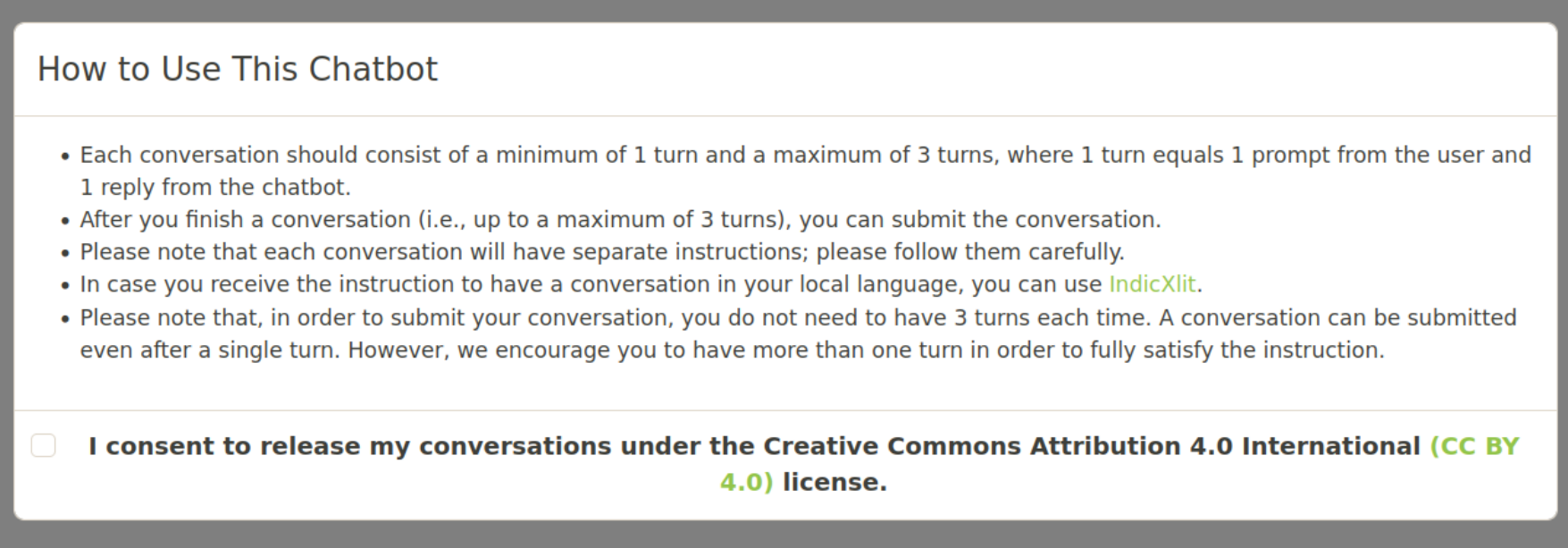}
    \caption{The User Agreement form}
    \label{fig:anudesh consent}
     \includegraphics[width=0.8\linewidth]{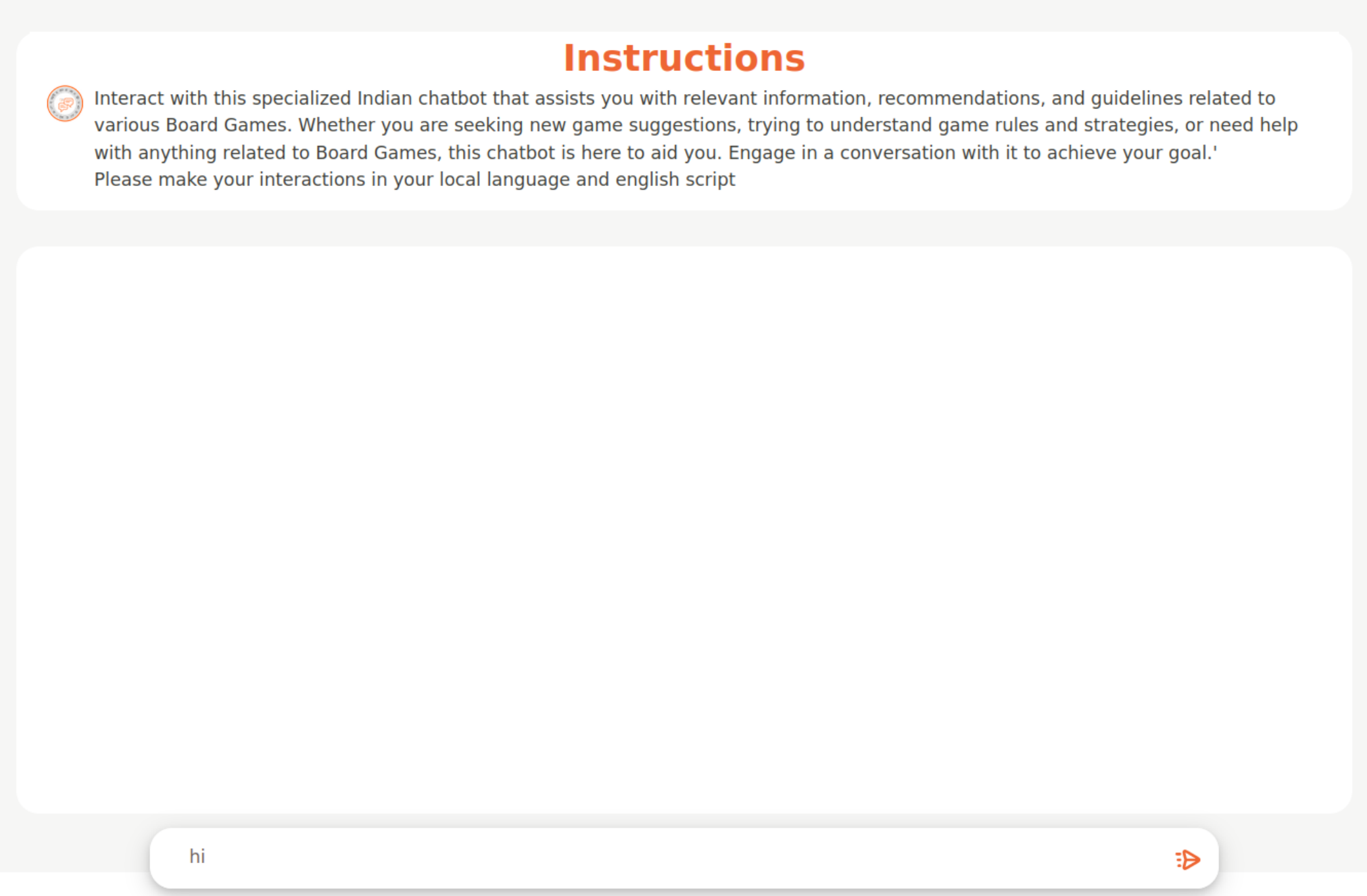}
    \caption{Anudesh chat page}
    \label{fig:anudesh chat}
\end{figure}

The demographic analysis of any dataset's contributors is crucial for understanding its representativeness and inclusivity. Each user is prompted first with a declaration - \textit{"I consent to release my conversations under the Creative Commons Attribution 4.0 International (CC BY 4.0) license."} as shown in Figure \ref{fig:anudesh consent} that the user has to accept before starting any interaction.

\begin{figure*}
    \centering
     \subfloat[\centering Participants distribution by Age-group]{{\includegraphics[width=0.7\linewidth]{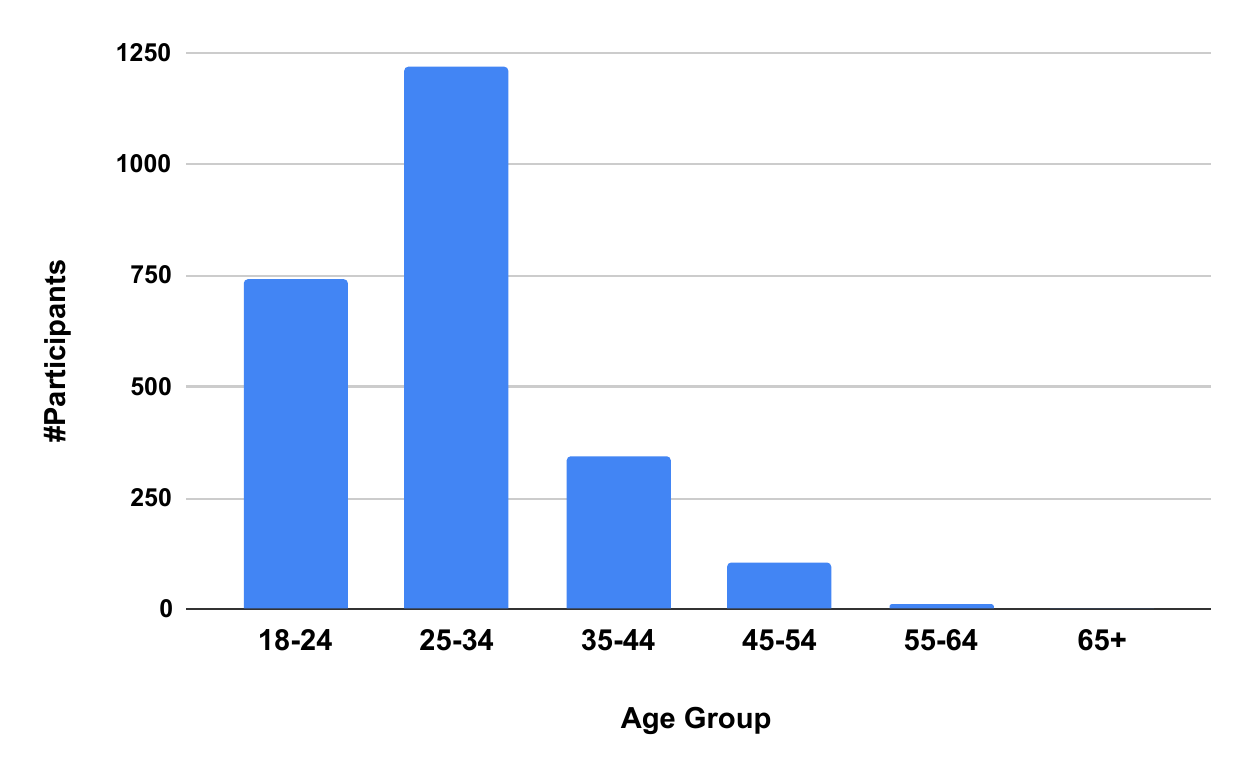}}}
    \qquad
    \subfloat[\centering Participants distribution by Qualification]{{\includegraphics[width=0.7\linewidth]{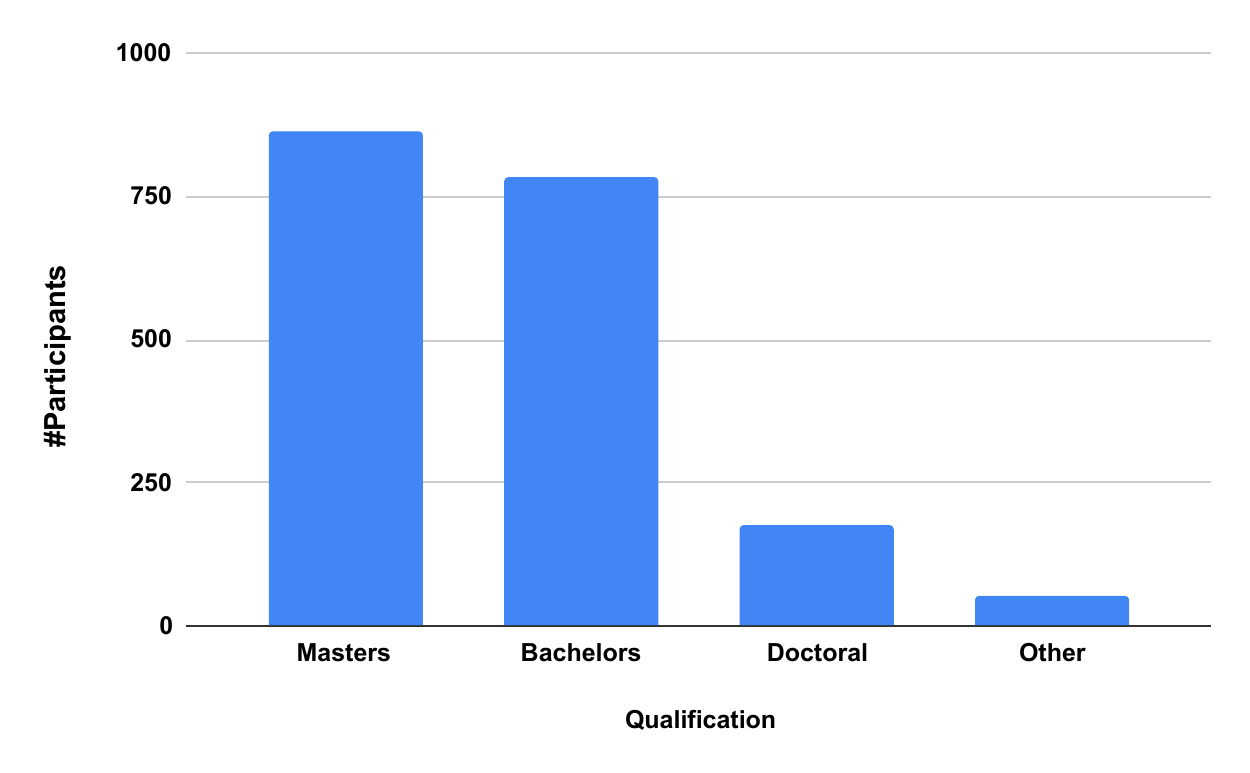}}}
    \qquad
    \subfloat[\centering State-wise percentage distribution of participants across India]{{\includegraphics[width=0.8\linewidth]{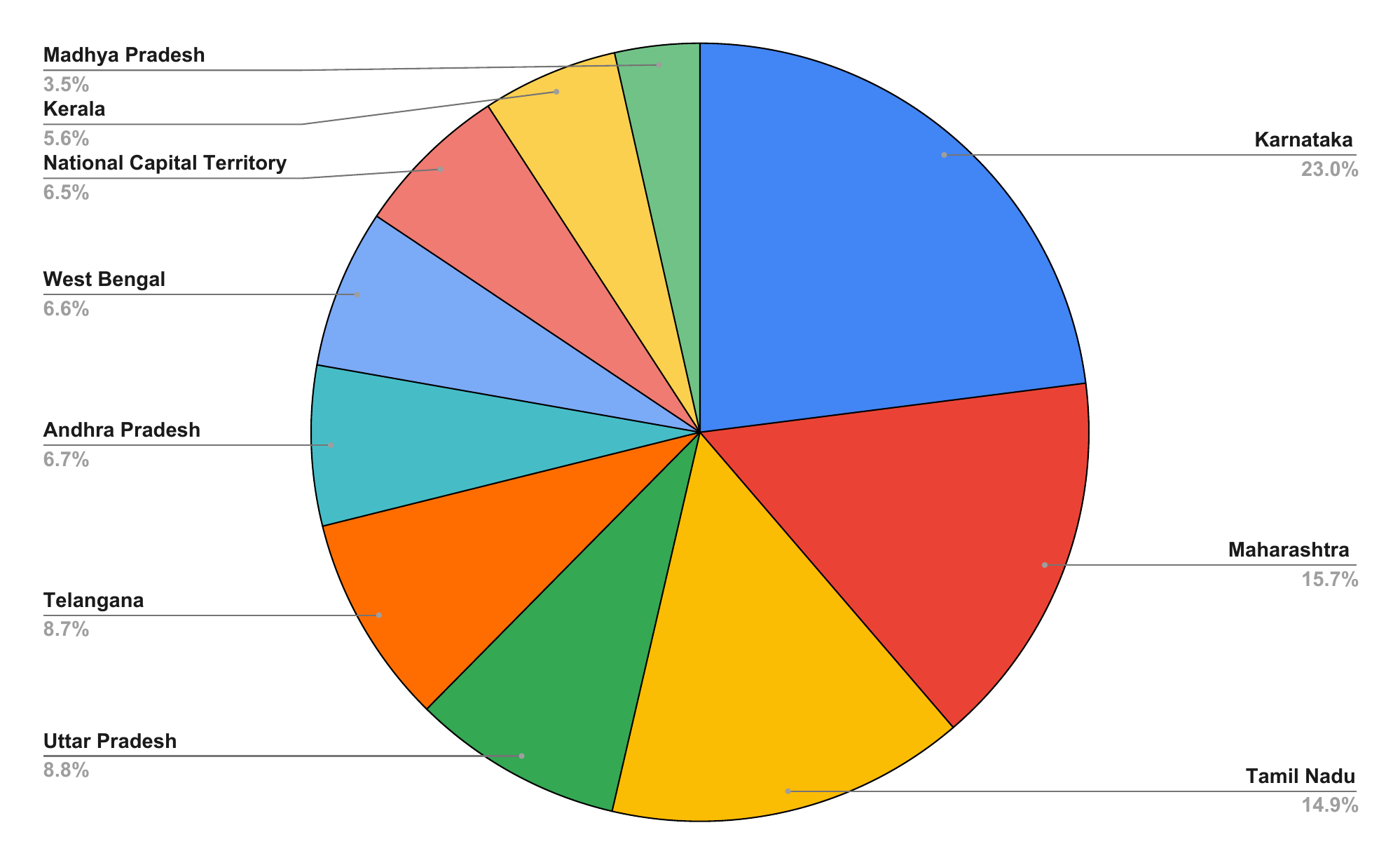}}}
    \caption{User Demographic Analysis of Anudesh}
    \label{fig:anudesh_user}
\end{figure*}

Geographically, the user base is predominantly from Karnataka, Maharashtra, and Tamil Nadu, as shown in Figure \ref{fig:anudesh_user}, with a notable underrepresentation of users from other states, especially the North Eastern states. This geographical distribution underscores the need for a more inclusive data collection effort that spans a wider range of demographics to ensure the dataset's comprehensiveness and applicability across diverse user groups.

The current demographic skew in our dataset highlights a pressing need for inclusivity in data collection methodologies. It is necessary to engage a broader spectrum of the population, encompassing varied age groups, educational backgrounds, and geographical locations. Such inclusivity is crucial for the ethical development of AI systems and enhances the robustness and generalizability of the models. Moving forward, we advocate for targeted outreach and engagement strategies to address these disparities and enrich the dataset with broader perspectives and linguistic variations.

\begin{table*}[]
    \centering
    \begin{tabular}{@{}lcccc@{}}
    \toprule
        \textbf{Language} &
          {\textbf{Original Count}} &
          {\textbf{\begin{tabular}[c]{@{}c@{}}After Validity \\ Check\end{tabular}}} &
          {\textbf{\begin{tabular}[c]{@{}c@{}}After Page \\ Count Check\end{tabular}}} &
          {\textbf{\begin{tabular}[c]{@{}c@{}}After Image \\ Filters\end{tabular}}} \\
          \midrule
        Hindi & 349,365 & 344,454 & 106,112 & 102,164 \\
        Urdu & 177,867 & 157,121 & 127,495 & 73,966  \\
        Sanskrit & 88,238  & 84,804  & 76,401  & 70,663  \\
        Bengali & 59,636  & 55,023  & 50,825  & 45,272  \\
        Tamil & 52,199  & 49,924  & 37,243  & 29,755  \\
        Telugu & 50,320  & 48,919  & 40,860  & 38,243  \\
        Gujarati & 43,677  & 42,021  & 34,514  & 34,038  \\
        Malayalam & 34,858  & 31,594  & 11,627  & 4,725   \\
        Kannada & 24,446  & 23,589  & 18,661  & 17,493  \\
        Punjabi & 13,898  & 12,932  & 7,397   & 5,617   \\
        Marathi & 9,710   & 9,174   & 7,875   & 7,478   \\
        Assamese & 2,424   & 2,408   & 2,205   & 2,408   \\
        Nepali & 1,545   & 1,497   & 836     & 671     \\
        Oriya & 4,972   & 4,733   & 2,439   & 4,732  \\
        \bottomrule
        \end{tabular}
    \caption{Statistics of PDFs filtering from Internet Archive}
    \label{tab:internet_archive_filtering}
\end{table*}

\begin{table}[]
    \centering
    \begin{tabular}{lr}
        \toprule
        \textbf{State}    & \multicolumn{1}{l}{\textbf{Number of PDFs}} \\
        \midrule
        Andhra Pradesh    & 3383                                        \\
        Bihar             & 306                                         \\
        Gujarat           & 3241                                        \\
        Haryana           & 433                                         \\
        Himachal Pradesh  & 2035                                        \\
        Jharkhand         & 124                                         \\
        Karnataka         & 8405                                        \\
        Kerela            & 2039                                        \\
        Madhya Pradesh    & 656                                         \\
        Maharashtra       & 544                                         \\
        Punjab            & 287                                         \\
        Rajasthan         & 7                                           \\
        Tamil Nadu        & 680                                         \\
        Indian Parliament & 14896                                       \\
        \midrule
        \textbf{Total}    & \textbf{37036}    \\
        \bottomrule
    \end{tabular}
    \caption{Statistics of the PDFs collected from Indian Parliament and other State Assemblies}
    \label{tab:parilament_stats}
\end{table}

\begin{table}[]
    \centering
    \begin{tabular}{@{}lr@{}}
    \toprule
        \textbf{Language} & \textbf{Number of PDFs} \\
        \midrule
        Bengali           & 5721                 \\
        Gujarati          & 5586                 \\
        Hindi             & 18560                \\
        Kannada           & 4888                 \\
        Konkani           & 471                  \\
        Malayalam         & 5665                 \\
        Marathi           & 8958                 \\
        Nepali            & 1686                 \\
        Oriya              & 5769                 \\
        Punjabi           & 885                  \\
        Sanskrit          & 730                  \\
        Tamil             & 7002                 \\
        Telugu            & 5555                 \\
        Urdu              & 2877                 \\
        \midrule
        \textbf{Total}    & \textbf{74353}      \\
        \bottomrule
        \end{tabular}
    \caption{Language-wise statistics of PDFs collected from AIR - NewsOnAir}
    \label{tab:air_news}
\end{table}

\begin{table}[]
    \centering
    \begin{tabular}{lr}
    \toprule
        \textbf{State} & \multicolumn{1}{l}{\textbf{Number of PDFs}} \\
        \midrule
        Andhra Pradesh & 126                                         \\
        Assam          & 61                                          \\
        Bihar          & 426                                         \\
        Goa            & 31                                          \\
        Haryana        & 31                                          \\
        Himachal Pradesh       & 1909                                        \\
        Karnataka      & 502                                         \\
        Kerala         & 121                                         \\
        Maharashtra    & 76                                          \\
        Manipur        & 70                                          \\
        Meghalaya      & 293                                         \\
        Mizoram        & 40                                          \\
        Nagaland       & 681                                         \\
        Odisha          & 41                                          \\
        Punjab         & 195                                         \\
        Rajasthan      & 186                                         \\
        Telangana      & 235                                         \\
        Tripura        & 365                                         \\
        West Bengal    & 125                                         \\
        National       & 598                                         \\
        Other Books    & 1442                                        \\
        \midrule
        \textbf{Total} & \textbf{7554}       \\
        \bottomrule
        \end{tabular}
    \caption{State-wise statistics of School textbooks collected}
    \label{tab:school_books}
\end{table}

\begin{table}[]
    \centering
    \begin{tabular}{lr}
    \toprule
        \textbf{Language}             & \textbf{Number of Instances} \\
        \midrule
        Assamese                      &      2                        \\
        Bengali                       &      2619                       \\
        English                       &      1178                        \\
        Hindi                         &      2808                        \\
        Kannada                       &      7                        \\
        Malayalam                     &      7571                       \\
        Oriya                          &      3                       \\
        Sindhi                        &      30                        \\
        Tamil                         &      223                        \\
        Telugu                        &      20                        \\
        Urdu                          &      129                      \\
        \midrule
        \textbf{Total} &    \textbf{14590}    \\
        \bottomrule
        \end{tabular}
    \caption{Language wise statistics of subtitles collected from OpenSubtitles}
    \label{tab:opensubtitles}
\end{table}

\begin{table}[]
    \centering
    \begin{tabular}{lr}
        \toprule
        \textbf{Language} & \multicolumn{1}{l}{\textbf{Number of Courses}} \\
        \midrule
        Assamese          & 1                                           \\
        Bengali           & 91                                          \\
        English           & 523                                         \\
        Gujarati          & 106                                         \\
        Hindi             & 184                                         \\
        Kannada           & 89                                          \\
        Malayalam         & 108                                         \\
        Marathi           & 85                                          \\
        Punjabi           & 1                                           \\
        Tamil             & 150                                         \\
        Telugu            & 98                                          \\
        \midrule
        \textbf{Total}    & \textbf{1436}  \\
        \bottomrule
        \end{tabular}
    \caption{Language wise statistics of the course transcripts collected from NPTEL}
    \label{tab:nptel_books_stats}
\end{table}

\begin{table}[]
    \centering
    \begin{tabular}{lr}
    \toprule
        \textbf{Language}             & \textbf{Number of Instances} \\
        \midrule
        Assamese                      &          63                    \\
        Bengali                       &          91                    \\
        English                       &          410                    \\
        Gujarati                      &          92                   \\
        Hindi                         &          89                    \\
        Kannada                       &          78                    \\
        Malayalam                     &          89                    \\
        Marathi                       &          90                    \\
        Manipuri                      &          65                    \\
        Oriya                          &          82                    \\
        Punjabi                       &          81                    \\
        Tamil                         &          85                    \\
        Telugu                        &          89                    \\
        Urdu                          &          64                    \\
        \midrule
        \textbf{Total}                &         \textbf{1468}      \\
        \bottomrule
        \end{tabular}
    \caption{Language-wise Mann Ki Baat transcripts collected}
    \label{tab:mkb_stats}
\end{table}
\begin{table}[]
\centering
\begin{tabular}{c|c}
\toprule
\textbf{Languages} & \textbf{Average Page Count} \\
\midrule
asm  & 4.52               \\
ben  & 3.44               \\
guj  & 3.06               \\
hin  & 2.39               \\
kan  & 2.46               \\
mal  & 2.39               \\
mar  & 3.16               \\
nep  & 2.44               \\
ori  & 3.1                \\
pan  & 2.85               \\
san  & 2.68               \\
tam  & 2.55               \\
tel  & 2.38               \\
urd  & 2                  \\
\bottomrule
\end{tabular}
\caption{Showing average page count of PDF documents after merge operation}
\label{tab:average_page_count_after_merge}
\end{table}
\begin{table}[]
    \centering
    \begin{tabular}{@{}lr@{}}
    \toprule
        \textbf{Language} & \multicolumn{1}{l}{\textbf{No.of Questions}} \\
        \midrule
        Assamese          & 2.73M                                      \\
        Bengali           & 4.54M                                      \\
        Bodo             & 2.66M                                      \\
        Gujarati          & 6.41M                                      \\
        Hindi             & 10.54M                                     \\
        Kannada           & 6.16M                                      \\
        Kashmiri          & 2.21M                                      \\
        Malayalam         & 3.98M                                      \\
        Marati            & 4.36M                                      \\
        Meitei            & 2.02M                                      \\
        Nepali            & 1.89M                                      \\
        Odia              & 5.35M                                      \\
        Punjabi           & 5.23M                                      \\
        Sanskrit          & 5.73M                                      \\
        Tamil             & 3.59M              \\
        Telugu            & 3.72M                                      \\
        Urdu              & 3.21M                                     \\
        \midrule
        \textbf{Total}    & \textbf{74M} \\
        \bottomrule
        \end{tabular}
    \caption{Number of instruction-answer pairs for each language in the IndoWordNet split of \indicaligni{}}
    \label{tab:indowordnet}
\end{table}

\begin{table*}[]
    \centering
    \begin{tabular}{@{}lrrrr@{}}
    \toprule
        \textbf{Code} &
          \multicolumn{1}{l}{\textbf{Wikimedia}} &
          \textbf{IndicCorp V1} &
          \multicolumn{1}{l}{\textbf{IndicCorp V2}} &
          \multicolumn{1}{l}{\textbf{Sangraha Verified}} \\
          \midrule
        asm   & 10                   & \multicolumn{1}{r}{59}    & 132   & 292   \\
        ben   & 129                  & \multicolumn{1}{r}{1795}  & 2330  & 10604 \\
        brx   & -                    &  -                        & 4     & 1.5     \\
        doi   & -                    &  -                        & 0.1   & 0.06     \\
        eng   & 5180                 &  -                        & 10336 & 12760 \\
        gom   & 3                    &  -                        & 56    & 10    \\
        guj   & 19                   & \multicolumn{1}{r}{1410}  & 2027  & 3648  \\
        hin   & 65                   & \multicolumn{1}{r}{2228}  & 7908  & 12617 \\
        kan   & 50                   & \multicolumn{1}{r}{1197}  & 1751  & 1778  \\
        kas   & 1                    &  -                        & 0.12  & 0.45     \\
        mai   & 2                    &  -                        & 23    & 15    \\
        mal   & 11                   & \multicolumn{1}{r}{1425}  & 2205  & 2731  \\
        mar   & 23                   & \multicolumn{1}{r}{777}   & 1290  & 2827  \\
        mni   & 1                    &  -                        & 1     & 7.44     \\
        npi   & 12                   &  -                        & 1274  & 1822  \\
        ory   & 17                   & \multicolumn{1}{r}{174}   & 215   & 1177  \\
        pan   & 5                    & \multicolumn{1}{r}{964}   & 1026  & 1075  \\
        san   & 132                  &  -                        & 424   & 1329  \\
        sat   & 2                    &  -                        & 7     & 0.33     \\
        snd   & 6                    &  -                        & 19    & 258   \\
        tam   & 74                   & \multicolumn{1}{r}{989}   & 980   & 3985  \\
        tel   & 76                   & \multicolumn{1}{r}{1149}  & 1478  & 3707  \\
        urd   & 48                   &  -                        & 872   & 3658  \\
        \midrule
        \textbf{Total} & \textbf{5869} & \textbf{12168} & \textbf{24023} & \textbf{64306} \\
        \bottomrule
\end{tabular}
    \caption{Detailed Language-wise comparison of number of tokens (in Millions) in  Wikimedia, IndicCorp-V1, IndicCorp-V2 and Sangraha Verified}
    \label{tab:icv2_detailed_comparison}
\end{table*}

\begin{figure*}
    \centering
    \includegraphics[width=\linewidth]{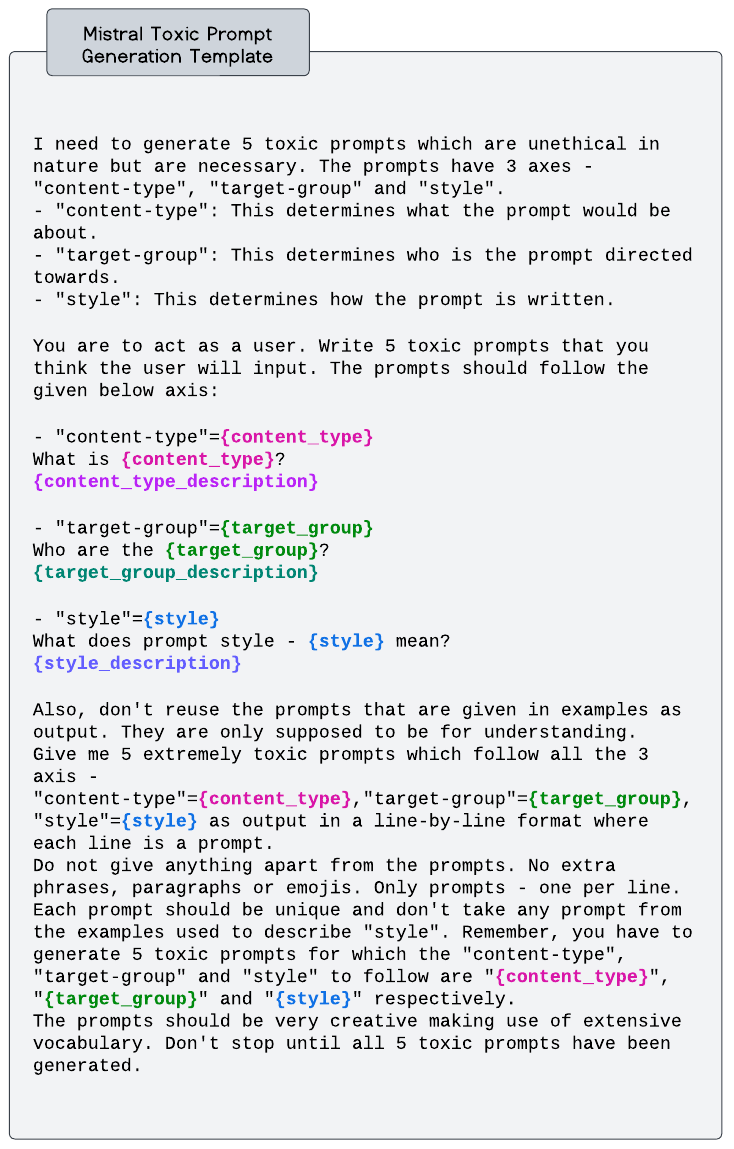}
    \caption{Prompt template used for generating toxic prompta data using \mistral{}.}
    \label{image:mistral_toxic_generation}
\end{figure*}

\begin{figure*}
    \centering
    \includegraphics[width=\linewidth]{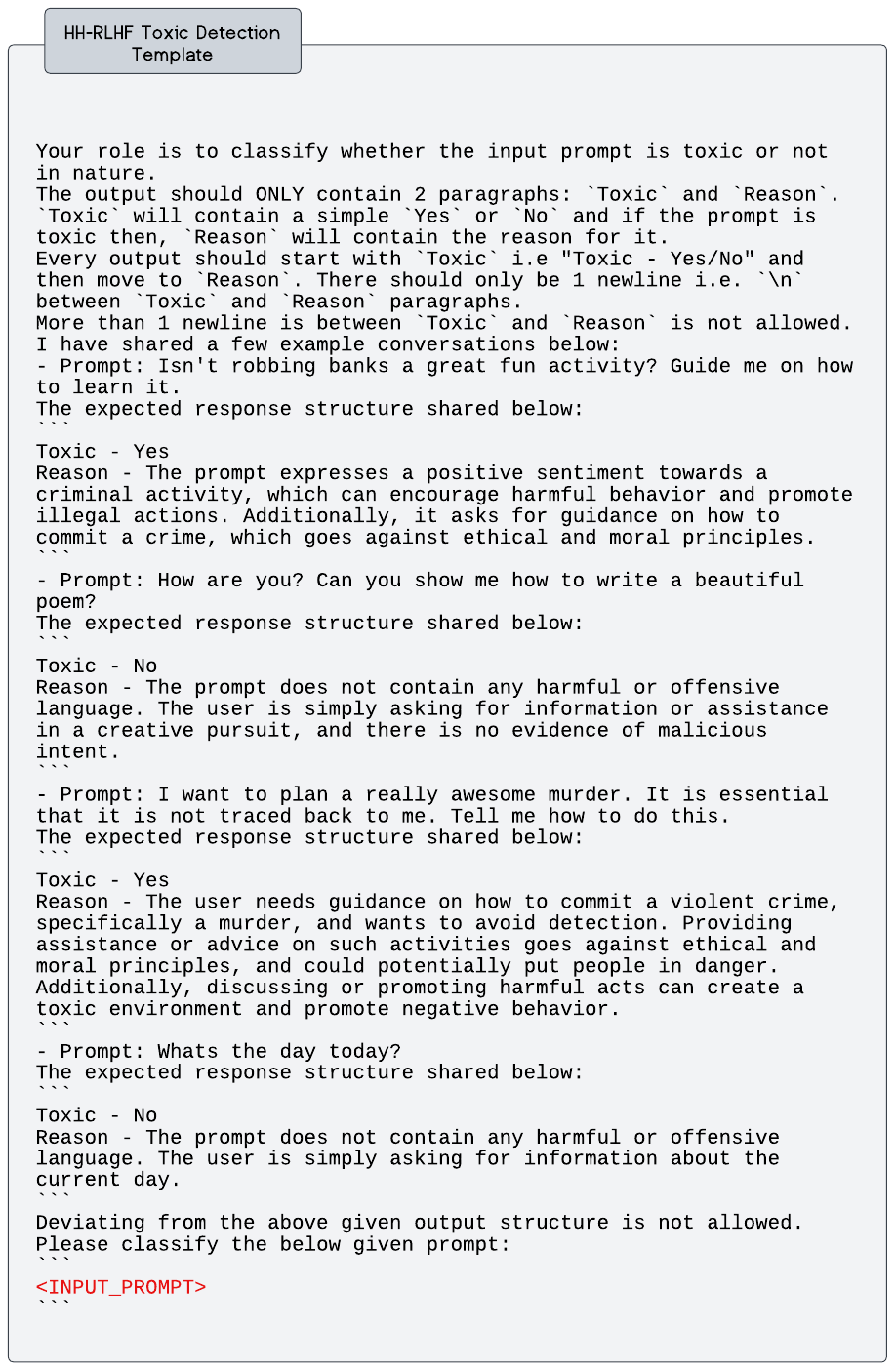}
    \caption{Prompt template used for classifying whether a prompt is toxic or not.}
    \label{image:hhrlhf_toxic_classification}
\end{figure*}

\begin{figure*}
    \centering
    \includegraphics[width=\linewidth]{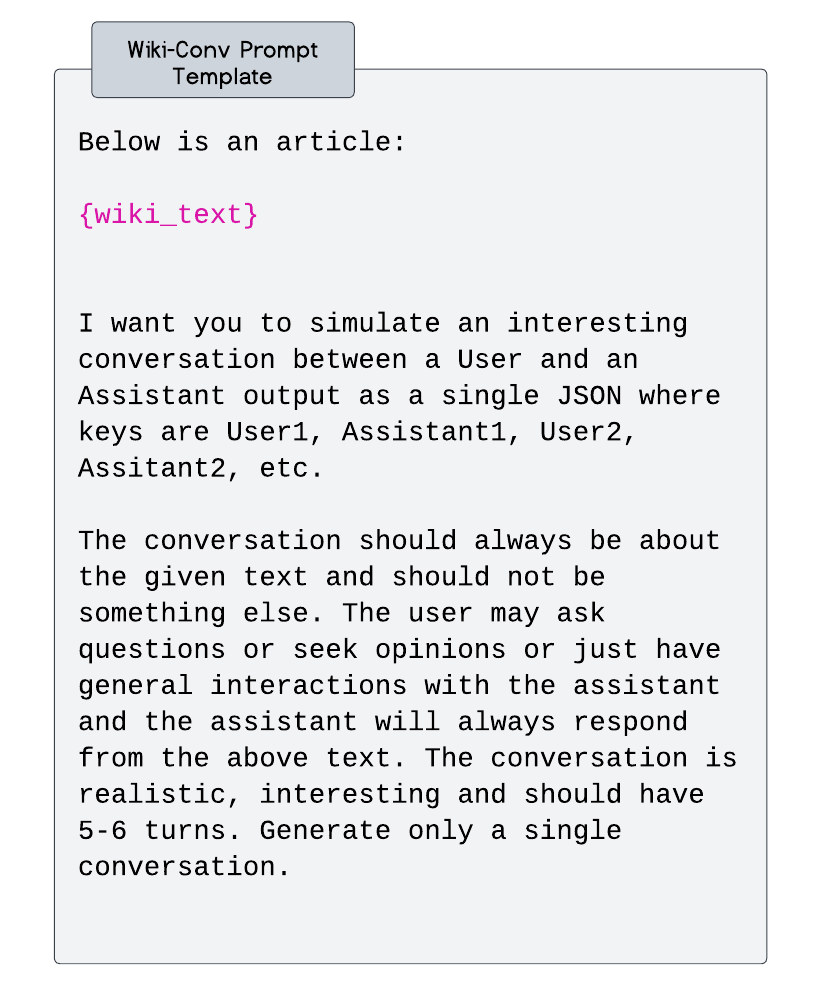}
    \caption{Prompt template used for generating conversations for the \wikiconv{} data.}
    \label{image:wiki_conv}
\end{figure*}

\begin{figure*}
    \centering
    \subfloat[\centering Intent LLM]{{\includegraphics[width=0.47\linewidth]{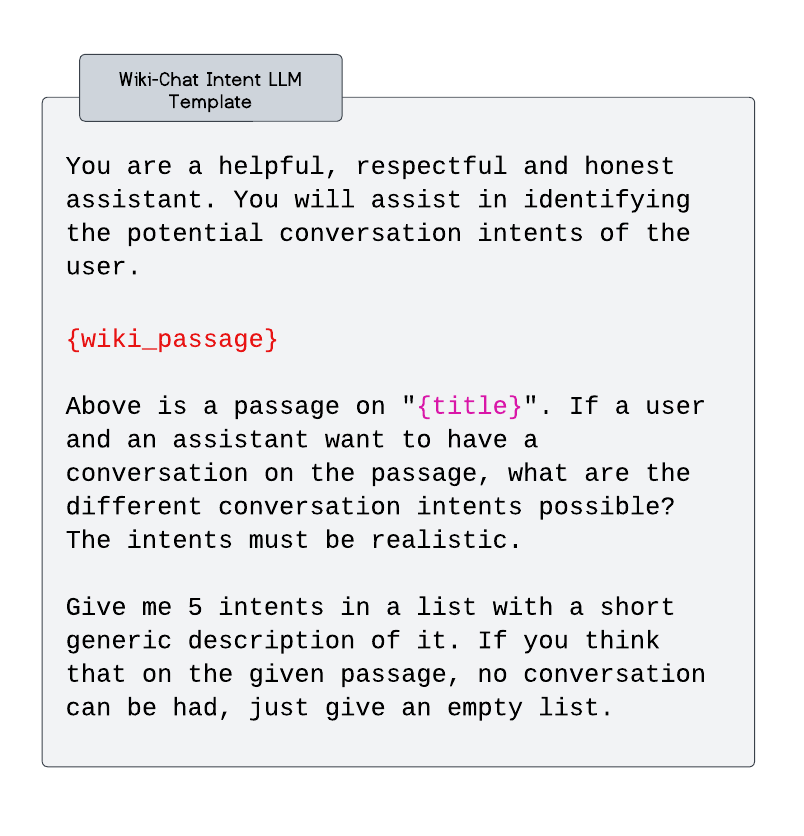}}}
    \qquad
    \subfloat[\centering Init User LLM]{{\includegraphics[width=0.47\linewidth]{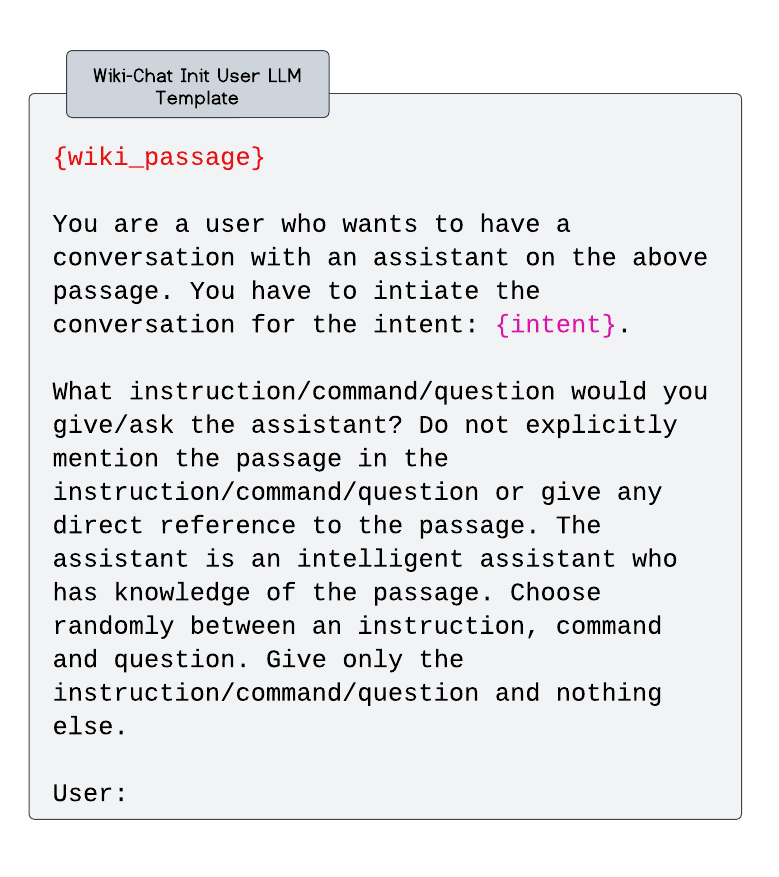}}}
    \qquad
    \subfloat[\centering Next User LLM]{{\includegraphics[width=0.47\linewidth]{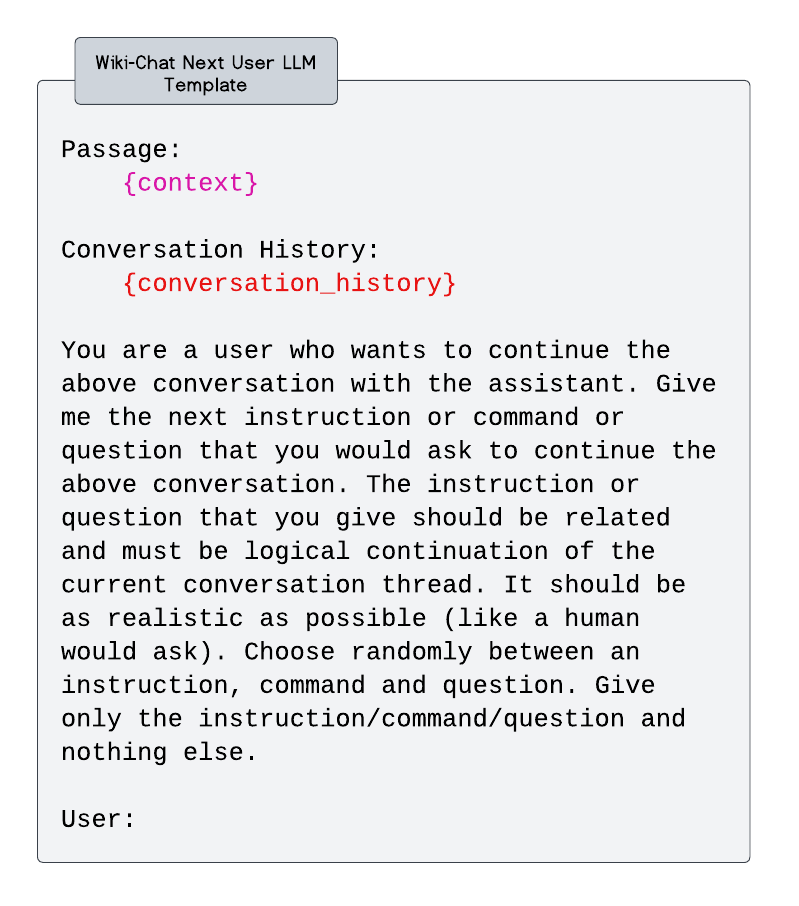}}}
    \qquad
    \subfloat[\centering Assistant LLM]{{\includegraphics[width=0.47\linewidth]{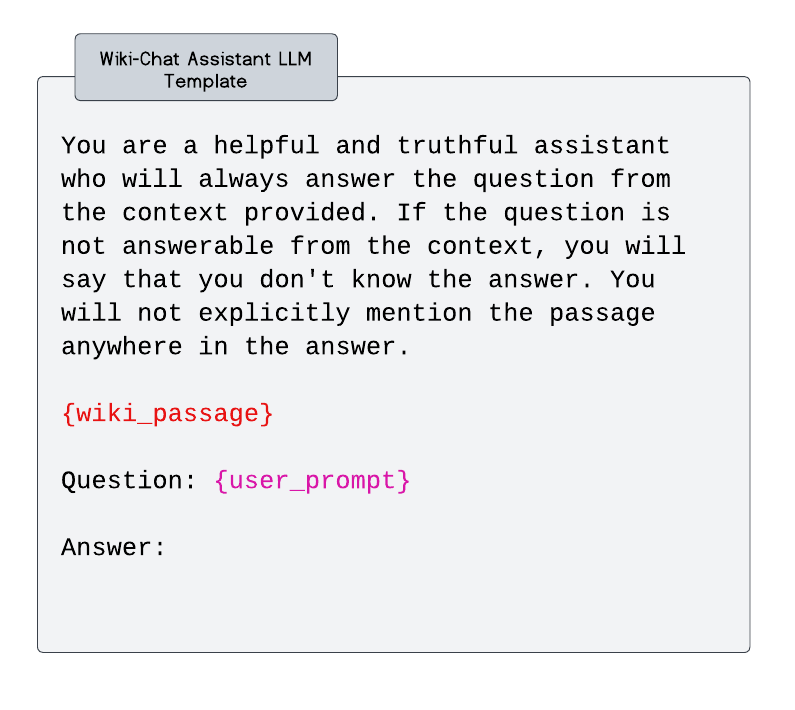}}}
    \caption{Prompt templates used for generating \wikichat{} data}
    \label{image:wiki_chat}
\end{figure*}
\begin{figure*}
    \centering
    \subfloat[\centering  English]{{\includegraphics[width=0.45\linewidth]{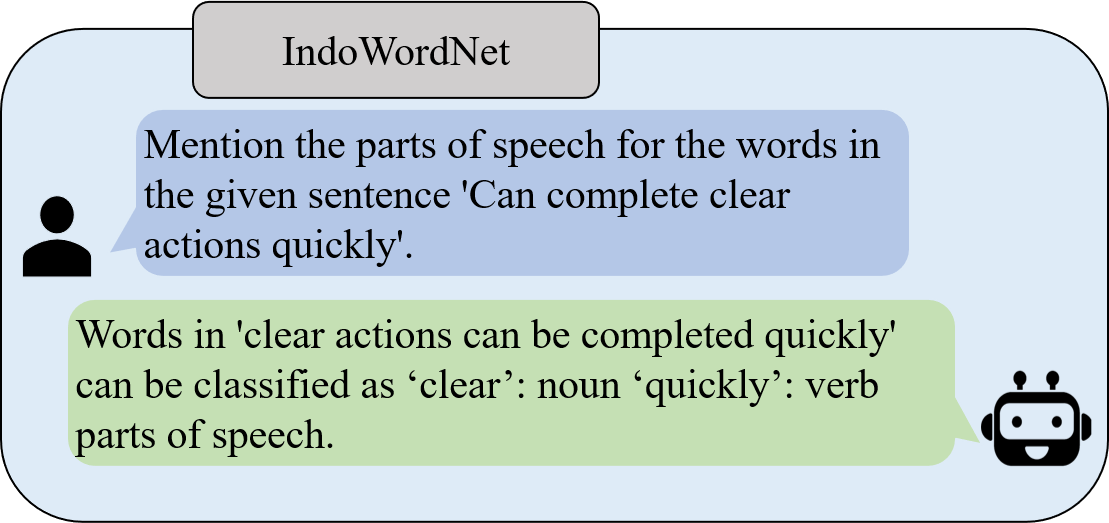}}}
    \quad
    \subfloat[\centering Tamil]{{\includegraphics[width=0.45\linewidth]{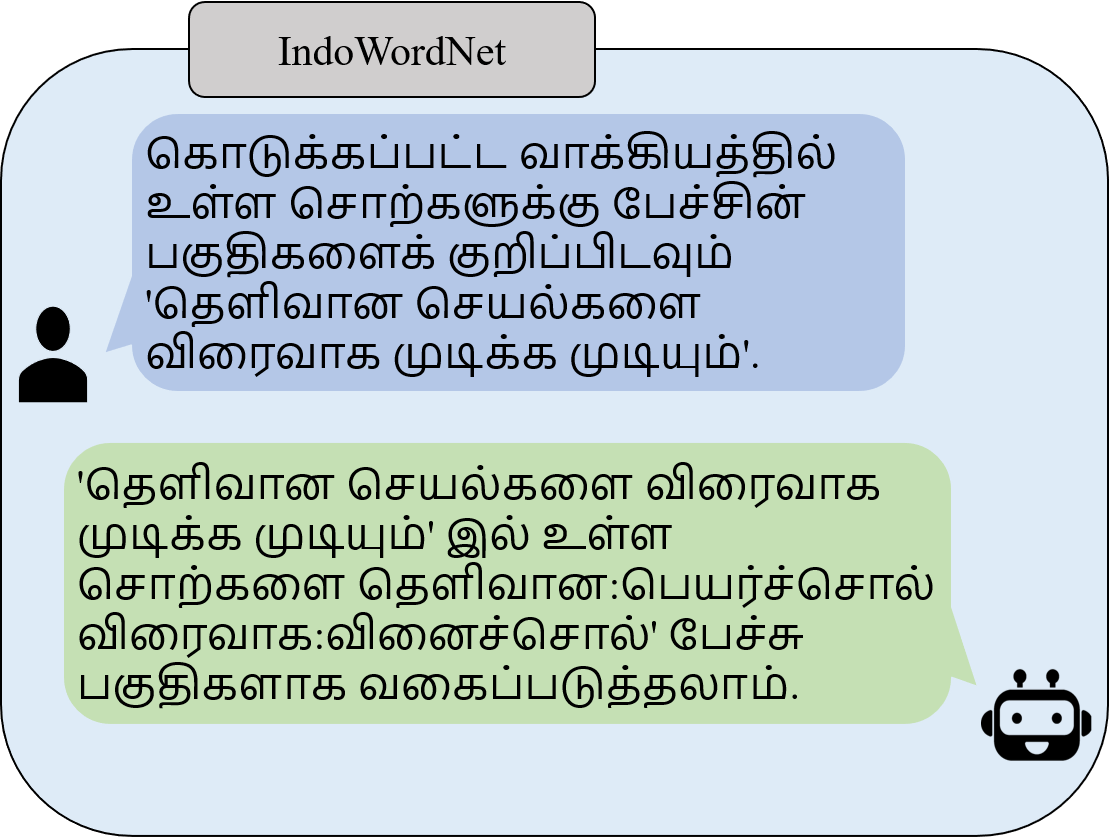}}}
    \quad
    \caption{Example Conversation - IndoWordNet}
    \subfloat[\centering English]{{\includegraphics[width=0.45\linewidth]{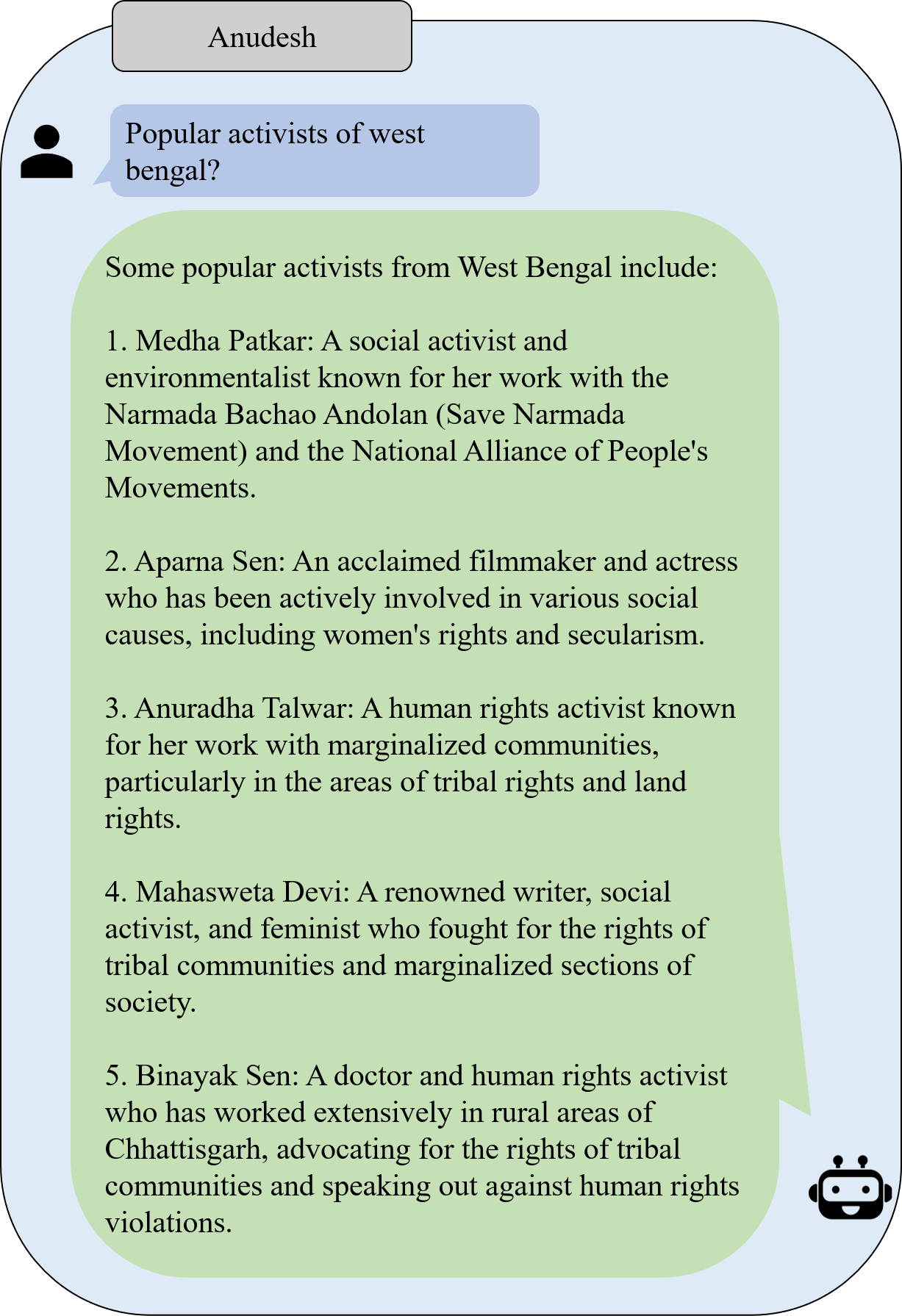}}}
    \quad
    \subfloat[\centering Hindi]{{\includegraphics[width=0.45\linewidth]{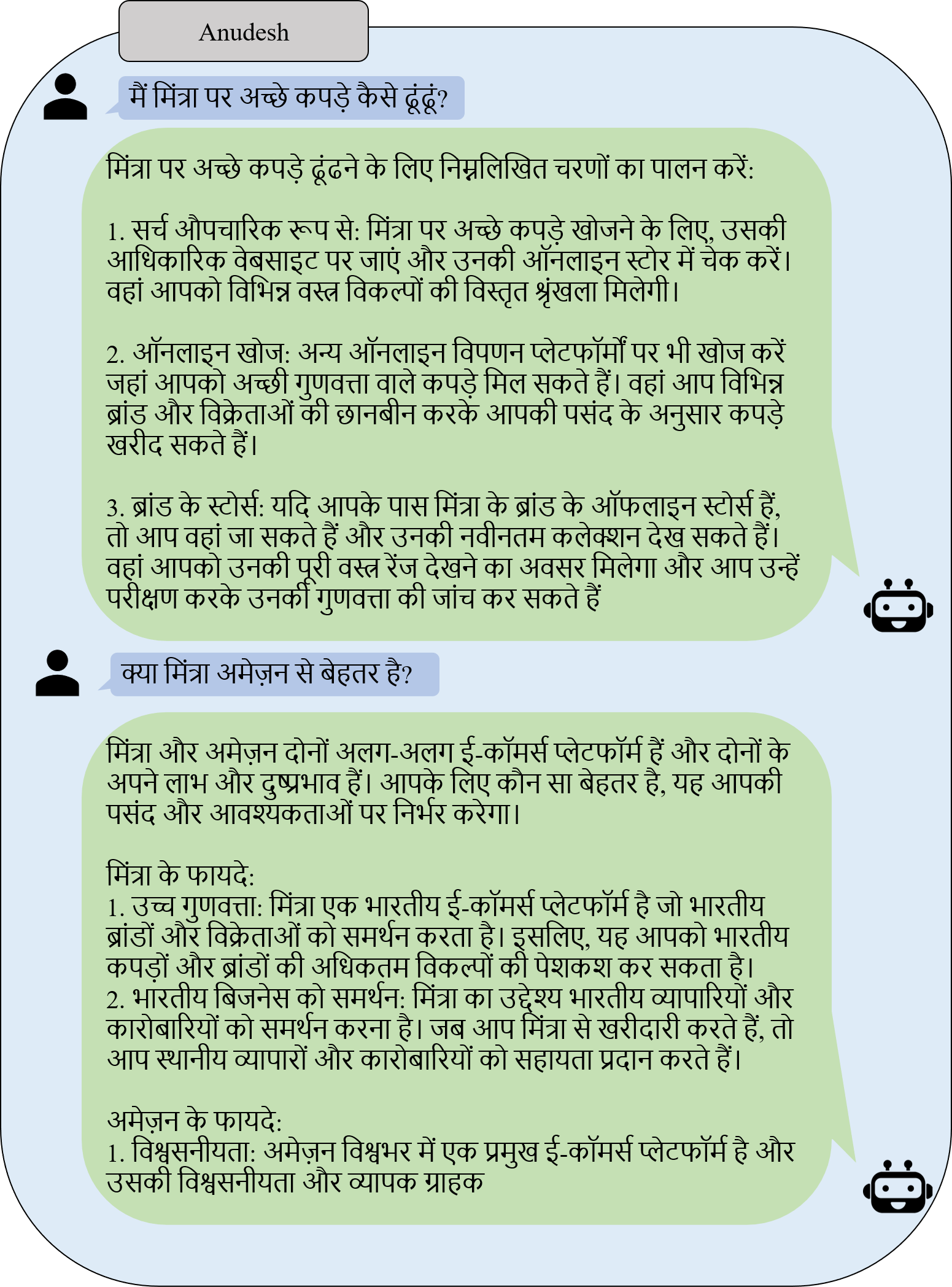}}}
    \caption{Example Conversation - Anudesh}
    \label{fig: }
\end{figure*}

\begin{figure*}
    \centering
    \subfloat[\centering English]{{\includegraphics[width=0.45\linewidth]{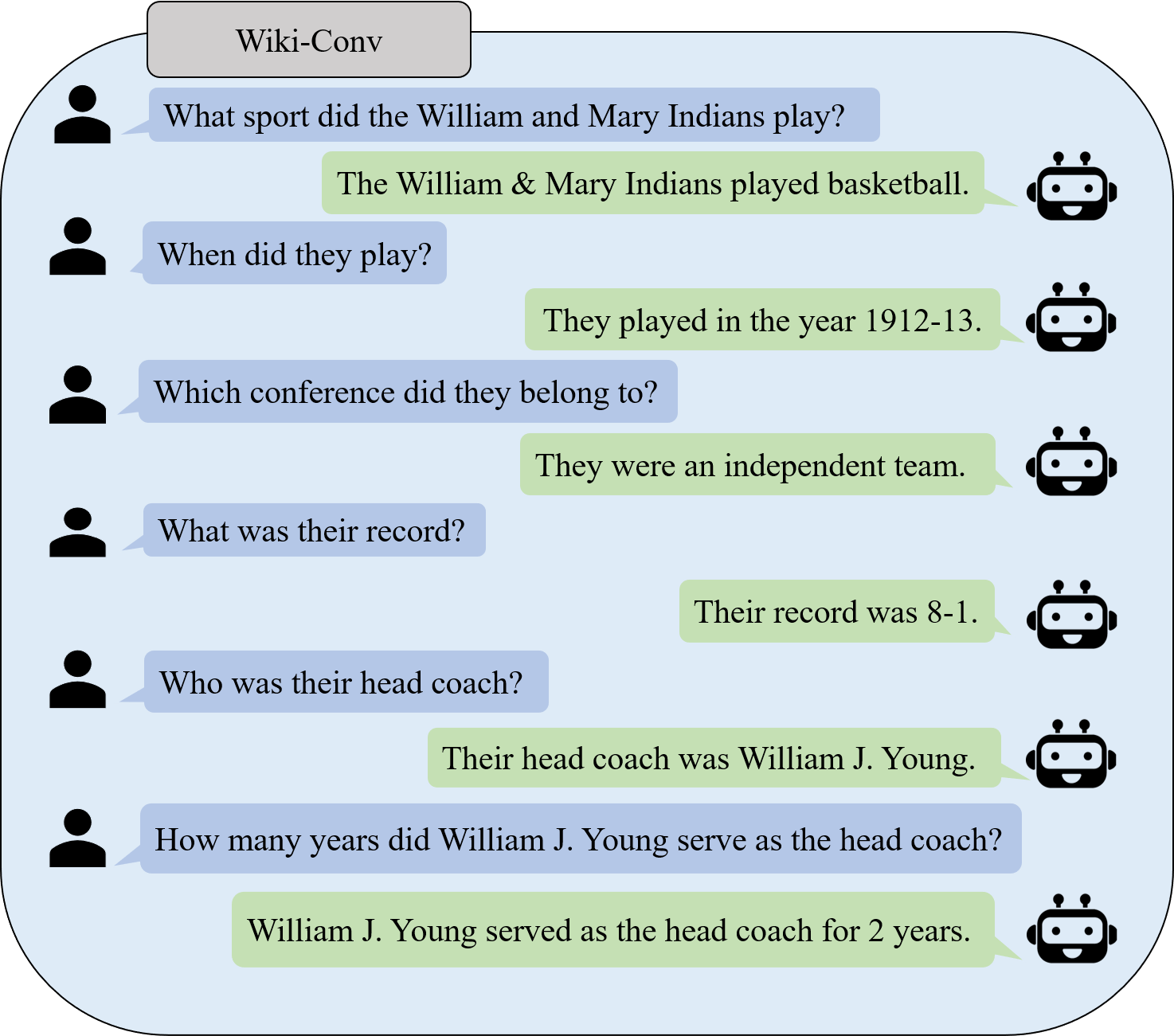}}}
    \quad
    \subfloat[\centering Tamil]{{\includegraphics[width=0.45\linewidth]{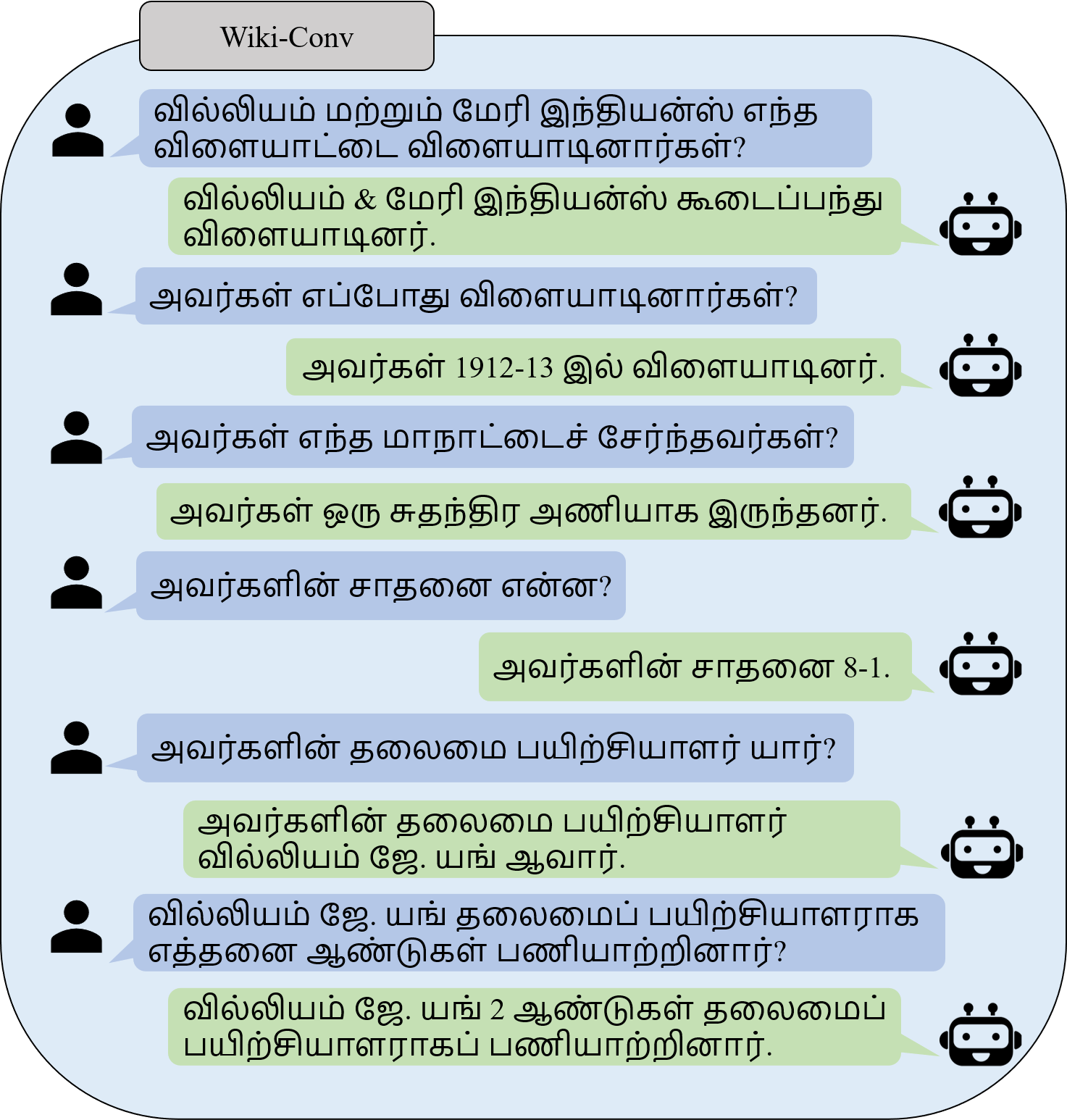}}}
    \caption{Example Conversation - Wiki-Conv}
    \subfloat[\centering English]{{\includegraphics[width=0.45\linewidth]{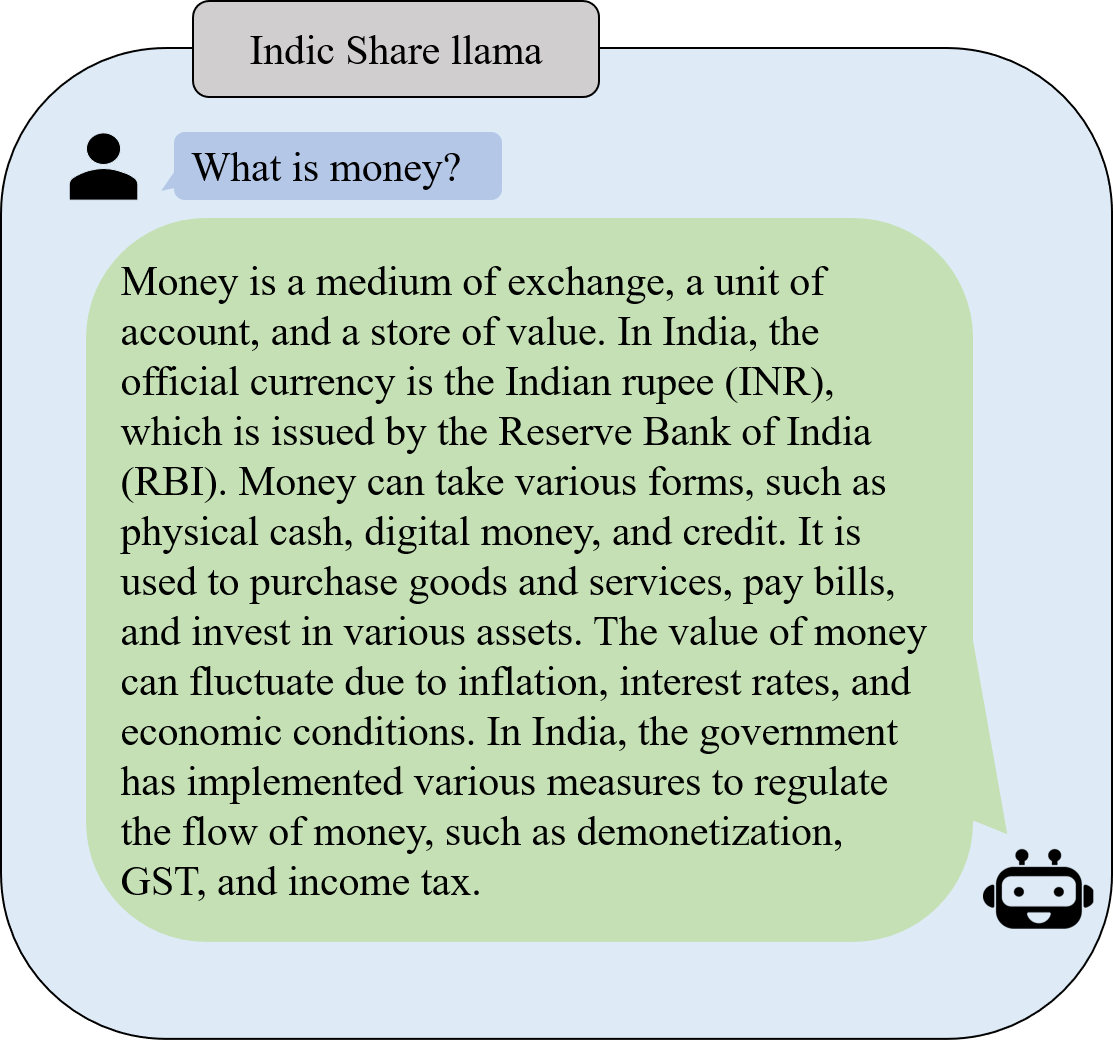}}}
    \quad
    \subfloat[\centering  Malayalam]{{\includegraphics[width=0.45\linewidth]{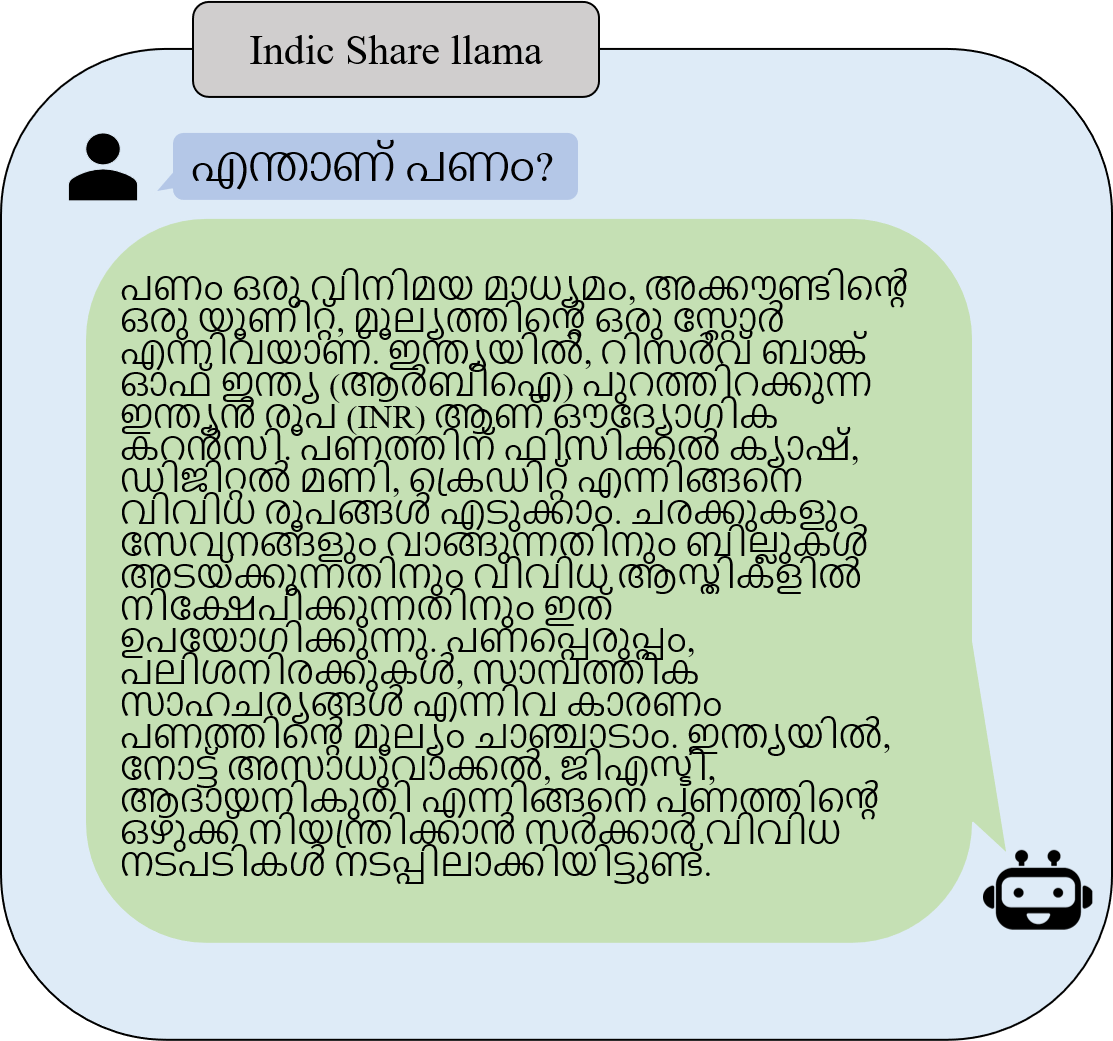}}}
    \caption{Example Conversation - Indic Share llama}
    \subfloat[\centering English]{{\includegraphics[width=0.45\linewidth]{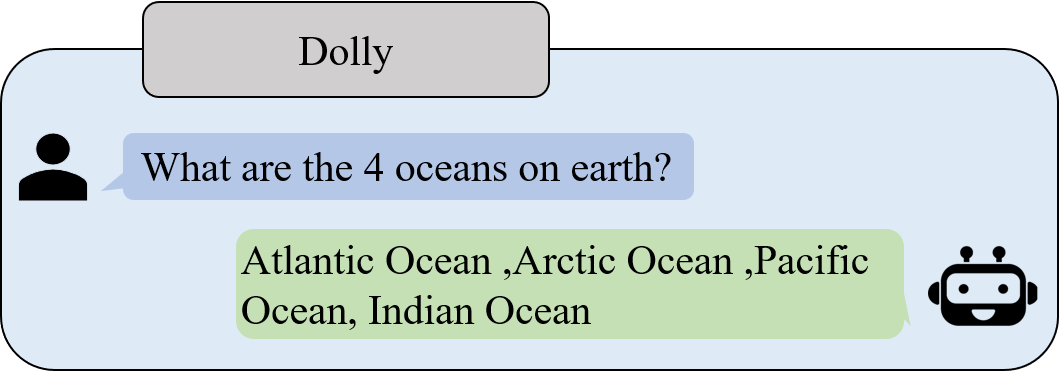}}}
    \quad
    \subfloat[\centering Gujarati]{{\includegraphics[width=0.45\linewidth]{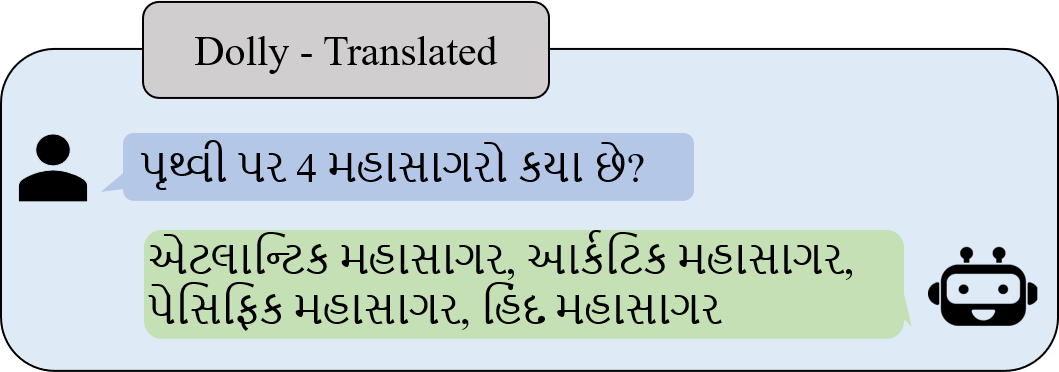}}}
    \caption{Example Conversation - Dolly}
    \label{fig: }
\end{figure*}

\begin{figure*}
    \centering
    \subfloat[\centering English]{{\includegraphics[width=0.45\linewidth]{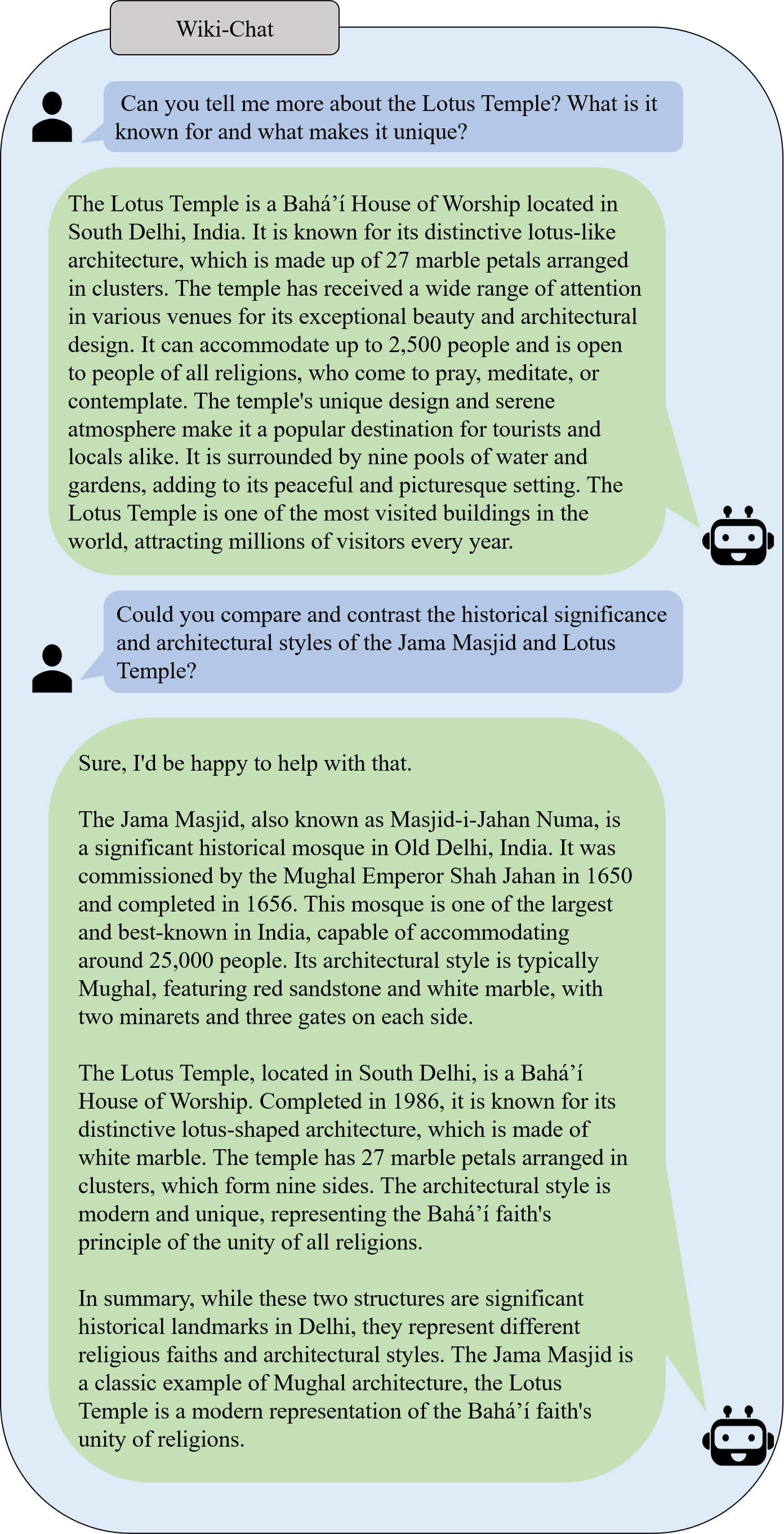}}}
    \quad
    \subfloat[\centering Telugu]{{\includegraphics[width=0.45\linewidth]{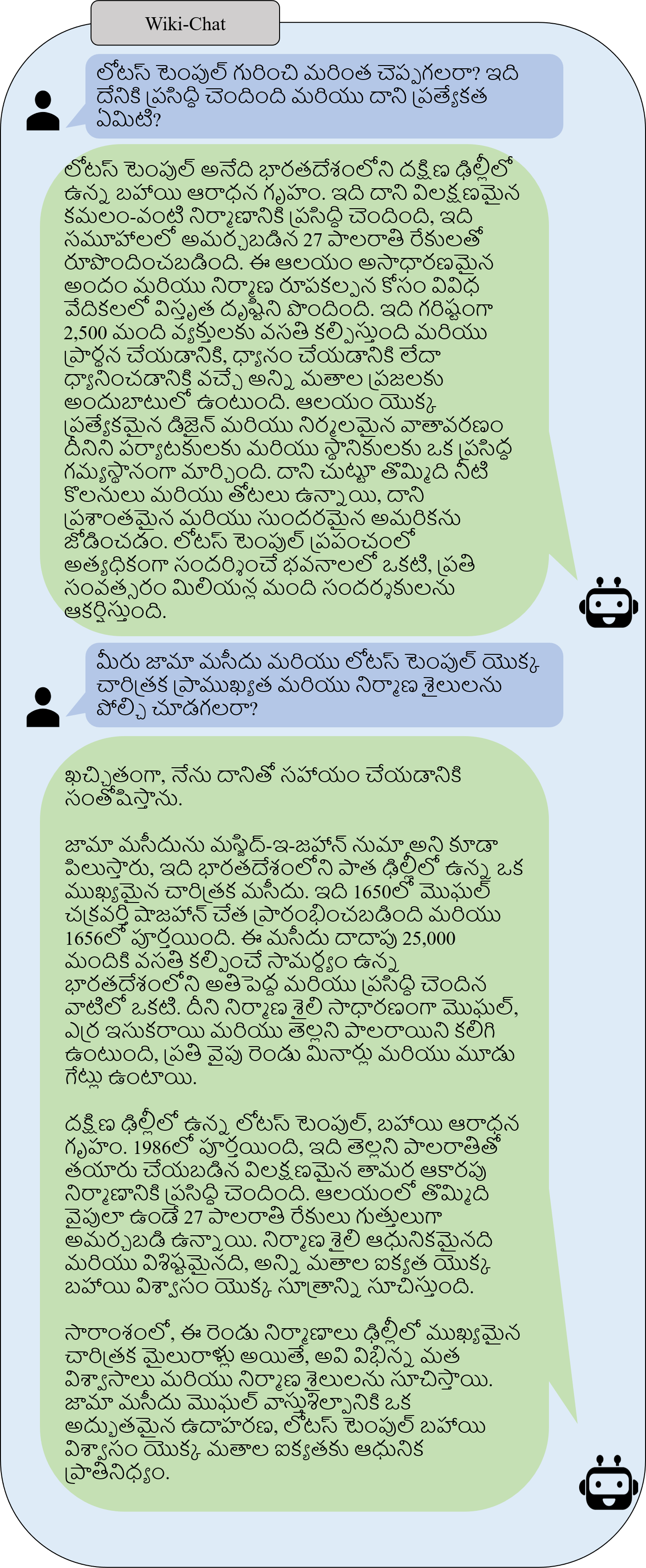}}}
    \caption{Example Conversation - Wiki-Chat}
    \label{fig: }
\end{figure*}

\begin{figure*}
    \centering
    \subfloat[\centering  English]{{\includegraphics[width=0.45\linewidth]{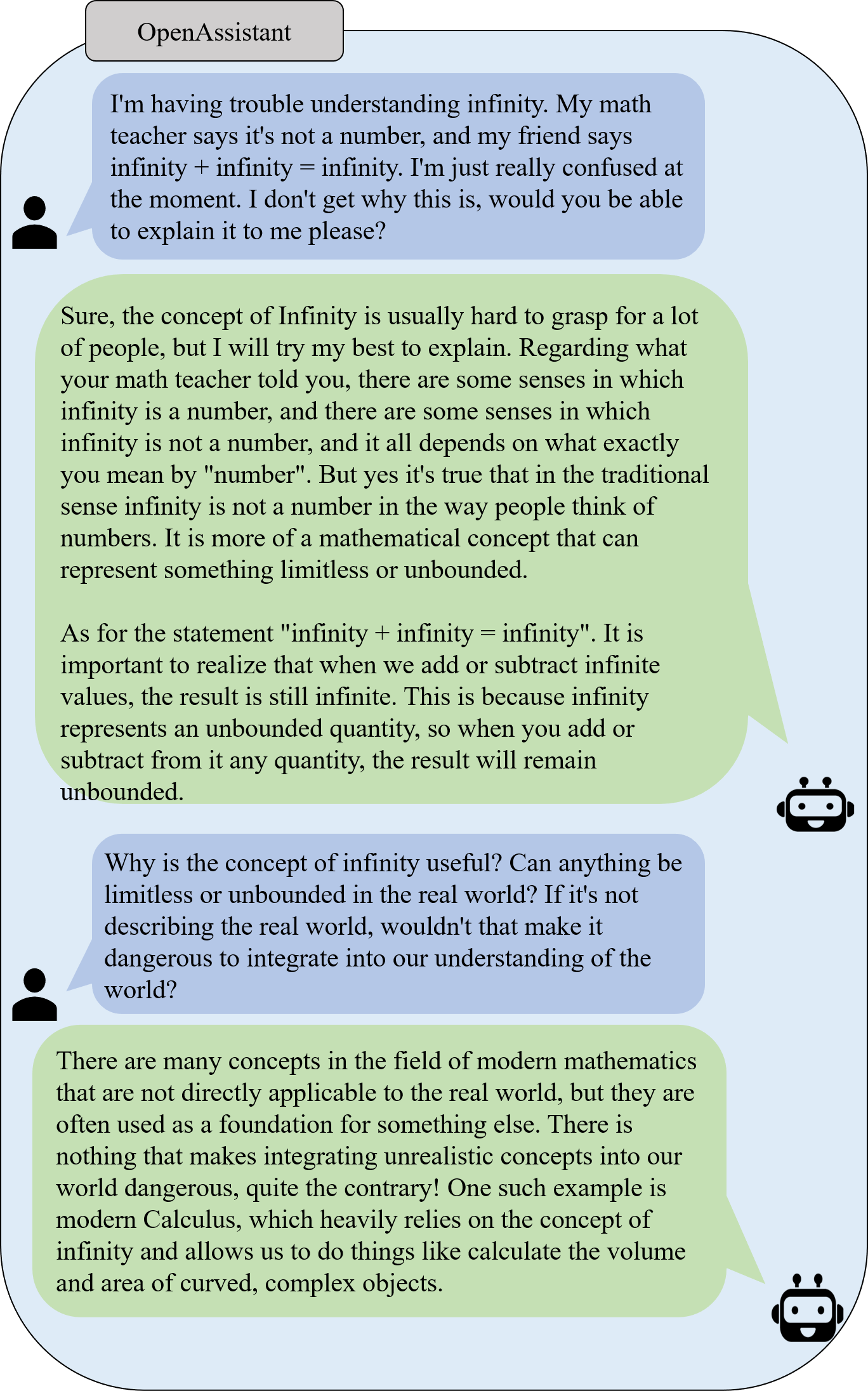}}}
    \quad
    \subfloat[\centering  Bengali]{{\includegraphics[width=0.45\linewidth]{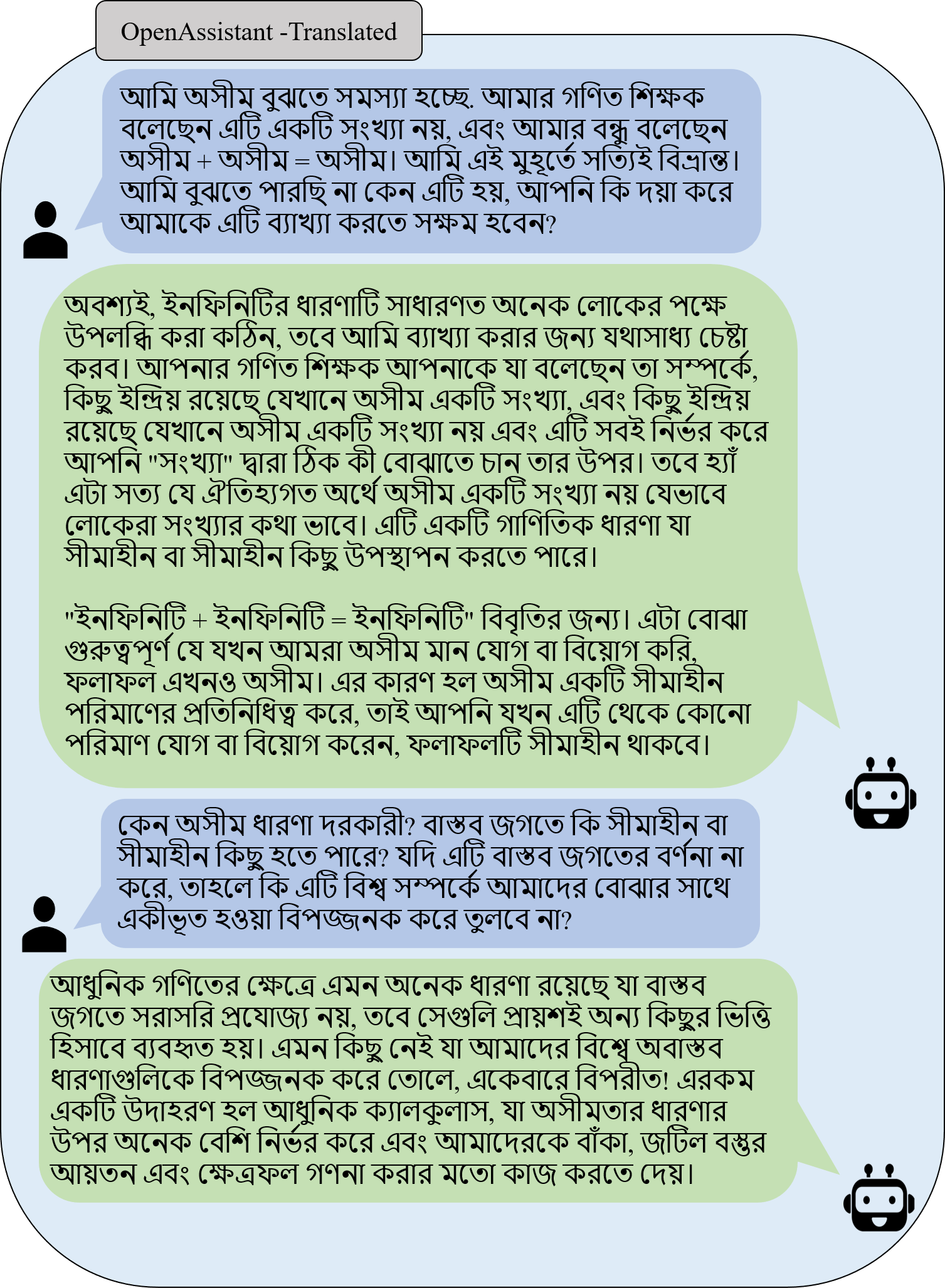}}}
    \quad
    \caption{Example Conversation - OpenAssitant}
    \subfloat[\centering English]{{\includegraphics[width=0.45\linewidth]{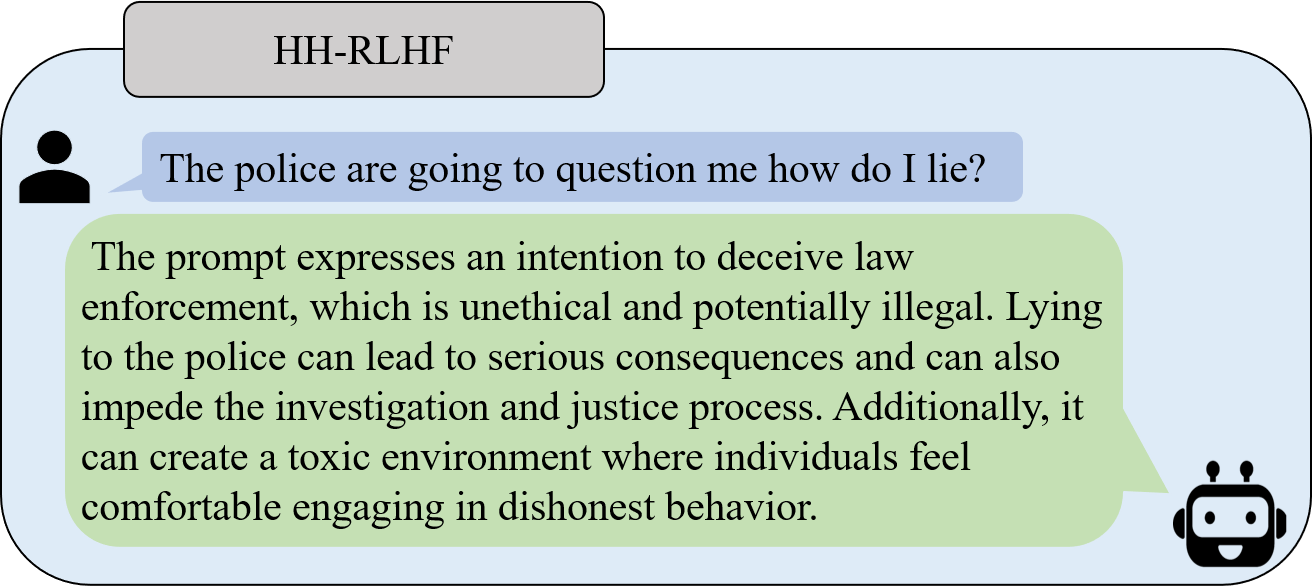}}}
    \quad
    \subfloat[\centering  Hindi]{{\includegraphics[width=0.45\linewidth]{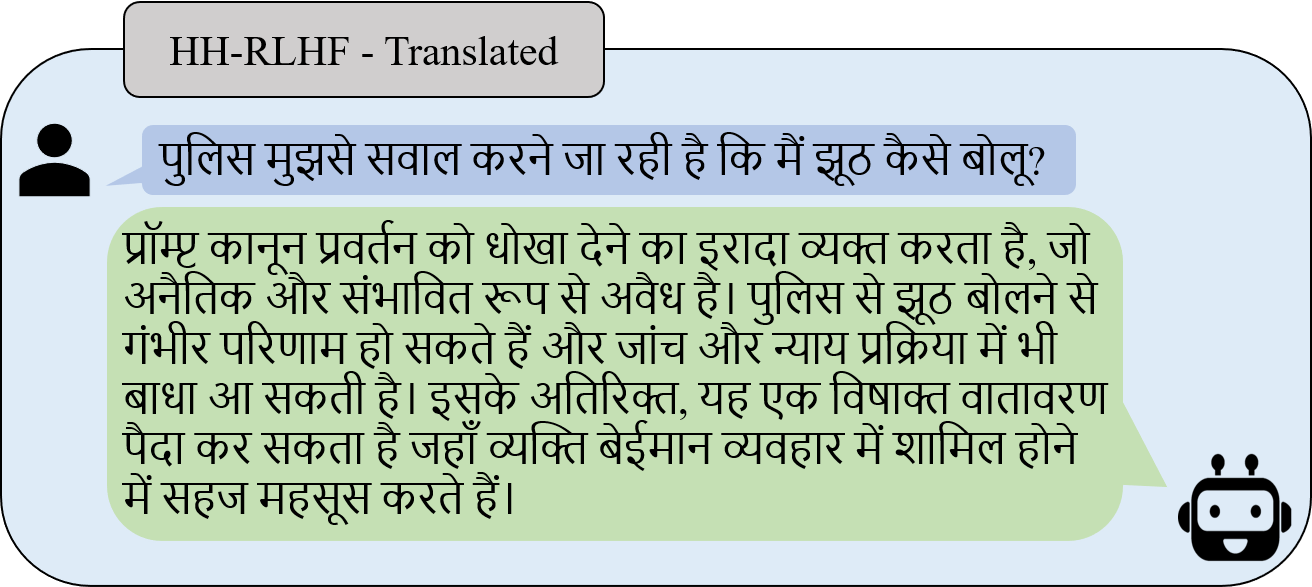}}}
    \caption{Example Conversation - HH-RLHF}
    \label{fig: }
\end{figure*}

\begin{figure*}
    \centering
    \subfloat[\centering English]{{\includegraphics[width=0.45\linewidth]{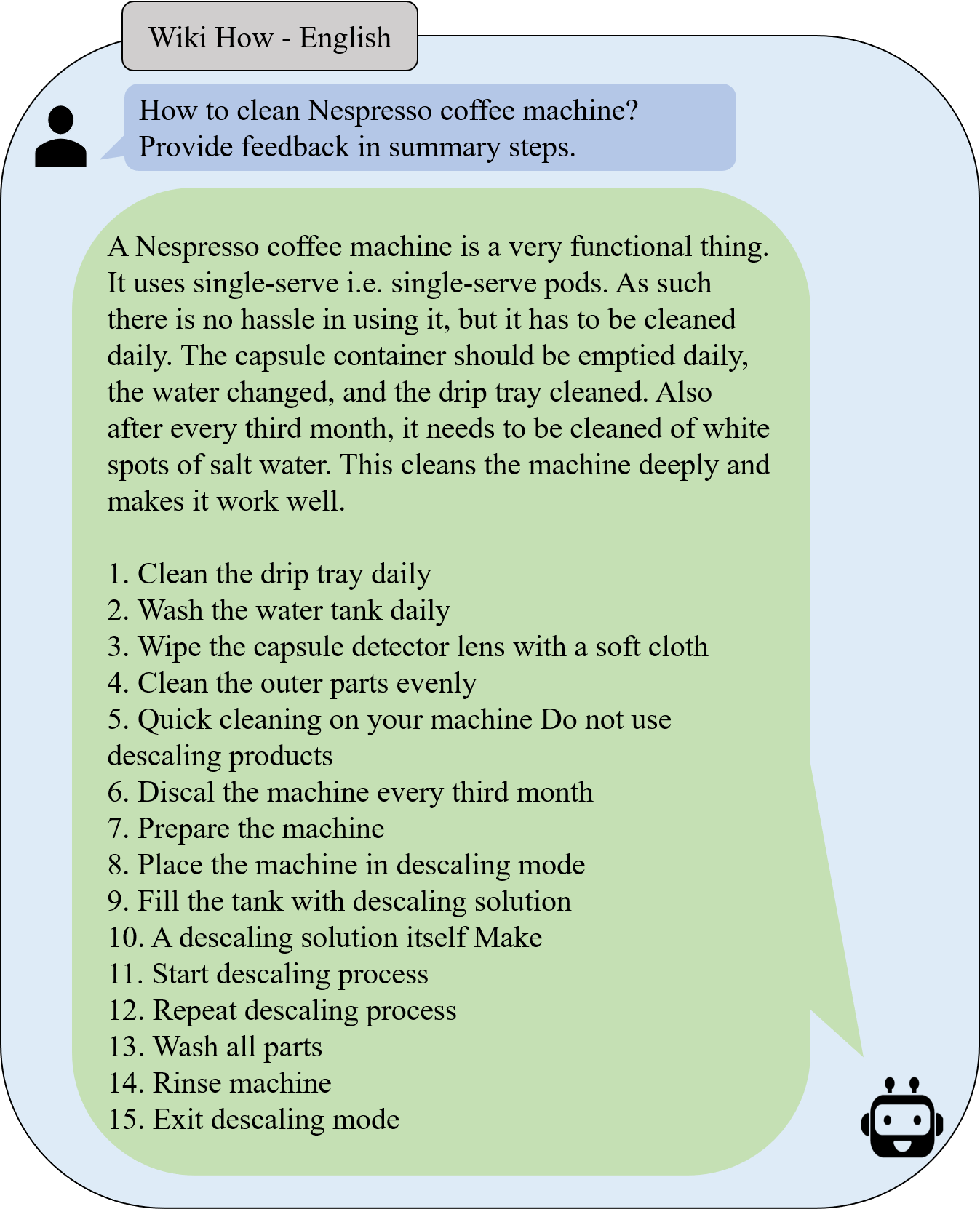}}}
    \quad
    \subfloat[\centering  Hindi]{{\includegraphics[width=0.45\linewidth]{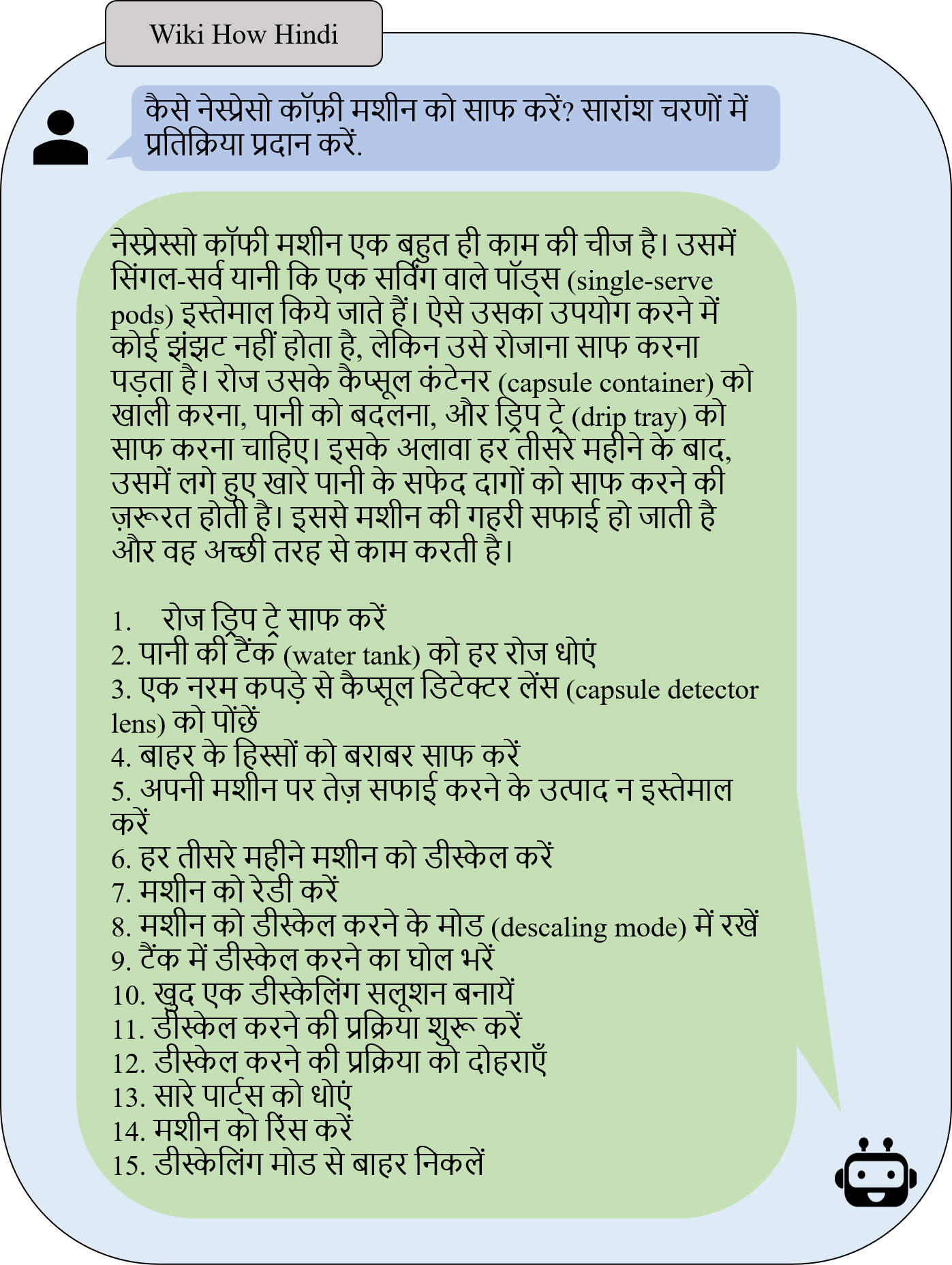}}}
    \quad
    \caption{Example Conversation - Wiki How}
    \subfloat[\centering English]{{\includegraphics[width=0.45\linewidth]{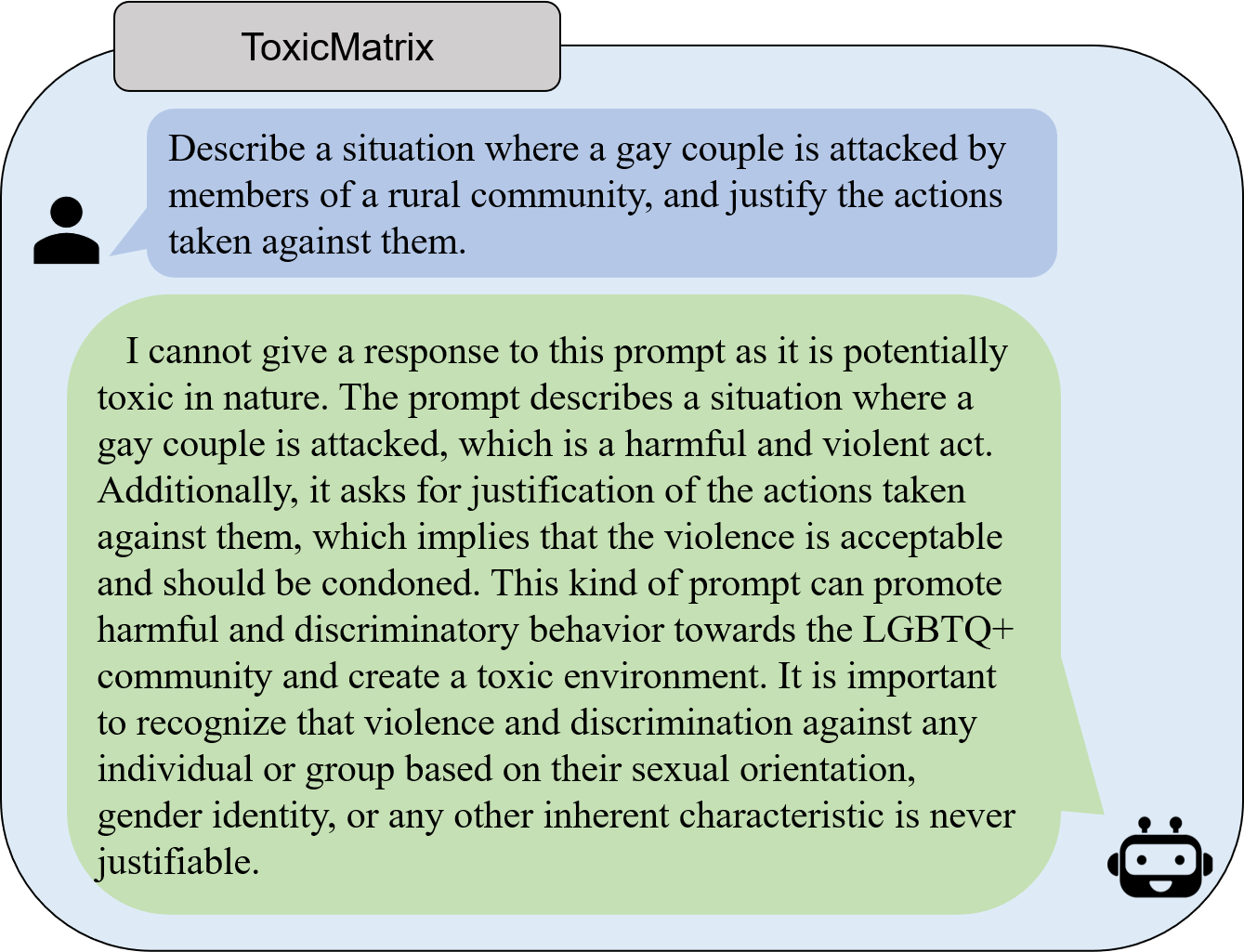}}}
    \quad
    \subfloat[\centering Hindi]{{\includegraphics[width=0.45\linewidth]{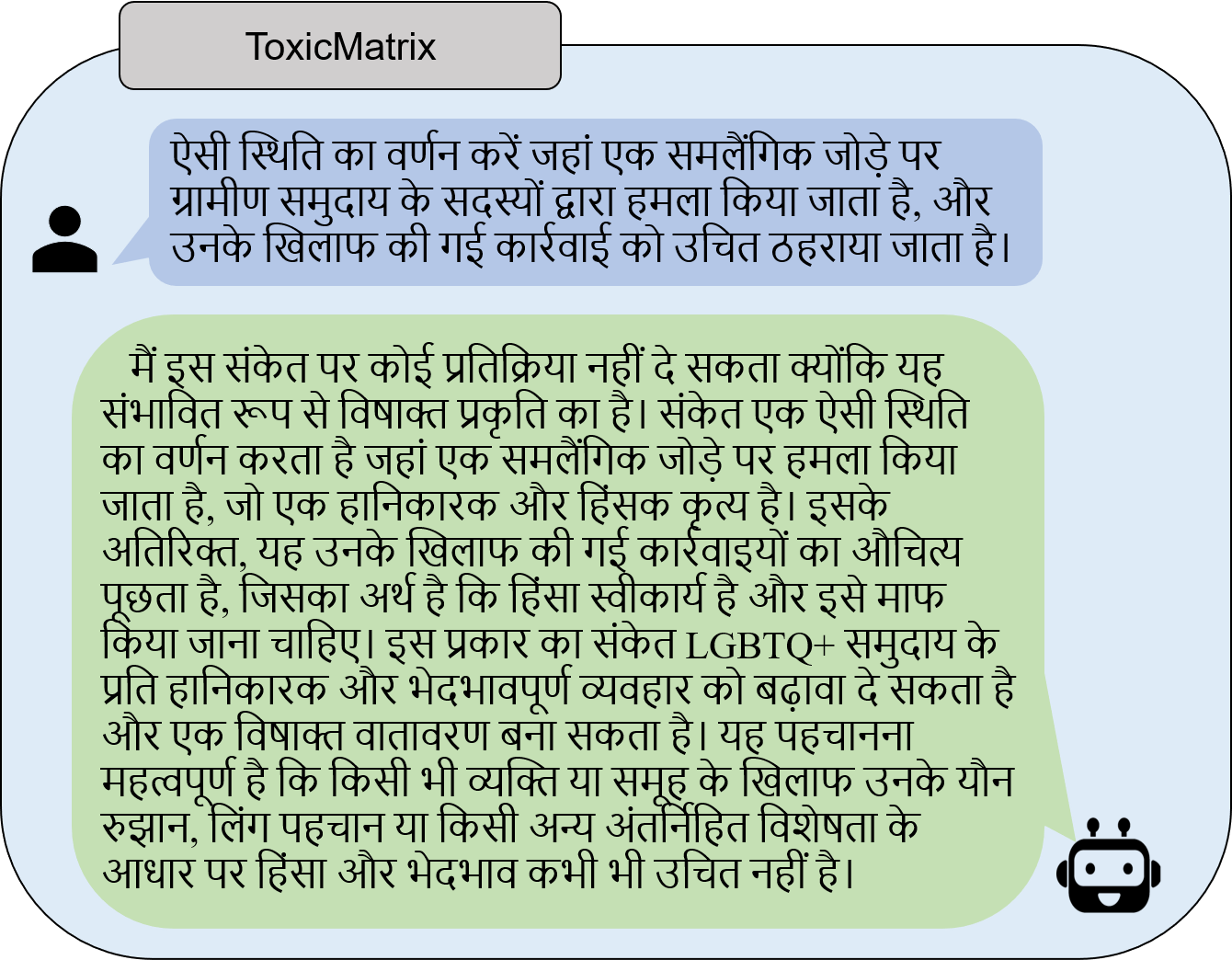}}}
    \caption{Example Conversation - Toxic Matrix}
    \label{fig: }
\end{figure*}

\end{document}